\documentclass[10pt]{article} 

\usepackage[utf8]{inputenc}
\usepackage{amsmath}
\usepackage{amsfonts}
\usepackage{amssymb}
\usepackage{mathtools}
\usepackage{listings}
\usepackage{mathrsfs}
\usepackage{times}
\usepackage{float}
\usepackage{color}
\usepackage{setspace}
\usepackage{color,soul}

\usepackage{amsmath,amsfonts,amsthm,eucal}
\usepackage{enumerate}
\usepackage[letterpaper,margin=3cm]{geometry}
\usepackage{float}
\usepackage[title]{appendix}
\usepackage{color}
\usepackage{tabularx}
\usepackage{siunitx}
\usepackage[tableposition=below]{caption}
\usepackage{array}

\usepackage[
pdfstartview=XYZ,
bookmarks=true,
colorlinks=true,
linkcolor=blue,
urlcolor=blue,
citecolor=blue,
pdftex,
bookmarks=true,
linktocpage=true,   
hyperindex=true
]{hyperref}

\usepackage{lipsum}
\usepackage{lineno}

\usepackage[FIGBOTCAP,TABTOPCAP,bf,tight]{subfigure}

\usepackage{natbib}
\setcitestyle{authoryear}

\usepackage[pdftex]{graphicx}
\usepackage{epstopdf}
\usepackage{booktabs} 
\usepackage{tikz,mathpazo}
\usetikzlibrary{shapes.geometric, arrows}
\usepackage{caption}

\newcommand{\tensor}[1]{\ensuremath{\boldsymbol{#1}}}

\DeclareMathOperator{\grad}{\nabla^{\tensor x}}
\DeclareMathOperator{\diver}{\nabla^{\tensor x}\cdot}

\DeclareMathOperator{\tr}{tr}

\DeclareMathAlphabet{\mathpzc}{OT1}{pzc}{m}{it}

\DeclareMathOperator*{\argmin}{arg\,min}

\theoremstyle{remark}

\newtheorem{remark}{Remark}
\renewcommand{\vec}[1]{\ensuremath{\boldsymbol{#1}}}

\usepackage[super]{nth}
\usepackage{array}
\newcommand{\fig}[1]{Fig. #1}
\newcommand{\Fig}[1]{Figure #1}
\newcommand{\norm}[1]{\left\lVert#1\right\rVert}

\newcommand{\mat}[1]{\ensuremath{\tensor{#1}}}

\newcommand{\dataSetFully}{\mathcal{D}^{\mathrm{sf}}}
\newcommand{\dataSetSolid}{\mathcal{D}^{\mathrm{s}}}
\newcommand{\dataSetFluid}{\mathcal{D}^{\mathrm{f}}}
\newcommand{\psSolid}{\mathcal{Z}^{\mathrm{s}}}
\newcommand{\psElemSolid}{\vec{z}^{\mathrm{s}}}
\newcommand{\psFluid}{\mathcal{Z}^{\mathrm{f}}}
\newcommand{\psElemFluid}{\vec{z}^{\mathrm{f}}}
\newcommand{\psFully}{\mathcal{Z}^{\mathrm{sf}}}
\newcommand{\psElemFully}{\vec{z}^{\mathrm{sf}}}
\newcommand{\distFluid} {d^{\mathrm{f}}}
\newcommand{\distSolid}{d^{\mathrm{s}}}
\newcommand{\distFully}{d^{\mathrm{sf}}}

\usepackage{algorithm}
\usepackage[noend]{algpseudocode}

\title{An accelerated hybrid data-driven/model-based approach for poroelasticity problems with multi-fidelity multi-physics data} 

\begin{document}

\author{Bahador Bahmani\thanks{Department of Civil Engineering and Engineering Mechanics, 
 Columbia University, 
 New York, NY 10027.     \textit{bb2969@columbia.edu}  } 
\and
        WaiChing Sun\thanks{Department of Civil Engineering and Engineering Mechanics, 
 Columbia University, 
 New York, NY 10027.
  \textit{wsun@columbia.edu}  (corresponding author)   }
}

\maketitle
\begin{abstract}
We present a hybrid model/model-free data-driven approach to solve poroelasticity problems. Extending the data-driven modeling 
framework originated from \citet{kirchdoerfer2016data}, we introduce 
one model-free and two hybrid model-based/data-driven formulations capable of simulating the coupled diffusion-deformation of fluid-infiltrating porous media with different amounts of available data. 
To improve the efficiency of the model-free data search, we 
introduce a distance-minimized algorithm accelerated by a k-dimensional tree search. To handle the different fidelities of the solid elasticity and fluid hydraulic constitutive responses, we introduce a hybridized model in which either the solid and the fluid solver 
can switch from a model-based to a model-free approach depending on the availability and the properties of the data. Numerical experiments are designed to verify the implementation and compare the performance of the proposed model to other alternatives.  
\end{abstract}


\section{Introduction}
\label{intro}
The theory of poromechanics attempts to capture how infiltrating pore fluid interacts with the solid skeleton formed by the solid constituents at the scale of the representative elementary volume where an effective medium can be established 
 \citep{terzaghi1996soil, biot1941general, coussy2004poromechanics}. 
Poroelasticity is a sub-discipline of poromechanics problem that focuses on the path independent response of porous media. It has important applications 
across multiple disciplines including seismology \citep{cocco2002pore, chambon2001effects}, 
hydraulic fracture \citep{detournay1993fundamentals, detournay2016mechanics}, 
petroleum engineering, reservoir management, geological disposal \citep{sun:2015, na2017computational},
 and biomechanics modeling for soft tissues and bones \citep{cowin1999bone}. 
 
Due to the multiphysics nature of the coupled diffusion-deformation process, 
poromechanics models must combine field equations which 
constraints field variables to obey 
the balance principle with coupled constitutive models which characterize material laws for pore fluid and solid skeleton \citep{zienkiewicz1999computational, sun2013unified, wang2017unified, wang2019updated, de2017multiscale}. 
One important issue that affects the practicality, accuracy, and robustness of the poroelasticity model is the difference in fidelity of the solid and fluid constitutive laws. In particular, elastic responses of a variety of porous media such as sandstone \citep{renaud2013hysteretic}, clay \citep{borja1997coupling, bryant2019micromorphically, na2019configurational}, sand \citep{cameron2009constitutive} and bone \citep{cowin1999bone} can be captured quite adequately with the existing state-of-the-art models \citep{borja2013plasticity} such that the error of a well-calibrated prediction is often within a few percentages.
Hence, a simple elasticity model calibrated with simple compression or shear tests is often sufficient to make forecasting predictions with a 
narrow confidence interval. 

However, owing to the difficulty to conduct highly precise experiments and the lack of parametric space to characterize the hydraulic responses, a typical prediction of permeability based on the porosity-permeability model is expected to have a much higher variance and, in many cases, is considered accurate even if predicted benchmark permeability is just within the same order of magnitude \citep{paterson2005experimental,sun2011multiscale, sun2011connecting,  andra2013digital, sun2018prediction}. 
In this case, a calibrated hydraulic model that minimizes the mean square error of the Darcy's velocity or pressure gradient does not yield a reliable forecast 
due to the much wider confidence intervals. 
This disparity in the fidelity of the elasticity and hydraulic models for porous media has also been consistently observed in large-scale multi-research-group benchmark studies such as \citet{andra2013digitalpart1, andra2013digital} and has become a major bottleneck for poromechanics models. 

An alternative to handle this disparity is to introduce a variational model-free approach for the poromechanics problem. First introduced by \citet{kirchdoerfer2016data} and later extended for the constitutive manifold \citep{ibanez2017data,  he2020physics}, incorporated with digital image correlation \citep{leygue2018data} and adopted to diffusion \citep{nguyen2020variational}, the data-driven approach enables one to make predictions on the most plausible constitutive responses via a distance minimization algorithm. This distance minimization algorithm then chooses either a data point from a material point database or a linear embedding manifold that minimizes the error from the conservation laws 
and compatibility conditions such that a physics simulation can be carried out without an explicitly derived constitutive law. 
While this approach holds great promise when a large amount of data is available especially for a low-dimensional prediction (e.g., heat flux in 2D, axial stress of a truss element), the predictions in three-dimensional space for arbitrary loading paths can be difficult when the amount of data is not sufficient or not distributed with a sufficient density in the parametric space. 

In this work, we introduce a more flexible hybrid approach where the solid constitutive law can be either model-based (when the fidelity of the elasticity model is sufficiently high) or model-free (when there are sufficient data points). 
Meanwhile, the hydraulic model is replaced by the model-free approach to avoid the usage of models with high deviations to make predictions. 
To cut the CPU time required to conduct the search for the closest data point, 
we introduce a k-dimensional tree search that helps organizing data points to accelerate the simulations. Numerical experiments are then conducted to examine and compare the fully model-based, the fully data-driven, and the hybrid models. 

The remaining parts of the paper will proceed as follows. We first introduce the formulations that enable data-driven algorithm to replace parts or all of the constitutive laws required to generate incremental solution updates for the poroelasticity problems (Section \ref{sec:hybridizd}). 
We then introduce a search strategy that enables us to accelerate the time used to search for the optimal data points which are often the bottleneck of the speed for the data-driven models (Section \ref{sec:localminimization}). 
Section \ref{sec:algoDD} overviews the data-driven algorithm and its implementation aspects. 
Numerical experiments are then conducted to verify the formulation and test the accuracy, robustness, and fidelity of the hybridized and data-driven models (Section \ref{sec:numericalexamples}) followed by a brief conclusion that summarized the major findings.  

As for notations and symbols, bold-faced letters denote tensors (including vectors which are rank-one tensors); 
the symbol '$\cdot$' denotes a single contraction of adjacent indices of two tensors
(e.g.\ $\vec{a} \cdot \vec{b} = a_{i}b_{i}$ or $\tensor{c} \cdot \vec{d} = c_{ij}d_{jk}$ );
the symbol `:' denotes a double contraction of adjacent indices of tensor of rank two or higher
(e.g.\ $\tensor{C} : \vec{\varepsilon^{e}}$ = $C_{ijkl} \varepsilon_{kl}^{e}$);
the symbol `$\otimes$' denotes a juxtaposition of two vectors
(e.g.\ $\vec{a} \otimes \vec{b} = a_{i}b_{j}$)
or two symmetric second-order tensors
(e.g.\ $(\vec{\alpha} \otimes \vec{\beta})_{ijkl} = \alpha_{ij}\beta_{kl}$).

\section{Hybridized data-driven/model-based poromechanics problem}
\label{sec:hybridizd}

\begin{figure}[h]
 \centering
\includegraphics[width=0.6\textwidth]{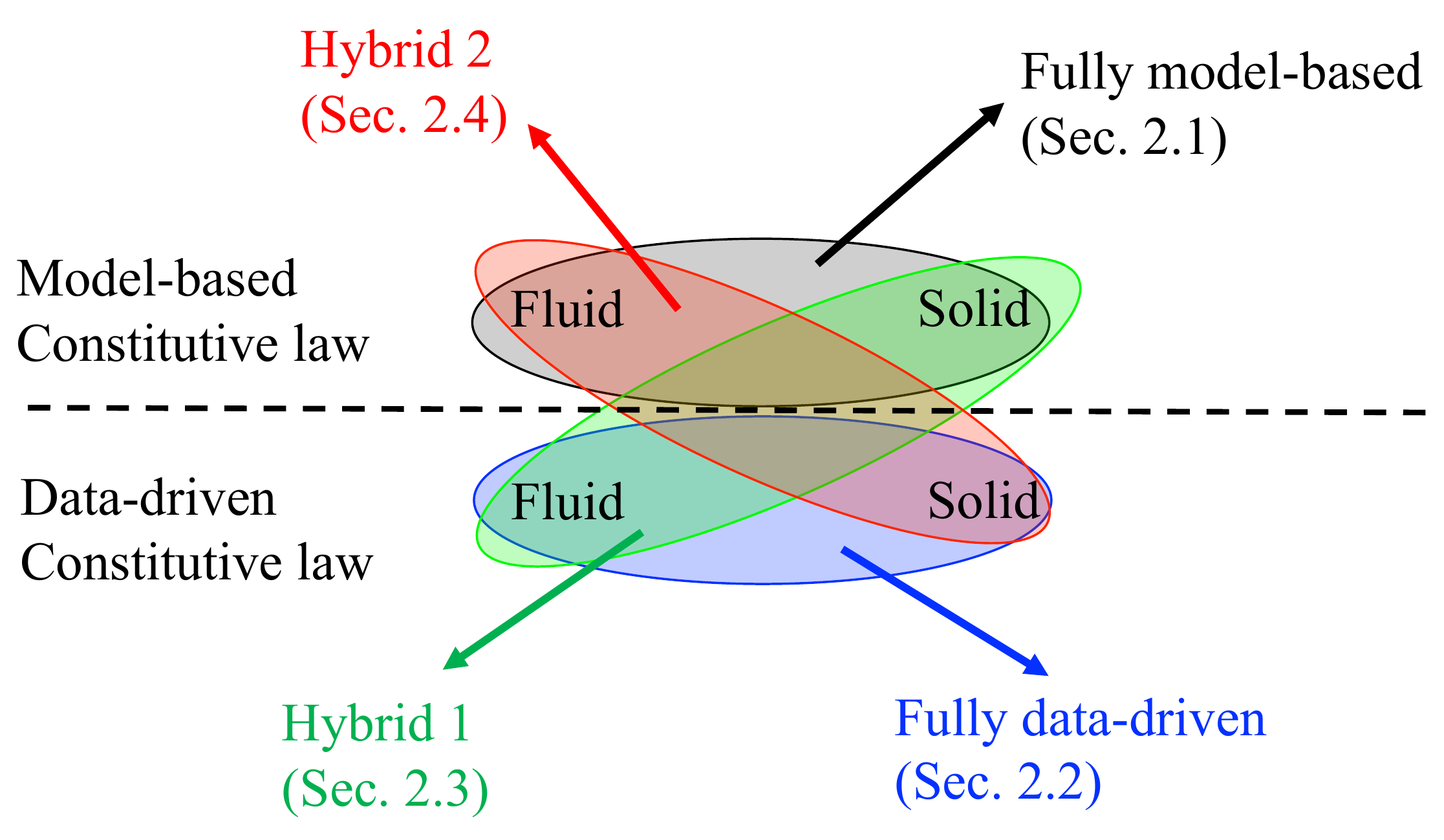}
  \caption{Subsections arrangement based on the constitutive assumptions for solid and fluid phases. The first row indicates the fully model-based assumption. The second row indicates the fully data-driven assumption.
  \label{fig::subsec-overview}}
\end{figure}

This section presents formulations that solve poroelasticity problems
either in a fully model-free fashion or in a hybridized model where either 
the solid or fluid constitutive law is replaced by a model-free data-driven approach. 
For completeness, we first review the classical poromechanics problem in which the solid displacement and pore pressure are the primary unknown variables. 
We then introduce a data-driven model-free algorithm to completely replace classical solid and fluid constitutive laws with the corresponding data-driven algorithm searched for the optimal data points. Finally, two hybridized formulations are provided in each of which either solid or fluid constitutive law follows the classical model-based approach.
The pros and cons of each formulation for different situations (e.g., availability of data, fidelity consideration, robustness) are discussed and will be further elaborated in the numerical experiments shown in Section \ref{sec:numericalexamples}. Figure \ref{fig::subsec-overview} overviews the main topic of each subsection of the current section.

\subsection{A brief review of field equations for poroelasticity problems}
For completeness, we provide a concise review of the poroelasticity model, 
which consists of two major components, i.e., the field theory that provides the necessary constraints for the field variables in the space-time domain and the material laws that provides the local constitutive updates for both the solid skeleton and the fluid constituents. Interest readers may refer to \citet{prevost1985wave, borja1995mathematical, zienkiewicz1999computational, coussy2004poromechanics, Sun.Ostien.Salinger:2013, Sun.Chen.Ostien:2013, na2017computational, na2019configurational} for a more comprehensive treatment for the topic. 

For simplicity, we assume that the deformation of the solid skeleton is, infinitesimal, path independent and elastic such that $\tensor{\epsilon} = \tensor{\epsilon}^{e}$ and $\tensor{\sigma} = \sigma(\tensor{\epsilon}^{e}, p) = \sigma(\tensor{\epsilon},p )$. We also assume that the flow in the pore space is a function of the pore pressure gradient such that the Darcy's velocity $\vec{q} = \vec{q}(\grad p)$ is a function of the pore pressure $p$. 
In this formulation, we follow the trajectory of the solid constituent and express 
the balance principles as a function of solid displacement $\vec{u} = \vec{u}^{s}$ 
and pore pressure $p$. 
By neglecting the inertial term, the balance of linear momentum and mass on the spatial domain $\Omega$ over time $t \in T$ read (cf. \citet{Sun.Chen.Ostien:2013}), 
\begin{align}
	&
	\diver{\tensor{\sigma}}(\tensor{\epsilon}, p) + \vec{\gamma} = \vec{0} \ \mathrm{in} \ \Omega \times T,
	\label{eq:linear-mom-spaceTime}
	\\&
	\frac{\dot{p}}{M} + B \dot{\epsilon}_{\mathrm{vol}}  + \diver{\vec{q}}(\grad p)  + s = 0 \ \mathrm{in} \ \Omega \times T,
		\label{eq:cons-mass-spaceTime}
\end{align}
where $\tensor{\sigma}$ is the total stress and $\epsilon_{\mathrm{vol}} = \tr \tensor{\epsilon}$ is the volumetric strain. We  postulate that the total stress can be partitioned into the effective stress of the solid skeleton and the pore fluid pressure according to the effective stress principle, i.e., 
\begin{equation}
	\tensor{\sigma} = \tensor{\sigma^\prime}(\tensor{\epsilon} ) - Bp\tensor{I} \ \mathrm{in} \ \Omega \times T. 
	\label{eq:eff_strs_rel}
\end{equation} 
Furthermore, 
 $\vec{\gamma}$ is the body force, 
$M$ is the Biot's modulus, $B$ is the Biot's coefficient, $\vec{q}$ is the Darcy's velocity, $s$ is the source (sink) term. 
The definitions of these physical quantities are listed below. 
\begin{align}
	&
	B = 1 - \frac{K}{K_{\mathrm{s}}}  \; , 
	\\&
	\vec{\gamma} = \vec{\gamma}^{\mathrm{s}}+  \vec{\gamma}^{\mathrm{f}} = (1-\phi^{\mathrm{f}}) \rho_{\mathrm{s}} \vec{g} + \phi^{\mathrm{f}} \rho_{\mathrm{f}} \vec{g} \; , 
	\\&
	M = \frac{K_{\mathrm{s}} K_{\mathrm{f}}}{K_{\mathrm{f}} (B - \phi^{\mathrm{f}}) + K_{\mathrm{s}} \phi^{\mathrm{f}} } \; , 
	\\ 
	& 
	\vec{q} = \phi^{f} (\vec{v}^{f} - \vec{v}),
\end{align}
where $K$ and $K_{s}$ are the bulk moduli of the solid skeleton and the solid constituent, respectively, $\gamma^{\mathrm{s}}$ and $\gamma^{\mathrm{f}}$ 
are the partial density of the solid and fluid constituents,
while $\rho_{\mathrm{s}}$ and $\rho_{\mathrm{f}}$ are the intrinsic density of the 
solid and fluid constituents. $\phi^{\mathrm{f}}$ is the porosity,  $K_{\mathrm{f}}$ is the 
bulk modulus of the fluid constituent, and $\vec{v}^{f}$ and $\vec{v}$ are the velocity of the fluid constituent and the solid skeleton, respectively. 

To compute the boundary value problem, the initial and boundary conditions are specified as follows. The initial conditions are $\vec{u} = \vec{u}_0 \ \mathrm{in}  \ \Omega \ \mathrm{at} \ t=0$ and $p = p_0 \ \mathrm{in}  \ \Omega \ \mathrm{at} \ t=0$. Meanwhile, the boundary conditions are $\vec{u} = \bar{\vec{u}} \ \mathrm{on} \ \partial \Omega_{u} \times T$ (prescribed displacement) and $\tensor{\sigma} \cdot \vec{n} = \bar{\vec{t}} \ \mathrm{on} \ \partial \Omega_{\sigma} \times T$ (prescribed traction), 
$ p = \bar{p} \ \mathrm{on} \ \partial \Omega_{p} \times T$ (prescribed pore pressure), 
and $ \vec{q} \cdot \vec{n} = \bar{q} \ \mathrm{on} \ \partial \Omega_{q} \times T$ (prescribed fluid flux) where $\vec{n}$ is an unit normal vector pointing outward to the boundary $\partial \Omega$. 
Furthermore, the following conditions must hold: 
$\partial \Omega_{\sigma} \cup \partial \Omega_{u} = \partial \Omega$, 
$\partial \Omega_{\sigma} \cap \partial \Omega_{u} =  \emptyset$, 
 $\partial \Omega_{q} \cup \partial \Omega_{p} = \partial \Omega$ and
$\partial \Omega_{q} \cap \partial \Omega_{p} =  \emptyset$.

For convenience, we first discretize the governing equations, Eq. \eqref{eq:linear-mom-spaceTime} and Eq. \eqref{eq:cons-mass-spaceTime}, in time via the implicit Euler scheme. 
Given the displacement and pore pressure at time $t_{n}$, 
the time-discretized balance principle  
within the time interval $T \in [t_n, t_{n+1}]$ can be expressed as, 
\begin{align}
	&
	\diver{\tensor{\sigma}_{n+1}}(\tensor{\epsilon}_{n+1}, p_{n+1}) + \vec{\gamma}_{n+1} = \vec{0} \ \mathrm{in} \ \Omega \; , 
	\label{eq:linear-mom-space}
	\\&
\frac{p_{n+1} }{M } + {B}  {\epsilon_{\mathrm{vol}}}_{n+1}   + \diver{\vec{q}_{n+1} (\grad p_{n+1})} \Delta t + s_{n+1} \Delta t = 
\frac{p_{n}}{M } + {B}  {\epsilon_{\mathrm{vol}}}_n 
 \ \mathrm{in} \ \Omega\;  .
		\label{eq:cons-mass-space}
\end{align}

To complete the initial boundary value problem, the classical model-based approach for poroelasticity problem required us to define material constitutive laws such that, given the strain and pore pressure  gradient, we may obtain updated effective stress and Darcy's velocity; the first one maps a given strain to the  effective stress $(\tensor{\epsilon}, \tensor{\sigma}')$, another one maps a given pore pressure gradient to the 
Darcy's velocity $(\grad p, \vec{q})$. 
In what follows, our goal is to introduce a new formulation such that \textit{at least one} of these two constitutive laws, $(\tensor{\epsilon}, \tensor{\sigma}')$ and $(\grad p, \vec{q})$ can be replaced by the model-free data-driven approach originally designed for elasticity problem in  \citet{kirchdoerfer2016data}. 

\remark{Notice that, for porous media with compressible constituents, 
 both the Biot's coefficient and the Biot's modulus may depend on the bulk modulus of the skeleton and hence may evolve if the elastic response of the solid skeleton 
is nonlinear. 
Here we limit our scope to the types of porous media 
with the bulk modulus of both constituents significantly larger than 
the effective bulk modulus of the solid skeleton. As such, the evolution 
of both the Biot's coefficient and Biot's modulus, $B$ and $M$, are neglected \citep{biot1941general, terzaghi1943theoretical, cryer1963comparison}. 
 This assumption and the effective stress principle together enable us to treat the constitutive laws for the solid deformation and fluid flow as two independent ordinary differential equations.}\label{rmk::const-biot} 

\subsection{Option 1: Pure data-driven poroelasticity} \label{sec:fullyDD}
Here, our goal is to present a new formulation constrained by the time-discretized balance principle listed in Eqs. \eqref{eq:linear-mom-space} and \eqref{eq:cons-mass-space} without employing any constitutive laws. 
To do so, we assume that there exists two databases, each have a finite number of data points for $(\tensor{\epsilon}, \tensor{\sigma}^\prime)$ and $(\grad p, \vec{q})$.
We further assume that there are sufficient data points distributed in the parametric space for both material laws such that a complete model-free approach is feasible. Notice that the data-driven model-free method developed in this research is categorized into non-parametric learning methods \citet{goodfellow2016deep} which are basically considered data-demanding since they have minimum assumptions about the data-driven model. This is in contrast to the parametric methods such as neural networks \citep{ghaboussi1998autoprogressive, wang2018multiscale, tartakovsky2020physics} where there is a stronger pre-assumption about the data-driven model.

In the following subsections, we first express the mathematical statement for the data-driven scheme as a double-minimization problem. Then, we provide a numerical strategy based on the fixed-point (staggered) method to solve this minimization problem in two steps: global and local minimization steps. Finally, we provide a numerical solution for the global minimization based on the Lagrange multiplier method. The local minimization step is not covered in this section and will be discussed in section \ref{sec:localminimization} separately since it has its specific treatment which is common between all the three formulations.
\subsubsection{Problem statement}
In this section, we present the fully data-driven, constitutive-model-free formulation for the poroelasticity problem. Our goal is to introduce a minimization problem that constitutes a model-free poroelasticity solver.
As such, we seek solutions from data sets of material responses that weakly satisfy the time-discretized balance principle listed in Eqs. \eqref{eq:linear-mom-space} and \eqref{eq:cons-mass-space} without explicitly 
introducing any constitutive law for either the solid skeleton and the pore fluid. 

If there are an infinite number of error-free data points populating the database, then 
a constitutive manifold can be identified such that one may select elements of the manifold 
that satisfy the constitutive laws. However, material databases is rarely populated 
with enough data and often contains data with noise that makes it impractical 
to impose such a strict requirement. As an alternative, we follow the idea of 
\cite{kirchdoerfer2016data} where we merely seek solutions that satisfy the balance principles while the resultant constitutive responses are the ones closest to but not necessary elements of 
the set of points in the databases. As such, we need to introduce the notion of "distance"  via 
 an appropriate norm we selected for the space of admissible solution  $(\tensor{\epsilon}, \tensor{\sigma})$ and $(\grad p, \vec{q})$. 

In other words, we regard the balance principles as the universal law that
should not be violated. The data-driven method is then designed to 
generate solutions that fulfill the balance principle while the 
material response at the integration points  $(\tensor{\epsilon}, \tensor{\sigma})$ and $(\grad p, \vec{q})$ are all closest to existing data points in the material database. The distance 
between an admissible response satisfied balance laws and an existing data point is then measured by an appropriate norm. 
We will express the above-mentioned statement as a double-minimization problem \citep{kirchdoerfer2016data,kirchdoerfer2018data,he2020physics,nguyen2020variational}. 
For brevity, we assign a new variable for pressure gradient as $\vec{r} = \grad{p}$. 

We then define the phase space for poroelasticity at time $t_{n+1}$ as all $ \psElemFully_{n+1} = (\tensor{\epsilon}_{n+1}, \tensor{\sigma}^\prime_{n+1} , \vec{r}_{n+1} , \vec{q}_{n+1} ) \in \psFully $ where $\psFully = \psSolid \times \psFluid$ is the product space of the solid phase space $\psSolid = V_{\epsilon} \times V_{\sigma^\prime} $ and the fluid phase space $\psFluid = V_{r} \times V_{q}$; 
in which $V_{\epsilon},\ V_{\sigma^{\prime}}=\left[  L^2(\Omega) \right]^{\sum_{A=1}^\text{ndim} A} $ 
are spaces of real-valued symmetric \nth{2} order tensor fields with square integrable components, and $V_{r},\ V_{q} = \left[  L^2(\Omega) \right]^{\text{ndim}}$ are spaces of real-valued vector fields with square integrable components. 
We define the space $\mathcal{C}^{\mathrm{momentum}}_{n+1}$ as follows:
\begin{multline*}
\mathcal{C}^{\mathrm{momentum}}_{n+1} =\left\{
(\tensor{\epsilon}_{n+1}(\vec{u}_{n+1}), \tensor{\sigma}^\prime_{n+1}, p_{n+1}) \in \psSolid \times V_p
\mid 
\diver{\tensor{\sigma}_{n+1}} + \vec{\gamma}_{n+1} = \vec{0} \ \mathrm{in} \ \Omega,\ \vec{u}_{n+1} = \bar{\vec{u}}_{n+1} \ \mathrm{on} \ \partial \Omega_u, \right.\\
\left.
\tensor{\sigma}_{n+1} \cdot \vec{n} = \bar{\vec{t}}_{n+1} \ \mathrm{on}\  \partial \Omega_{\sigma}, \
\tensor{\sigma}_{n+1} = \tensor{\sigma}^\prime_{n+1}-Bp_{n+1}\tensor{I} \ \mathrm{in} \ \Omega
\right\},
\end{multline*}
where $\vec{u}_{n+1} \in V_u$, $V_u= [H^1(\Omega)]^{\text{ndim}}$, $V_p=H^1(\Omega)$, and $H^1$ denotes the Sobolev space of square-integrable functions with square-integrable first derivative and $\text{ndim}$ denotes the dimension of the spatial domain. Here, strain tensor $\tensor{\epsilon}_{n+1}(\vec{u}_{n+1})$ is a derived quantity of displacement vector $\vec{u}_{n+1}$ through small strain relation.
All members in continuous set $\mathcal{C}^{\mathrm{momentum}}_{n+1}$ satisfy time discretized conservation of linear momentum equation, mechanical-related boundary conditions, small deformation relation for strain tensor, and effective stress principle at time $t_{n+1}$. The compatibility conditions of the strain field is automatically satisfied since the small strain tensor is the symmetric part of the displacement field gradient. The space of admissible solutions that satisfy the conservation of mass is denoted by $\mathcal{C}^{\mathrm{mass}}_{n+1}$ and defined as follows: 
\begin{multline*}
\mathcal{C}^{\mathrm{mass}}_{n+1} =\left\{
(\vec{r}_{n+1}(p_{n+1}), \vec{q}_{n+1}, \tensor{\epsilon}_{n+1}(\vec{u_{n+1}})) \in \psFluid \times V_{\epsilon}
\mid 
\frac{p_{n+1} }{M } + {B}  {\epsilon_{\mathrm{vol}}}_{n+1}  + \diver{\vec{q}_{n+1}} \Delta t + s_{n+1} \Delta t=\right.\\
\left.
\frac{p_{n}}{M } + {B}  {\epsilon_{\mathrm{vol}}}_n 
 \ \mathrm{in} \ \Omega,
 p_{n+1} = \bar{p}_{n+1} \ \mathrm{on} \ \partial \Omega_{p},\ 
 \vec{q}_{n+1}\cdot \vec{n} = \bar{q}_{n+1} \ \mathrm{on} \ \partial \Omega_{q}
\right\},
\end{multline*}
where the gradient of pore pressure $\vec{r}_{n+1}(p_{n+1})$ is a derived quantity of pressure field herein.
The admissible solution space $\mathcal{C}^{\mathrm{coupled}}_{n+1}$ for the poroelasticity problem at time $t_{n+1}$ includes field variables $\tensor{\epsilon_{n+1}}(\vec{u}_{n+1})$, $\tensor{\sigma}_{n+1}^\prime$, $\vec{r}_{n+1}(p_{n+1})$, and $\vec{q}_{n+1}$ that satisfy all physical constraints at time $t_{n+1}$. This space exists at the intersection of above defined continuous sets $\mathcal{C}^{\mathrm{coupled}}_{n+1} = \mathcal{C}^{\mathrm{momentum}}_{n+1} \cap \mathcal{C}^{\mathrm{mass}}_{n+1}$. The poroelasticity database at time $t_{n+1}$ is denoted by $\dataSetFully_{n+1} \subset \psFully$. This discrete set contains a finite number of elements that stores experimental data points corresponding to the poroelasticity constitutive laws at time $t_{n+1}$. The poroelasticity database format and its properties will be clarified in details later. 

The data-driven solution $\bar{\vec{z}}^{\mathrm{sf}}_{n+1} \in \psFully$ at time $t_{n+1}$ is the solution of the following double-minimization problem:
\begin{equation}
\bar{\vec{z}}^{\mathrm{sf}}_{n+1}  = \underset{\mathcal{C}^{\mathrm{coupled}}_{n+1}}{\arg} \left\{  
\min_{{\psElemFully}^*_{n+1}  \in \dataSetFully_{n+1}}
\min_{\psElemFully_{n+1}  \in \mathcal{C}^{\mathrm{coupled}}_{n+1}} 
\norm{{\psElemFully}^*_{n+1}  - \psElemFully_{n+1} }_{\psFully}^{2}
\right\},
\label{eq::fullyDD-prob-def}
\end{equation}
where ${\psElemFully}^*_{n+1}=(\tensor{\epsilon}^*_{n+1}, {\tensor{\sigma}^\prime}^*_{n+1}, \vec{r}^*_{n+1}, \vec{q}^*_{n+1})$ is an element of poroelasticity database, \\
${\psElemFully}_{n+1}=(\tensor{\epsilon}_{n+1}(\vec{u}_{n+1}), {\tensor{\sigma}^\prime}_{n+1}, \vec{r}_{n+1}(p_{n+1}), \vec{q}_{n+1})$ is an element of admissible poroelasticity solution space, and $\norm{\cdot}_{\psFully}$ is a norm associated with the space $\psFully$ which measures closeness of ${\psElemFully}^*_{n+1}$ and ${\psElemFully}_{n+1}$. We define the subtract operation on space $\psFully$ as follows:
\begin{equation}
{\psElemFully}^*_{n+1}  - \psElemFully_{n+1}  = (\tensor{\epsilon}^*_{n+1} - \tensor{\epsilon}_{n+1}, {\tensor{\sigma}^\prime}^*_{n+1}-{\tensor{\sigma}^\prime}_{n+1}, \vec{r}^*_{n+1}-\vec{r}_{n+1}, \vec{q}^*_{n+1}-\vec{q}_{n+1}) \in \psFully.
\label{eq::subtract-proelast-space}
\end{equation}
We define the product norm for the poroelasticity product phase space $\psFully$ as follows:
\begin{equation}
\norm{\psElemFully_{n+1} }_{\psFully} = \sqrt{\norm{\psElemSolid_{n+1} }_{\psSolid}^{2} + \Delta t \norm{\psElemFluid_{n+1} }_{\psFluid}^{2}}
\label{eq::norm-fully}
\end{equation}
where $\psElemSolid_{n+1}=(\tensor{\epsilon}_{n+1}, \tensor{\sigma}^\prime_{n+1}) \in \psSolid$ , $\psElemFluid_{n+1}=(\vec{r}_{n+1}, \vec{q}_{n+1}) \in \psFluid$
and $\norm{\cdot}_{\psSolid}$ and $\norm{\cdot}_{\psFluid}$ are the norms for the respectively solid and fluid phase spaces.  
The time step size $\Delta t$ in the second term of Eq. \eqref{eq::norm-fully} is introduced to make the unit consistent. 
 We define the following norm for $\psSolid$ :
\begin{equation}
 \norm{\psElemSolid_{n+1}}_{\psSolid}^{2} =  \norm{(\tensor{\epsilon}_{n+1}, \tensor{\sigma}^\prime_{n+1})}_{\psSolid}^{2}= \int_{\Omega} \frac{1}{2} \tensor{\epsilon}_{n+1} : \tensor{\mathbb{C}}_{\mathrm{s}} :\tensor{\epsilon}_{n+1} +\frac{1}{2} \tensor{\sigma}^\prime_{n+1} : \tensor{\mathbb{S}}_{\mathrm{s}} :\tensor{\sigma}^\prime_{n+1} d\Omega,
\label{eq::norm-solid}
\end{equation}
where $\tensor{\mathbb{C}}_{\mathrm{s}}$ and $\tensor{\mathbb{S}}_{\mathrm{s}}$ are \nth{4} order symmetric positive definite tensors. 
As shown in \citet{kirchdoerfer2016data, he2020physics, nguyen2020variational},
specific $\tensor{\mathbb{C}}_{\mathrm{s}}$ and $\tensor{\mathbb{S}}_{\mathrm{s}}$
can be chosen to from different equivalent norms, provided that both tensors remain 
positive definite. The choices of $\tensor{\mathbb{C}}_{\mathrm{s}}$ and $\tensor{\mathbb{S}}_{\mathrm{s}}$ may affect the values of the norm due to the weighting but the
resultant normed space is topologically identical to a Euclidean space. 

The weighting tensor $\tensor{\mathbb{C}}_{\mathrm{s}}$  share the same unit as the elasticity tensor, i.e., Force/Length$^{2}$, while the unit of $\tensor{\mathbb{S}}_{\mathrm{s}}$ is the reciprocal of that of 
$\tensor{\mathbb{C}}_{\mathrm{s}}$. Both the eigenvalues and the spectral directions 
of these tensors affect the values of norms and therefore change how distance is measured
and could affect the efficiency of the search problems (cf. \citet{mota2016cartesian, heider2020so}). 
It is suggested in \citep{kirchdoerfer2018data, leygue2018data,he2020physics, nguyen2020variational} to select $\tensor{\mathbb{S}}_{\mathrm{s}}=\tensor{\mathbb{C}}_{\mathrm{s}}^{-1}$ for solid mechanics applications. 
On the other hand, the norm for $\psFluid$  is defined as, 
\begin{equation}
 \norm{\psElemFluid_{n+1}}_{\psFluid}^{2} =  \norm{(\vec{r}_{n+1}, \vec{q}_{n+1})}_{\psFluid}^{2}= \int_{\Omega} \frac{1}{2} \vec{r}_{n+1} \cdot \tensor{C}_{\mathrm{f}} \cdot \vec{r}_{n+1} + \frac{1}{2} \vec{q}_{n+1} \cdot \tensor{S}_{\mathrm{f}} \cdot \vec{q}_{n+1} d\Omega,
\label{eq::norm-fluid}
\end{equation}
where $\tensor{C}_{\mathrm{f}}$ and $\tensor{S}_{\mathrm{f}}$ are \nth{2} order symmetric positive definite tensors. Similarly, these numerical parameters control the importance of pressure gradient and Darcy's velocity vectors in the norm calculations. The unit for $\tensor{C}_{\mathrm{f}}$ is the same as hydraulic conductivity unit ($\frac{\mathrm{Length}^4}{\mathrm{Force} \times \mathrm{Time}}$). This norm, which in our case is of the unit of power, has been introduced in \citet{nguyen2020variational} for Poisson's equation.

Similar to the norm equipped by the solid phase space, the implication of the choice of the
specific weighting effect for $\tensor{S}_{\mathrm{f}}$ and $\tensor{C}_{\mathrm{f}}$ 
has not been previously studied in previous studies.  \citet{nguyen2020variational} suggests $\tensor{S}_{\mathrm{f}} = \tensor{C}_{\mathrm{f}}^{-1}$. The time increment $\Delta t$ used in \eqref{eq::norm-fully} is the scaling factor to make both terms with the same unit as energy. This factor could be directly included in $\tensor{C}_{\mathrm{f}}$ and $\tensor{S}_{\mathrm{f}}$, but we preferred to be consistent with power-like definition of fluid phase space metric.

To find the stationary points for the double minimization defined in Eq. \eqref{eq::fullyDD-prob-def} at time $t=t_{n+1}$, 
a global-local iteration (which is our method of preference and will be described later) is needed to find both the admissible solution $\psElemFully_{n+1}$
and the discrete data points that minimize the distance defined by the norm in Eq. \eqref{eq::norm-fully}. The latter can be done by comparing every point from the material databases to identify the optimized data point from $\dataSetFully_{n+1}$ for each integration point. 
However, searching the optimized data points from the entire data set can be inefficient 
for a large database.  
As such, we 
 consider an adaptive poroelasticity database where a subset of the plausible data points are collected for each time step to constitute a temporal-varying material database constituted
 by the union of an adaptive solid effective-stress-strain database $\dataSetSolid_{i}$ and an adaptive fluid pressure-gradient-Darcy-velocity database $\dataSetFluid_{i}$ 
\begin{equation}
\dataSetFully_{i} = \{ (\tensor{\epsilon}^*, {\tensor{\sigma}^\prime}^*, {\vec{r}}^*, {\vec{q}}^* ) |  (\tensor{\epsilon}^*, {\tensor{\sigma}^\prime}^*)\in \dataSetSolid_{i} ,  ({\vec{r}}^*, {\vec{q}}^* ) \in \dataSetFluid_{i}   \},
\end{equation}
where the subscript $i$ indicates the snapshot taken at a discrete time step $t_{i}$. 
For instance, an admissible subset of data can be identified via prior knowledge (e.g., upper and lower bounds of the porosity-permeability relationship, correlation structures, etc.) and deductive reasoning. In our last numerical experiment, 
we use porosity (the ratio between the void and solid phase volume) to filter out 
the implausible data points.
This treatment enables the data-driven solver to  narrow down the search of possible solutions and therefore 
enhance the efficiency and reduce the memory requirement. 
Furthermore, the trade-off between computational time and memory will be discussed later in Sec. \ref{sec:localminimization}. 
Note that the adaptive database design can also be used as a mean to incorporate an active learning algorithm that generates new data points on demand \citep{lookman2019active, wang2020non}. The active learning approach as well as other algorithms that 
may identify the feasible subset of data through clustering \citep{liu2016self, zhang2019fast} or other techniques are not discussed in this work but will further be explored in the future.

Here, the solid data set $\dataSetSolid_{n+1}$ is the set of experimental strain and effective stress pairs $(\tensor{\epsilon}^*, {\tensor{\sigma}^\prime}^*)$; one point in this data set is a bundle of strain and effective stress together corresponding to one experimental observation. Fluid data set $\dataSetFluid_{n+1}$ is the set of all experimental pressure gradient and Darcy's velocity pairs $({\vec{r}}^*, {\vec{q}}^* )$; one point in this data set is a bundle of pressure gradient and Darcy's velocity together corresponding to one experimental observation.
The union of separate databases for stress-strain and hydraulic responses is designed for practical reasons because experiments that obtain the stress-strain curves and the effective permeability of a specimen are often conducted separately \citep{bardet1997experimental, paterson2005experimental, sun2018prediction}.

\subsubsection{Solution strategy: fixed-point iteration}
Here, we use the fixed-point method to numerically solve the double-minimization statement Eq. \eqref{eq::fullyDD-prob-def} associated with the fully data-driven poroelasticity problem. The use case of this method is initially shown by \citet{kirchdoerfer2016data} for data-driven elasticity problems. 

The data-driven solution must minimize an objective, i.e., the norm defined over poroelasticity phase space Eq. \eqref{eq::norm-fully}, with two different sets of constraints. One set of constraints belongs to a discrete set, $\dataSetFully_{n+1}$, with a finite number of members, but the other one belongs to continuous space, i.e., $\mathcal{C}^{\mathrm{coupled}}_{n+1}$, that satisfies conservation laws. This minimization statement is categorized into combinatorial optimization problems due to the discrete nature of $\dataSetFully_{n+1}$, making the problem NP-hard. The fixed point method, or staggered method, reduces complexities by proposing a sequential solution algorithm, and it has been used in many applications \citep{felippa2001partitioned, borden2012phase, hu2020phase}. Here, it is used to break down the double-minimization into two separate, simpler minimization problems. We solve for one minimization problem by assuming the solution for the other one is fixed. In this way, we iteratively solve a minimization problem and update for another one until convergence of the solution.

\begin{figure}[h]
 \centering
\includegraphics[width=0.5\textwidth]{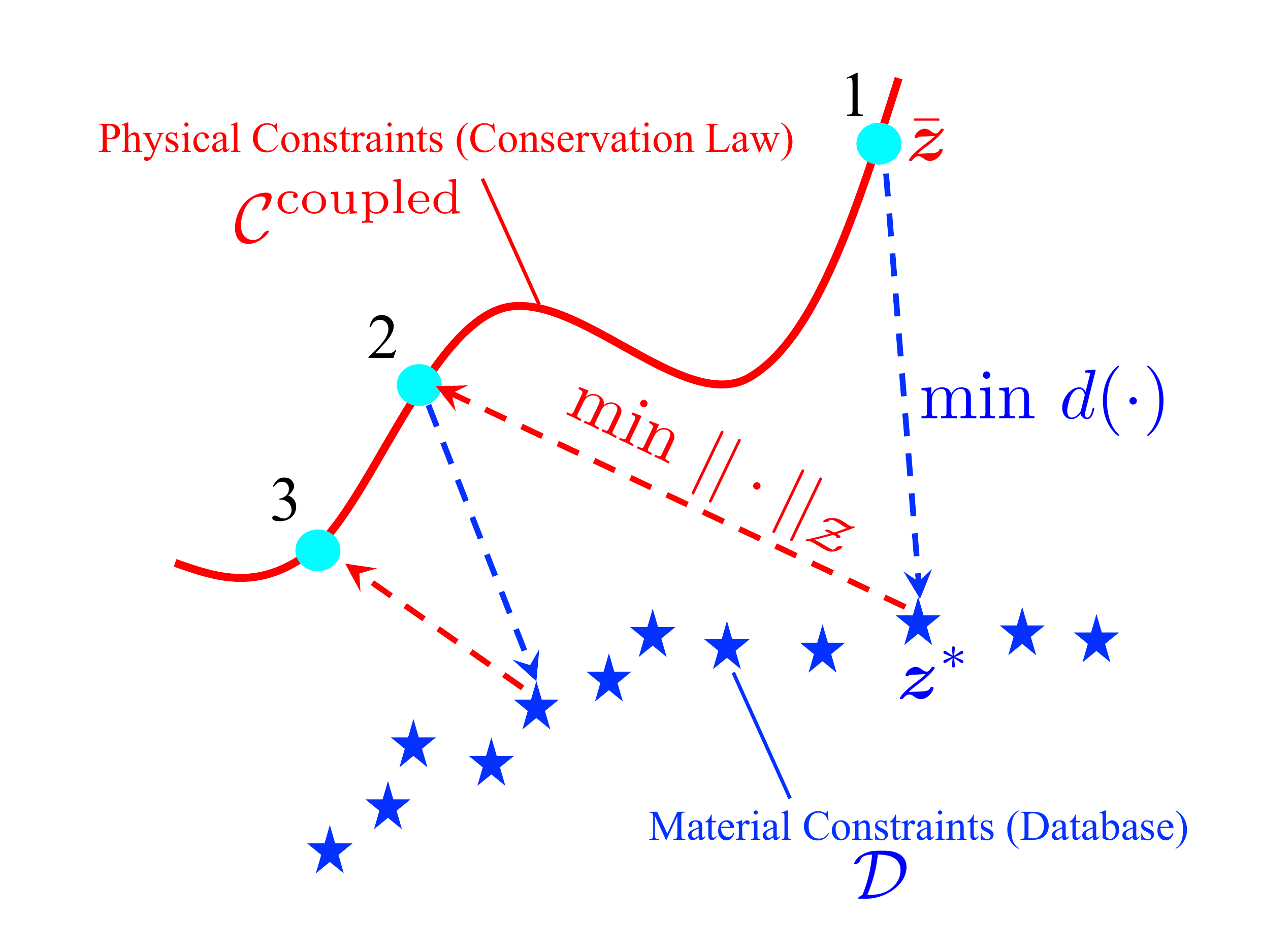}
  \caption{Schematic representation of material $\vec{z}^*$ and physical $\bar{\vec{z}}$ responses at a quadrature point during fixed-point iterations. Numbers show iteration numbers. Star points present the entire database. The solid line describes the continuous admissible solution space that respects conservation laws. Dash blue arrow lines depict the local minimization step from the physical manifold to the material manifold for the quadrature point. Dash red arrow lines depict the global minimization step from the material manifold to the physical manifold. %
  Note that the global step minimizes the norm over the entire domain (for all quadrature points). However, the local step minimizes the distance locally at each quadrature point.
  \label{fig::double_min}}
\end{figure}

According to the fixed-point method, we assume that the optimal data points ${\psElemFully}^*_{n+1}$ are known in the objective function Eq. \eqref{eq::fullyDD-prob-def}, therefore we just need to minimize the objective function for unknowns $\psElemFully_{n+1}$ as follows:
\begin{equation}
\vec{z}^{\mathrm{sf}}_{n+1}  = 
\argmin_{\psElemFully_{n+1}  \in \mathcal{C}^{\mathrm{coupled}}_{n+1}} 
\norm{{\psElemFully}^*_{n+1}  - \psElemFully_{n+1} }_{\psFully}^{2}, \mathrm{given} \ {\psElemFully}^*_{n+1} \in \dataSetFully_{n+1} .
\label{eq::fullyDD-global-min-def}
\end{equation}
We call this minimization step global minimization. The global minimization step, geometrically, project discrete points ${\psElemFully}^*_{n+1}$ onto the continuous space $\mathcal{C}^{\mathrm{coupled}}_{n+1}$ according to the defined norm in Eq. \eqref{eq::norm-fully}; see red dash lines in \fig \ref{fig::double_min}. In other words, this step finds solutions belong to the physical space $\mathcal{C}^{\mathrm{coupled}}_{n+1}$ that are closest points to the selected data points (from material space).

In the next step of fixed-point method, we assume the solutions $\psElemFully_{n+1}$ are known, then we find optimal data points ${\psElemFully}^*_{n+1}$ that minimize the objective function Eq. \eqref{eq::fullyDD-prob-def} as follows:
\begin{equation}
{\psElemFully}^*_{n+1} = 
\argmin_{{\psElemFully}^*_{n+1}   \in \dataSetFully_{n+1}}
\norm{{\psElemFully}^*_{n+1}  - \psElemFully_{n+1} }_{\psFully}^{2},  \mathrm{given} \ \psElemFully_{n+1}  \in \mathcal{C}^{\mathrm{coupled}}_{n+1}.
\label{eq::fullyDD-local-min-def}
\end{equation}
This minimization is defined over the discrete space  $\dataSetFully_{n+1}$. Since choices for ${\psElemFully}^*_{n+1}$ are finite and there is no constraint on data points in the database,
the global objective function Eq. \eqref{eq::fullyDD-local-min-def} defined over $\Omega$ 
is minimized if the integrand is locally minimized. In this work, Gaussian quadrature is used to approximate the spatial integration. Let all the integration points be elements of a finite set 
$\{ \bar{\vec{x}}_{1}, \bar{\vec{x}}_{2}, ..., \bar{\vec{x}}_{n_{\text{int}}} \}$, then the local minimization problem for an integration point $\bar{\vec{x}}_{a}$ reads,  

\begin{equation}
{\psElemFully}^*_{n+1}(\bar{\vec{x}_{a}}) = \argmin_{{\psElemFully}^*_{n+1}   \in \dataSetFully_{n+1}} [ \distFully({\psElemFully}^*_{n+1}, {\psElemFully}_{n+1} (\bar{\vec{x}})) ]^{2},\  \mathrm{given} \ \psElemFully_{n+1}(\bar{\vec{x}_{a}}) \in \mathcal{C}^{\mathrm{coupled}}_{n+1},
\label{eq::local-dist-fullyDD}
\end{equation}
where $\distFully (\cdot)$ is a local distance function between ${\psElemFully}^*_{n+1} $ and $\psElemFully_{n+1} $, which is defined as, 
 Eqs. \eqref{eq::subtract-proelast-space}, \eqref{eq::norm-fully}, \eqref{eq::norm-solid}, and \eqref{eq::norm-fluid}:
\begin{align}
&
\distFully({\psElemFully}^*_{n+1}, \psElemFully_{n+1}) = [ \distSolid({\psElemSolid}^*_{n+1}, \psElemSolid_{n+1})^{2} + \Delta t \distFluid({\psElemFluid}^*_{n+1}, \psElemFluid_{n+1})^{2}]^{1/2}, \label{eq::dist-fully}
\\&
\distSolid({\psElemSolid}^*_{n+1}, \psElemSolid_{n+1})^{2} = 
\frac{1}{2} ( \tensor{\epsilon}^*_{n+1} - \tensor{\epsilon}_{n+1} ) : \tensor{\mathbb{C}}_{\mathrm{s}} : ( \tensor{\epsilon}^*_{n+1} - \tensor{\epsilon}_{n+1} )
+
\frac{1}{2}( {\tensor{\sigma}^\prime}^*_{n+1} - {\tensor{\sigma}^\prime}_{n+1}  ) : \tensor{\mathbb{S}}_{\mathrm{s}} :( {\tensor{\sigma}^\prime}^*_{n+1} - {\tensor{\sigma}^\prime}_{n+1}  ),
\label{eq::dist-solid}
\\&
\distFluid({\psElemFluid}^*_{n+1}, \psElemFluid_{n+1})^{2} = 
\frac{1}{2}( \vec{r}^*_{n+1} - \vec{r}_{n+1} ) \cdot \tensor{C}_{\mathrm{f}} \cdot ( \vec{r}^*_{n+1} - \vec{r}_{n+1} )
+
\frac{1}{2}( \vec{q}^*_{n+1} - \vec{q}_{n+1}) \cdot \tensor{S}_{\mathrm{f}} \cdot ( \vec{q}^*_{n+1} - \vec{q}_{n+1}),
\label{eq::dist-fluid}
\end{align}
where $\distSolid (\cdot)$ and $\distFluid (\cdot)$ denote the local distances between solid-related and fluid-related components, respectively. The technique that solves the local minimization will be discussed in Sec. \ref{sec:localminimization}. Geometrically, the local minimization step project a point in $\mathcal{C}^{\mathrm{coupled}}_{n+1}$ onto the data set $\dataSetFully_{n+1}$, see blue dash lines in \fig \ref{fig::double_min}. 
By minimizing the distance defined in  Eq. \eqref{eq::dist-fully}, we determine points in the data set (material space) that are closest to the conservation laws. 

Each fixed-point iteration consists of two steps \citep{he2020physics}. First, we solve the global minimization Eq. \eqref{eq::fullyDD-local-min-def} to project the solution coming from material space (database) $\dataSetFully_{n+1}$ onto physical space $\mathcal{C}^{\mathrm{coupled}}_{n+1}$; see blue dash arrow lines in \fig \ref{fig::double_min}. Second, we solve local minimization problems to project the most recent solutions belong to the physical space onto the material space; see red dash arrow lines \fig \ref{fig::double_min}. Fixed-point iterations continue until there is no change more than a user-defined tolerance in optimal solutions 
at time step $t_{n+1}$.
In summary, we solve the double-minimization problem in two consecutive steps:\\
\textbf{Global Step:} find the physical field $\vec{z}^{\mathrm{sf}}_{n+1} = (\tensor{\epsilon}, \tensor{\sigma}^\prime , \vec{r} , \vec{q} )_{n+1}$ for a given material field ${\psElemFully}^*_{n+1} \in \dataSetFully_{n+1}$ via:
\begin{align*}
\min_{\psElemFully_{n+1}} 
\norm{{\psElemFully}^*_{n+1}  - \psElemFully_{n+1} }_{\psFully}^{2}, \text{such that} \ \vec{z}^{\mathrm{sf}}_{n+1} \in \mathcal{C}^{\mathrm{coupled}}_{n+1}.
\end{align*}
\textbf{Local Step:} at each local spatial point $\bar{\vec{x}}_{i}$ (quadrature), find the local material point ${\psElemFully}^*_{i,\ n+1} = (\tensor{\epsilon}^*, {\tensor{\sigma}^\prime}^* , \vec{r}^* , \vec{q}^* )_{i,\ n+1}$ for a given physical point ${\psElemFully}_{i,\ n+1} (\bar{\vec{x}}_{i})\in \mathcal{C}^{\mathrm{coupled}}_{n+1}$ via:
\begin{align*}
\argmin_{{\psElemFully}^*_{i,\ n+1}} [ \distFully({\psElemFully}^*_{i,\ n+1}, {\psElemFully}_{i,\ n+1}) ]^{2},\  \text{such that} \ {\psElemFully}^*_{i,\ n+1} \in \dataSetFully_{n+1}.
\end{align*}
We will explain how the global and local steps can be formulated for numerical solutions in Sec. \ref{sec:global-fullyDD} and Sec. \ref{sec:localminimization}, respectively.

\subsubsection{Global minimization}\label{sec:global-fullyDD}
We introduce the functional associated with the global constrained optimization for the fully data-driven poroelasticity problem.
The objective function in Eq. \eqref{eq::fullyDD-local-min-def} is minimized along with the set of constraints defined in $\mathcal{C}^{\mathrm{coupled}}_{n+1}$. As such, the  trial 
spaces $V_u$ and $V_p$ for the u/p poromechanics formulation are chosen to strongly satisfy Dirichlet boundary conditions,
\begin{align}
&
V_u =\left\{  \vec{u}:\Omega \to \mathbb{R}^3  | \vec{u} \in \left[ H^1(\Omega) \right]^3, \vec{u} = \bar{\vec{u}} \ \mathrm{on} \ \partial \Omega_u \right\},
\\&
V_p =\left\{  p:\Omega \to \mathbb{R} | p \in H^1(\Omega), p = \bar{p} \ \mathrm{on} \ \partial \Omega_p \right\}.
\end{align}
The optimal solutions $\vec{u}_{n+1} \in V_u$, $\tensor{\sigma^\prime_{n+1}} \in V_{\sigma^\prime}$, $p_{n+1} \in V_p$, and $\vec{q}_{n+1} \in V_{q}$ are the stationary points of the following functional:
\begin{equation}
\mathcal{L}^{\text{DD}}_{\mathrm{tot}} (\psElemFully_{n+1}, \mathcal{B}_{n+1}; {\psElemFully}^*_{n+1})= 
\mathcal{L}^{\text{DD}}_{\mathrm{loss}} 
+ \mathcal{L}^\text{DD}_{\mathrm{momentum}} 
+ \mathcal{L}^\text{DD}_{\mathrm{mass}},
\label{eq::lagrangian-fullyDD}
\end{equation}
where $\mathcal{L}^\text{DD}_{\mathrm{loss}}$ is the original objective (loss) function, and $\mathcal{L}^\text{DD}_{\mathrm{momentum}}$ and $\mathcal{L}^\text{DD}_{\mathrm{mass}}$ are contributions from constraints defined in $\mathcal{C}^{\mathrm{momentum}}_{n+1}$ and $\mathcal{C}^{\mathrm{mass}}_{n+1}$, respectively. We group all the Lagrange multipliers in
$\mathcal{B}_{n+1} = \{ \vec{\beta}^{\mathrm{mon}}_{n+1}, \vec{\beta}^{\sigma}_{n+1}, \beta^{\mathrm{mass}}_{n+1}, \beta^{q}_{n+1} \}$ where
$\vec{\beta}^{\mathrm{mon}}_{n+1}$ and $\vec{\beta}^{\sigma}_{n+1}$ are real-valued vector fields to weakly enforce the balance of linear momentum and traction boundary conditions, respectively, and 
$\beta^{\mathrm{mass}}_{n+1}$ and $\beta^{q}_{n+1}$ are real-valued scalar fields to weakly enforce conservation of mass, and normal Darcy's velocity boundary conditions. These terms are obtained as follows:
\begin{align}
\begin{split}
	\mathcal{L}^\text{DD}_{\mathrm{loss}}  =
	{}&
			\int_{\Omega} (\distFully({\psElemFully}^*_{n+1},
			 	\psElemFully_{n+1}))^{2} d\Omega,
\end{split}\\
\begin{split}\label{eq::mom-term-lagrangian-fullyDD}
	\mathcal{L}^\text{DD}_{\mathrm{momentum}}   =
	{}&
		 \int_{\Omega} \vec{\beta}^{\mathrm{mom}}_{n+1}  
		 		\cdot \left( \diver{(\tensor{\sigma}^\prime_{n+1}
		 		-Bp_{n+1}\tensor{I})} + \vec{\gamma}_{n+1}  \right)
		 		d\Omega
	\\&
		+\int_{\partial \Omega_{\sigma}} \vec{\beta}^{\sigma}_{n+1} 
		\cdot \left( (\tensor{\sigma}^\prime_{n+1}-Bp_{n+1}
		\tensor{I} \right) \cdot \vec{n} - \bar{\vec{t}}_{n+1}) d\Gamma,
\end{split}\\
	\mathcal{L}^\text{DD}_{\mathrm{mass}}   =
	{}&
		\int_{\Omega} \beta^{\mathrm{mass}}_{n+1} \ 
				(\frac{p_{n+1} }{M }
				+
				{B}  {\epsilon_{\mathrm{vol}}}_{n+1}
				+
				\diver{\vec{q}_{n+1}} \Delta t
				+
				s_{n+1} \Delta t
				-
				\frac{p_{n}}{M } - {B}  {\epsilon_{\mathrm{vol}}}_n)
				 d\Omega
	\\&
		+\int_{\partial \Omega_q} \beta^{q}_{n+1} \ (\vec{q}_{n+1}
				\cdot \vec{n} - \bar{q}_{n+1})  d\Gamma.
	\label{eq::mass-term-lagrangian-fullyDD}
\end{align}

Taking the first variation of Eq. \eqref{eq::lagrangian-fullyDD}, using common rules of the calculus of variations \citep{felippa1994survey,nguyen2020variational}, after applying the divergence theorem leads to
\begin{align*}
\delta \mathcal{L}^\text{DD}_{\text{tot}} =& 
\delta \mathcal{L}^\text{DD}_{\mathrm{loss}} 
+ \delta \mathcal{L}^\text{DD}_{\mathrm{momentum}} 
+ \delta \mathcal{L}^\text{DD}_{\mathrm{mass}} 
\\
=& 
\delta \mathcal{L}^\text{DD}_{u} 
+ \delta \mathcal{L}^\text{DD}_{p}
+ \delta \mathcal{L}^\text{DD}_{\beta^{\text{mom}}}
+ \delta \mathcal{L}^\text{DD}_{\beta^{\text{mass}}}
+ \delta \mathcal{L}^\text{DD}_{\sigma^\prime}
+\delta \mathcal{L}^\text{DD}_{q}
+ \delta \mathcal{L}^\text{DD}_{\beta^{\sigma}}
+ \delta \mathcal{L}^\text{DD}_{\beta^{q}}  = 0,
\end{align*}
where each contribution is as follows:
\begin{align*}
\begin{split}
	\delta \mathcal{L}^\text{DD}_{u}  =
	{}&
			\int_{\Omega} \delta \vec{u}_{n+1} \cdot \frac{\partial
	 				\tensor{\epsilon}(\vec{u}_{n+1})}{\partial \vec{u}} :
	 				\mathbb{C}_{\mathrm{s}} : (\tensor{\epsilon}_{n+1} 
	 				-\tensor{\epsilon}^*_{n+1}) d\Omega 
	 		+\int_{\Omega}  B \delta \vec{u}_{n+1} \cdot \frac{\partial
	 		 		\epsilon_{\mathrm{vol}}(\vec{u}_{n+1})}{\partial \vec{u}}
	 		 		 {\beta}^{\mathrm{mass}}_{n+1} d\Omega,
\end{split}\\
\begin{split}
    \delta \mathcal{L}^\text{DD}_{p}  =
    {}&
    			\int_{\Omega} \grad{\delta p}_{n+1} 
    					\cdot \tensor{C}_{\text{f}} \cdot ( \vec{r}_{n+1} -
    					 \vec{r}^*_{n+1} ) \Delta t d\Omega
		   +\int_{\Omega} \frac{1}{M} \delta p_{n+1}
		   			\beta^{\mathrm{mass}}_{n+1} d\Omega		   			
		   + \int_{\Omega} B \delta p_{n+1} \grad{{\vec{\beta}}
		   			^{\mathrm{mom}}_{n+1}} : \tensor{I} d\Omega
	\\&
		   -\int_{\partial \Omega_{\sigma}} \delta p_{n+1} B \vec{\beta}
		   			^{\sigma}_{n+1}  \cdot \vec{n} d\Gamma
		   -\int_{\partial \Omega} \delta p_{n+1} B \vec{\beta}
		   			^{\text{mom}}_{n+1}  \cdot \vec{n} d\Gamma,
\end{split}\\
\begin{split}
    \delta \mathcal{L}^\text{DD}_{\beta^{\text{mom}}} = 
    {}&
	       -\int_{\Omega} \grad{\delta \vec{\beta}^
	       			{\mathrm{mom}}_{n+1}  } : ( \tensor{\sigma}^
	       			\prime_{n+1} - B p_{n+1} \tensor{I}) d\Omega
		  +\int_{\partial \Omega } \delta \vec{\beta}
		  			^{\mathrm{mom}}_{n+1}  \cdot
		  			( \tensor{\sigma}^\prime_{n+1} - B p_{n+1} \tensor{I})
		  			\cdot \vec{n}
		  			d\Gamma
	\\&
	   	  +\int_{\Omega} \delta \vec{\beta}^{\mathrm{mom}}_{n+1}  
	   	  			\cdot \vec{\gamma}_{n+1} d\Omega,
\end{split}\\
\begin{split}
    \delta \mathcal{L}^\text{DD}_{\beta^{\text{mass}}} = 
    {}&
			\int_{\Omega} \delta {\beta}^{\mathrm{mass}}_{n+1}
					\left[ \frac{1}{M} (p_{n+1} - p_{n}) + B
					({\epsilon_{\mathrm{vol}}}_{n+1} -
					 {\epsilon_{\mathrm{vol}}}_n) + s_{n+1} \Delta t \right]
					 d\Omega
    		   -\int_{\Omega} \grad{\delta {\beta}^{\mathrm{mass}}_{n+1}}
    		    			\cdot \vec{q}_{n+1} \Delta t d\Omega
    \\&
    		  +\int_{\partial \Omega} \delta {\beta
    		  			}^{\mathrm{mass}}_{n+1} \vec{q}_{n+1} \cdot \vec{n}
    		  			\Delta t d\Gamma,
\end{split}\\
\begin{split}
    \delta \mathcal{L}^\text{DD}_{\sigma^\prime} = 
    {}&
			\int_{\Omega} \delta \tensor{\sigma^\prime}_{n+1} :
	 				\mathbb{S}_{\mathrm{s}} : (\tensor{\sigma^\prime}_{n+1} 
	 				-\tensor{\sigma^\prime}^*_{n+1}) d\Omega
	 		- \int_{\Omega} \delta \tensor{\sigma^\prime}_{n+1}:
	 				\grad{\vec{\beta}^{\text{mom}}_{n+1}} d\Omega
	 		+\int_{\partial \Omega_{\sigma}} \delta 
	 				\tensor{\sigma^\prime}_{n+1} : ( \vec{\beta}
	 				^{\sigma}_{n+1} \otimes \vec{n} ) d\Gamma
    \\&
	 		+\int_{\partial \Omega} \delta 
	 				\tensor{\sigma^\prime}_{n+1} : ( \vec{\beta}
	 				^{\text{mom}}_{n+1} \otimes \vec{n} ) d\Gamma,
\end{split}\\
\begin{split}
    \delta \mathcal{L}^\text{DD}_{q} = 
    {}&
    			\int_{\Omega} \delta \vec{q}_{n+1}
    					\cdot \tensor{S}_\text{f} \cdot ( \vec{q}_{n+1} -
    					 \vec{q}^*_{n+1} ) \Delta t d\Omega
    			-\int_{\Omega} \delta \vec{q}_{n+1}
    					\cdot \grad{\beta}^{\text{mass}} \Delta t d\Omega
    			+\int_{\partial \Omega_q} \delta \vec{q}_{n+1}
    					\cdot \beta^q \vec{n} d\Gamma
    \\&
    			+\int_{\partial \Omega} \delta \vec{q}_{n+1}
    					\cdot \beta^{\text{mass}} \vec{n} \Delta t d\Gamma,
\end{split}\\
\begin{split}
     \delta \mathcal{L}^\text{DD}_{\beta^{\sigma}} =
    {}&
		  \int_{\partial \Omega_{\sigma} } \delta \vec{\beta}
		  			^{\sigma}_{n+1}  \cdot
		  			( (\tensor{\sigma}^\prime_{n+1} - B p_{n+1} \tensor{I})
		  			  \cdot \vec{n} - \bar{\vec{t}}_{n+1} )
		  			d\Gamma,
\end{split}\\
     \delta \mathcal{L}^\text{DD}_{\beta^{q}} = 
    {}&
    			\int_{\partial \Omega_q} \delta \beta^q_{n+1}
    					( \vec{q}_{n+1} \cdot \vec{n}  - \bar{q}_{n+1} ) d\Gamma.
\end{align*}
We reduce the number of field variables by setting $\vec{\beta}^{\sigma}_{n+1} = - \vec{\beta}^{\mathrm{mom}}_{n+1}$ defined on the boundary $\partial \Omega_{\sigma}$ and $\beta^{q}_{n+1} = - \Delta t \beta^{\mathrm{mass}}_{n+1}$ defined on the boundary $\partial \Omega_{q}$. After some mathematical manipulations, we obtain the following residuals (corresponding to Euler-Lagrange equations) along with the additional restrictions on fields  $\vec{\beta}^{\text{mom}}_{n+1}$ and  $\beta^{\text{mass}}_{n+1}$ as extra boundary conditions $\vec{\beta}^{\text{mom}}_{n+1} = \vec{0}$ on $\partial \Omega_u$ and $\beta^{\text{mass}}_{n+1} = 0$ on $\partial \Omega_p$:
\begin{align}
\begin{split}\label{eq::res-u-FullyDD} 
	\mathcal{R}^{u}_{n+1}   =
	{}&
			\int_{\Omega} \delta \vec{u}_{n+1} \cdot \frac{\partial
					 \tensor{\epsilon}(\vec{u}_{n+1})}{\partial \vec{u}} :
					  \mathbb{C}_{\mathrm{s}} : (\tensor{\epsilon}_{n+1} -
					  \tensor{\epsilon}^*_{n+1}) d\Omega
 			+\int_{\Omega}  B \delta \vec{u}_{n+1} \cdot \frac{\partial
 			 		  \epsilon_{\mathrm{vol}}(\vec{u}_{n+1})}{\partial \vec{u}}
 			 		   {\beta}^{\mathrm{mass}}_{n+1} d\Omega = 0,
\end{split}\\
\begin{split}\label{eq::res-p-FullyDD}
   \mathcal{R}^p_{n+1}  =
    {}&
			\int_{\Omega} \grad{\delta p}_{n+1} \cdot 
					\tensor{C_{\mathrm{f}}} \cdot ( \vec{r}_{n+1} - 
					\vec{r}^*_{n+1} ) \Delta t d\Omega
			+\int_{\Omega} \frac{1}{M} \delta p_{n+1}
			 		\beta^{\mathrm{mass}}_{n+1} d\Omega
	\\&
			+\int_{\Omega} B \delta p_{n+1} \grad{{\vec{\beta}}
					^{\mathrm{mom}}_{n+1}} : \tensor{I} d\Omega 
	=0,
\end{split}\\
\begin{split}\label{eq::res-betaMom-FullyDD} 
    \mathcal{R}^{\beta^\text{mom}}_{n+1} = 
    {}&
	    		-\int_{\Omega} \grad{\delta \vec{\beta}^
	    				{\mathrm{mom}}_{n+1}  } : ( \tensor{\sigma}^
	    				\prime_{n+1} - B p_{n+1} \tensor{I}) d\Omega
		   +\int_{\partial \Omega_{\tensor{\sigma}} } \delta \vec{\beta}
		   			^{\mathrm{mom}}_{n+1}  \cdot \bar{\vec{t}}_{n+1}
		   			d\Gamma
		   +\int_{\Omega} \delta \vec{\beta}^{\mathrm{mom}}_{n+1}
		    			 \cdot \vec{\gamma}_{n+1} d\Omega = 0,
\end{split}\\
\begin{split}\label{eq::res-betaMass-FullyDD}
    \mathcal{R}^{\beta^\text{mass}}_{n+1}  = 
    {}&
 			\int_{\Omega} \delta {\beta}^{\mathrm{mass}}_{n+1} \left[
 					 \frac{1}{M} (p_{n+1} - p_{n}) + B
 					  ({\epsilon_{\mathrm{vol}}}_{n+1} -
 					  {\epsilon_{\mathrm{vol}}}_n) + s_{n+1} \Delta t \right]
 					  d\Omega
	\\& 
    		   -\int_{\Omega} \grad{\delta {\beta}^{\mathrm{mass}}_{n+1}}
    		   			 \cdot \vec{q}_{n+1} \Delta t d\Omega	    
    		   +\int_{\partial \Omega_{\vec{q}} } \delta {\beta}
    		   			^{\mathrm{mass}}_{n+1} \bar{q}_{n+1} \Delta t d\Gamma
    		   	= 0,
\end{split}\\
\begin{split}\label{eq::res-sigmaPrime-FullyDD}
    \mathcal{R}^{\sigma^\prime}_{n+1} = 
    {}&
			\int_{\Omega} \delta \tensor{\sigma^\prime}_{n+1} :
	 				\left( \mathbb{S}_{\mathrm{s}} :
	 					(\tensor{\sigma^\prime}_{n+1} 
	 					-\tensor{\sigma^\prime}^*_{n+1})
	 				- \grad{\vec{\beta}^{\text{mom}}_{n+1}} \right) d\Omega
	 		= 0, 
\end{split}\\
    \mathcal{R}^{q}_{n+1} = 
    {}&
    			\int_{\Omega} \delta \vec{q}_{n+1}
    					\cdot \left(
 	   						\tensor{S}_\text{f} \cdot ( \vec{q}_{n+1} -
    						 		\vec{q}^*_{n+1} )
    						 	-   \grad{\beta}^{\text{mass}} \right)
    					 \Delta t d\Omega = 0.
	\label{eq::res-q-FullyDD}
\end{align}
We further reduce number of independent fields (and equations) by the local (point-wise) satisfaction of Eqs. \eqref{eq::res-sigmaPrime-FullyDD} and \eqref{eq::res-q-FullyDD} via :
\begin{align}
&
\tensor{\sigma}^\prime_{n+1} =  {\tensor{\sigma}^\prime}^*_{n+1} + \mathbb{S}_{\mathrm{s}}^{-1} : \grad{\vec{\beta}}^{\mathrm{mom}}_{n+1}  \ \mathrm{in} \ \Omega,
\label{eq::eff-stress-dd}
\\&
\vec{q}_{n+1} =\vec{q}^*_{n+1} + \tensor{S}_{\mathrm{f}}^{-1} \cdot \grad{\beta^\text{mass}_{n+1}} \ \mathrm{in} \ \Omega.
\label{eq::darcy-velo-dd}
\end{align}

\remark{The coupled system (Eqs. \eqref{eq::res-u-FullyDD}, \eqref{eq::res-p-FullyDD},  \eqref{eq::res-betaMom-FullyDD}, and \eqref{eq::res-betaMass-FullyDD}) is constant 
even if there is any hidden non-linearity in the database. As such, one may simply store the 
LU factorization (with pivoting) of the tangential matrix at the beginning and 
use the decomposition to facilitate the Gaussian elimination and therefore improve the 
efficiency by avoiding the use of a linear solver at each iteration.}

\subsection{Option 2: Hybrid data-driven poroelasticity 1 (model-based solid + data-driven fluid solver) } \label{sec:hybrid1}
\begin{figure}[h]
 \centering
 \subfigure[]
{\includegraphics[width=0.3\textwidth]{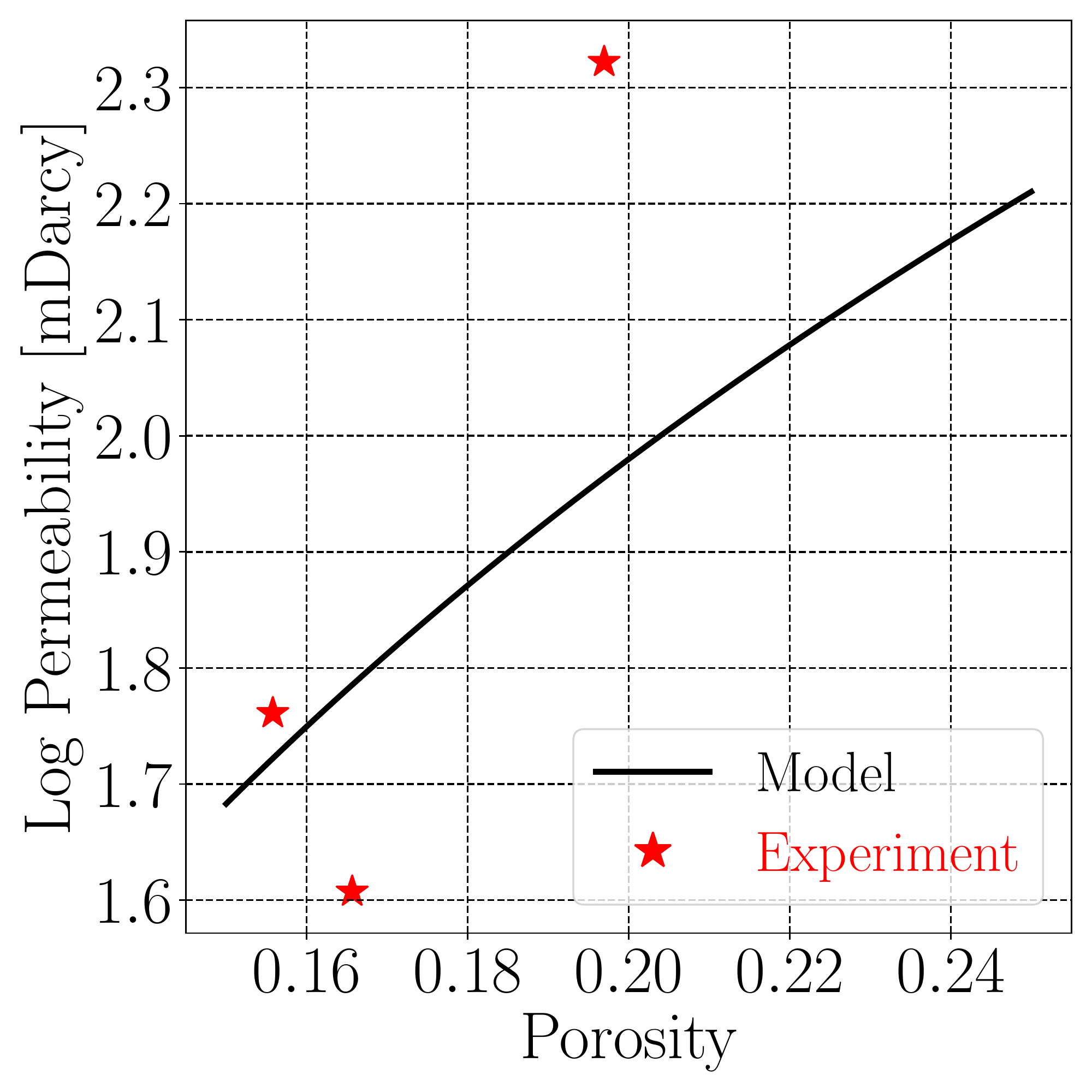}}
\hspace{0.01\textwidth}
 \subfigure[]
{\includegraphics[width=0.3\textwidth]{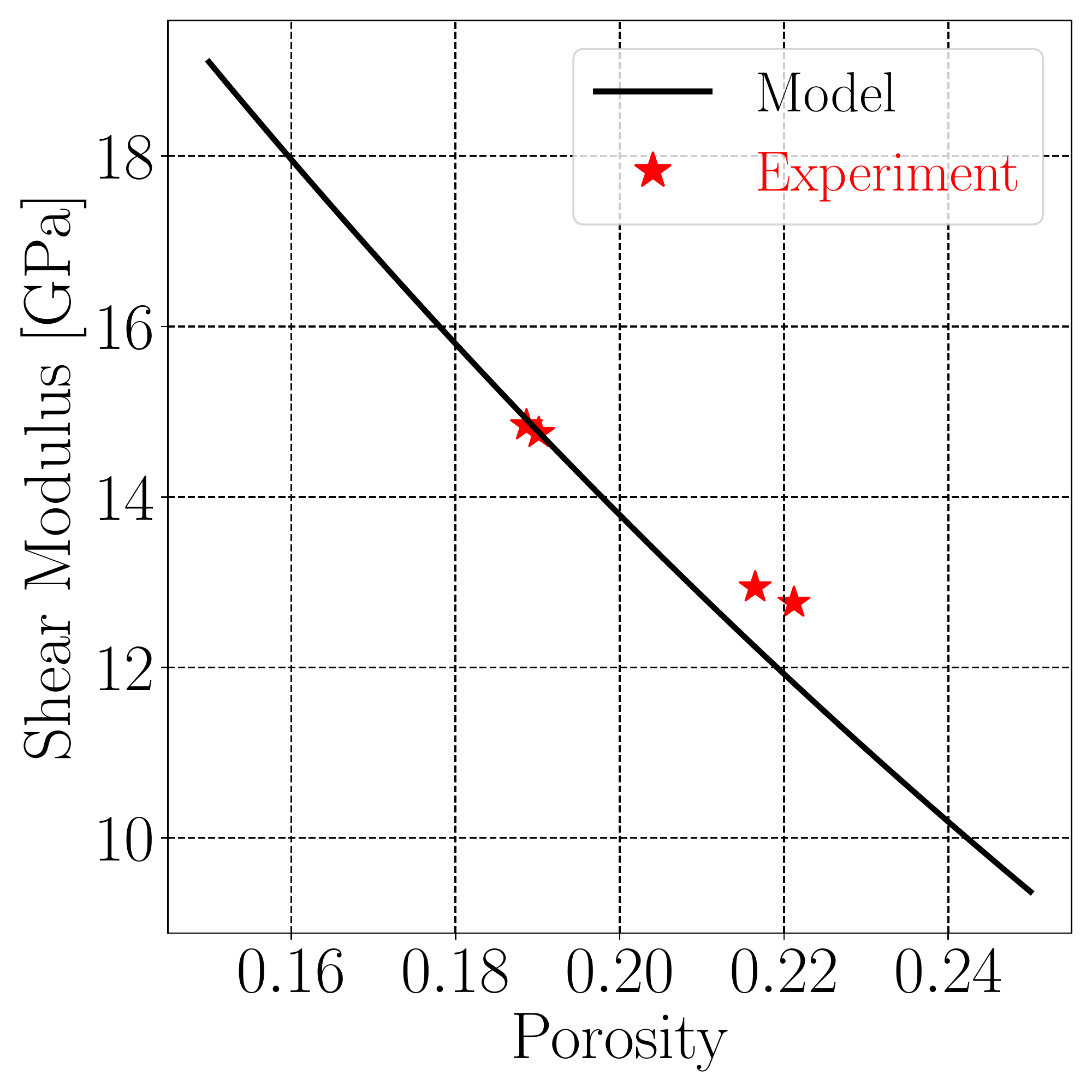}}
\hspace{0.01\textwidth}
 \subfigure[]
{\includegraphics[width=0.3\textwidth]{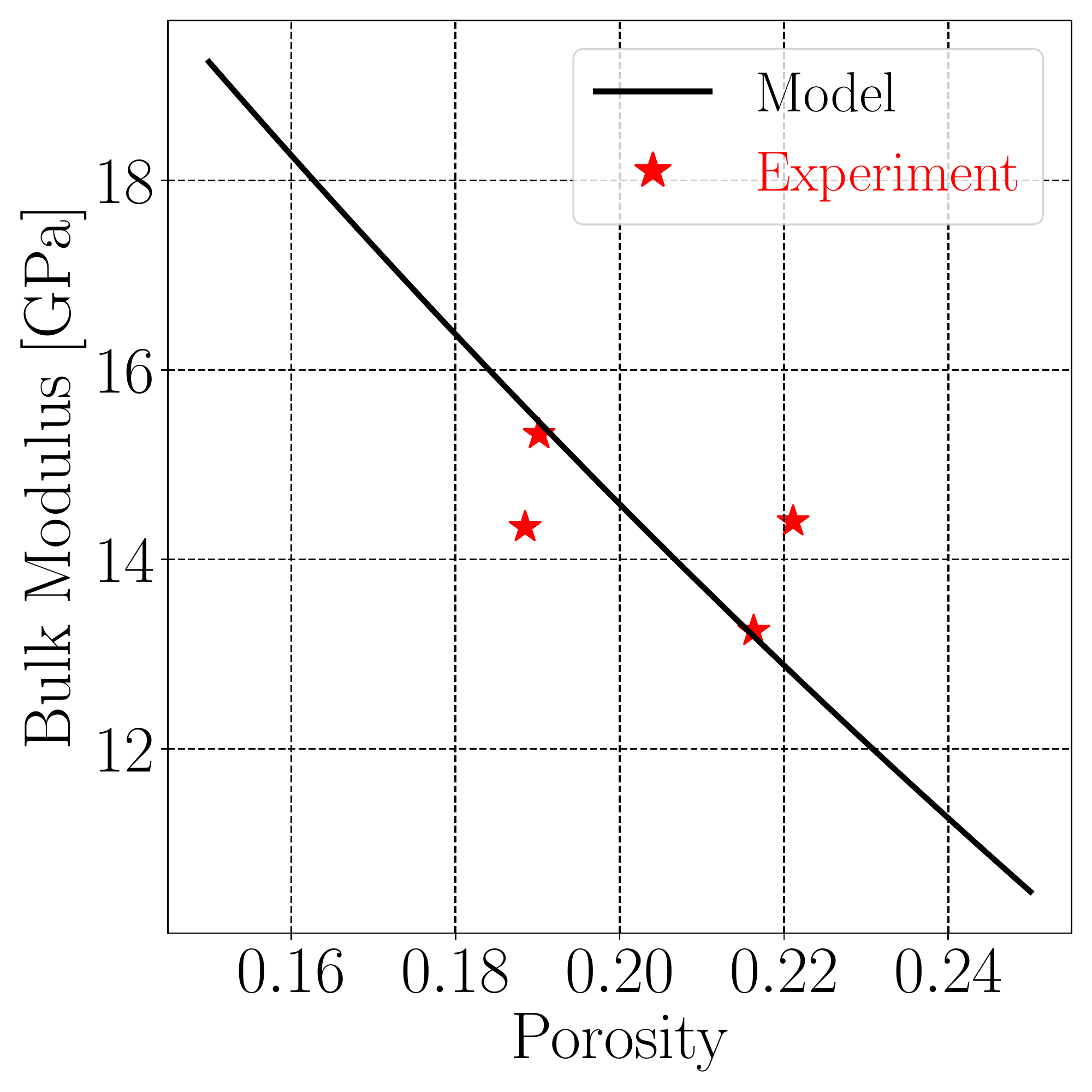}}
\hspace{0.01\textwidth}
  \caption{Black solid lines are model predictions reported in \citet{andra2013digital}. Red star points are experimental observations reported in \citet{andra2013digital} for Berea sandstone. The errors between model predictions and experimental data are considerably less for shear and bulk moduli than permeability. Notice that the permeability is plotted in the Log scale, and so the difference between model and experiments are even greater in real scale. These plots are reproduced from \citet{andra2013digital}, see figures 2(a), 4(a), and 6(a) in the original reference.
   \label{fig::comp-digital-rock-paper}}
\end{figure}

In this section, our goal is to introduce an alternative formulation where the 
fluid constitutive responses are determined from the data-driven approach whereas the solid constitutive responses are determined from a material model. 
This treatment is appropriate for a large variety of poroelasticity problems where 
the confidence interval for any given hydraulic model is expected to be significantly larger than the solid elasticity counterpart after normalization, e.g., see \fig \ref{fig::comp-digital-rock-paper}. Examples of these materials include sandstone, clay, rock, and biological tissues where the estimated effective permeability is often considered accurate if it is within the same order of the benchmark values whereas the elasticity error is expected to be much smaller \citep{paterson2005experimental}.

The idea is that we know an appropriate constitutive law for solid deformation, and there is a database $ \dataSetFluid_{n+1}$ for flow constitutive behavior which is a set of finite pairs of pressure gradient and Darcy's velocity. The following derivation is not restricted to a specific solid constitutive law. The only assumption is that the constitutive law is derived from an energy potential in the context of hyperelasticity for the small deformation limit.

Since the data-driven part is only accounted for the hydraulic constitutive law, the defined norm for the poroelasticity phase space Eq. \eqref{eq::norm-fully} includes only the fluid contribution. We incorporate solid constitutive model as additional constraint in the set $\mathcal{C}^{\mathrm{momentum}}_{n+1}$ defined in the fully data-driven formulation, and we designate the new set by $\bar{\mathcal{C}}^{\mathrm{momentum}}_{n+1}$ to distinguish them. The resultant problem statement for this hybrid option reads,
\begin{equation}
\bar{\vec{z}}^{\mathrm{sf}}_{n+1}  = \underset{\bar{\mathcal{C}}^{\mathrm{coupled}}_{n+1}}{\arg} \left\{  
\min_{{\psElemFluid}^*_{n+1}  \in \dataSetFluid_{n+1}}
\min_{\psElemFully_{n+1}  \in \bar{\mathcal{C}}^{\mathrm{coupled}}_{n+1}} 
\norm{{\psElemFluid}^*_{n+1}  - \psElemFluid_{n+1} }_{\psFluid}^{2}
\right\},
\label{eq::fluidDD-prob-def}
\end{equation}
where $\bar{\mathcal{C}}^{\mathrm{coupled}}_{n+1}=\bar{\mathcal{C}}^{\mathrm{momentum}}_{n+1} \cap \mathcal{C}^{\mathrm{mass}}_{n+1}$.
Recall that ${\psElemFluid}^*_{n+1}$ encodes the fluid-related variables of $\psElemFully_{n+1}$, i.e., pressure gradient and Darcy's velocity. The norm $\norm{\cdot}_{\psFluid}$ defined over the fluid phase space $\psFluid$ is the same as Eq. \eqref{eq::norm-fluid}. Note that the unit of the objective function in Eq. \eqref{eq::fluidDD-prob-def} is power, but it is energy for the fully data-driven Eq. \eqref{eq::fullyDD-prob-def}. Following the same procedure described in the fully data-driven formulation, we solve the above double-minimization by the fixed-point method consisting global and local steps. 

For the global minimization step we have:
\begin{equation}
\mathcal{L}^\text{HYB1}_{\mathrm{tot}} (\psElemFully_{n+1}, \mathcal{B}_{n+1}; {\psElemFluid}^*_{n+1})= 
\mathcal{L}^\text{HYB1}_{\mathrm{loss}} 
+ \mathcal{L}^\text{HYB1}_{\mathrm{momentum}} 
+ \mathcal{L}^\text{HYB1}_{\mathrm{mass}},
\label{eq::lagrangian-fluidDD}
\end{equation}
where $\mathcal{L}^\text{HYB1}_{\mathrm{momentum}}$ is almost the same as  $\mathcal{L}^{\text{DD}}_{\mathrm{momentum}}$ defined in Eq. \eqref{eq::mom-term-lagrangian-fullyDD} with the only difference that the effective stress term is replaced by the constitutive relation; there exits a potential $\psi(\tensor{\epsilon})$ such that $\tensor{\sigma^\prime} (\tensor{\epsilon}) = \frac{\partial \psi}{\partial \tensor{\epsilon}}$. In other words, the constitutive relation is imposed strongly (point-wise) herein. Notice that if there is a need to define an effective stress field as an independent field, similar to mixed formulations for elasticity \citep{washizu1975variational}, one could weakly impose the constitutive relation by adding its contribution through a tensorial Lagrange multiplier defined over the whole domain. Here, we do not intend to arrive at formulations with strain or stress fields as independent fields. Because such formulations increase the number of unknowns significantly, also they require solution spaces with higher regularity such as Hilbert space $H(\textrm{div}, \Omega)$ \citep{arnold1988new, korsawe2006finite, teichtmeister2019aspects, fahrendorf2020mixed}; this regularity is needed to fulfill the continuity condition of normal traction between elements while tangential traction can be discontinuous.
The term $\mathcal{L}^\text{HYB1}_{\mathrm{mass}}$ is exactly the same as $ \mathcal{L}^\text{DD}_{\mathrm{mass}}$ defined in Eq. \eqref{eq::mass-term-lagrangian-fullyDD} since nothing related to the mass balance is changed.
According to the norm defined in Eq. \eqref{eq::norm-fluid}, the original objective (loss) function is as follows:
\begin{equation}
\mathcal{L}^\text{HYB1}_{\mathrm{loss}} = \int_{\Omega} (\distFluid({\psElemFluid}^*_{n+1}, \psElemFluid_{n+1}))^{2} d\Omega,
\end{equation}
where the fluid distance function $\distFluid(\cdot)$ is defined in Eq. \eqref{eq::dist-fluid}.
In this hybrid formulation,  the effective stress $\tensor{\sigma}^\prime_{n+1}$ is obtained from 
a constitutive model, such as an hyperelastic strain-energy functional $\psi(\tensor{\epsilon}(\vec{u}))$, a key departure from the fully data-driven formulation in 
 Eq. \eqref{eq::lagrangian-fullyDD}. The corresponding first variation of Eq. \eqref{eq::lagrangian-fluidDD} reads, 
\begin{align}
\delta \mathcal{L}^\text{HYB1}_{\text{tot}} =& 
\delta \mathcal{L}^\text{HYB1}_{\mathrm{loss}} 
+ \delta \mathcal{L}^\text{HYB1}_{\mathrm{momentum}} 
+ \delta \mathcal{L}^\text{HYB1}_{\mathrm{mass}} 
\nonumber \\
=& 
\delta \mathcal{L}^\text{HYB1}_{u} 
+ \delta \mathcal{L}^\text{HYB1}_{p}
+ \delta \mathcal{L}^\text{HYB1}_{\beta^{\text{mom}}}
+ \delta \mathcal{L}^\text{HYB1}_{\beta^{\text{mass}}}
+\delta \mathcal{L}^\text{HYB1}_{q}
+ \delta \mathcal{L}^\text{HYB1}_{\beta^{\sigma}}
+ \delta \mathcal{L}^\text{HYB1}_{\beta^{q}}  = 0,
\end{align}
where:
\begin{align*}
\begin{split}
	\delta \mathcal{L}^\text{HYB1}_{u}  =
	{}&
	 		\int_{\Omega}  B \delta \vec{u}_{n+1} \cdot \frac{\partial
	 		 		\epsilon_{\mathrm{vol}}(\vec{u}_{n+1})}{\partial \vec{u}}
	 		 		 {\beta}^{\mathrm{mass}}_{n+1} d\Omega
			-\int_{\Omega} \delta \vec{u}_{n+1} \cdot 
					\grad{\beta}^\text{mom} : 
					\frac{\partial \tensor{\sigma}^
							\prime(\tensor{\epsilon}_{n+1})}
							{\partial \tensor{\epsilon}} :
					\frac{\partial \tensor{\epsilon}(\vec{u}_{n+1})}
							{\partial \vec{u}}
	 				d\Omega 
	\\&
			+\int_{\partial \Omega} \delta \vec{u}_{n+1} \cdot  
					\frac{\partial \tensor{\sigma}^
							\prime(\tensor{\epsilon}_{n+1})}
							{\partial \tensor{\epsilon}} :
					\frac{\partial \tensor{\epsilon}(\vec{u}_{n+1})}
							{\partial \vec{u}} :
					(  \vec{\beta}^\text{mom} \otimes \vec{n}  )
	 				d\Gamma
	\\&
			-\int_{\partial \Omega_{\sigma}} \delta \vec{u}_{n+1} \cdot 
					\frac{\partial \tensor{\sigma}^
							\prime(\tensor{\epsilon}_{n+1})}
							{\partial \tensor{\epsilon}} :
					\frac{\partial \tensor{\epsilon}(\vec{u}_{n+1})}
							{\partial \vec{u}} :
					(  \vec{\beta}^\sigma \otimes \vec{n}  )
	 				d\Gamma,
\end{split}\\
\begin{split}
    \delta \mathcal{L}^\text{HYB1}_{p}  =
    {}&
    			\int_{\Omega} \grad{\delta p}_{n+1} 
    					\cdot \tensor{C}_{\text{f}} \cdot ( \vec{r}_{n+1} -
    					 \vec{r}^*_{n+1} ) d\Omega
		   +\int_{\Omega} \frac{1}{M} \delta p_{n+1}
		   			\beta^{\mathrm{mass}}_{n+1} d\Omega		   			
		   + \int_{\Omega} B \delta p_{n+1} \grad{{\vec{\beta}}
		   			^{\mathrm{mom}}_{n+1}} : \tensor{I} d\Omega
	\\&
		   -\int_{\partial \Omega_{\sigma}} \delta p_{n+1} B \vec{\beta}
		   			^{\sigma}_{n+1}  \cdot \vec{n} d\Gamma
		   -\int_{\partial \Omega} \delta p_{n+1} B \vec{\beta}
		   			^{\text{mom}}_{n+1}  \cdot \vec{n} d\Gamma,
\end{split}\\
    \delta \mathcal{L}^\text{HYB1}_{q} = 
    {}&
    			\int_{\Omega} \delta \vec{q}_{n+1}
    					\cdot \tensor{S}_\text{f} \cdot ( \vec{q}_{n+1} -
    					 \vec{q}^*_{n+1} ) d\Omega
    			-\int_{\Omega} \delta \vec{q}_{n+1}
    					\cdot \grad{\beta}^{\text{mass}} \Delta t d\Omega
    			+\int_{\partial \Omega_q} \delta \vec{q}_{n+1}
    					\cdot \beta^q \vec{n} d\Gamma
    \\&
    			+\int_{\partial \Omega} \delta \vec{q}_{n+1}
    					\cdot \beta^{\text{mass}} \vec{n} \Delta t d\Gamma.
\end{align*}
The remaining terms are identical to the fully data-driven counterparts, i.e., 
$\delta \mathcal{L}^\text{HYB1}_{\beta^{\text{mom}}} = 
\delta \mathcal{L}^\text{DD}_{\beta^{\text{mom}}}$, 
$\delta \mathcal{L}^\text{HYB1}_{\beta^{\text{mass}}}=
\delta \mathcal{L}^\text{DD}_{\beta^{\text{mass}}}$, 
$\delta \mathcal{L}^\text{HYB1}_{\beta^{\sigma}} =
\delta \mathcal{L}^\text{DD}_{\beta^{\sigma}}
$, and
$\delta \mathcal{L}^\text{HYB1}_{\beta^{q}} =
\delta \mathcal{L}^\text{DD}_{\beta^{q}}$. For brevity,
 we refer to those terms defined in the fully data-driven section.

Similar to the fully data driven formulation, we first reduce the number of unknown fields by setting $\vec{\beta}^{\sigma}_{n+1} = - \vec{\beta}^{\mathrm{mom}}_{n+1}$ defined on boundary $\partial \Omega_{\sigma}$ and $\beta^{q}_{n+1} = -\Delta t \beta^{\mathrm{mass}}_{n+1}$ defined on boundary $\partial \Omega_{q}$. After some mathematical manipulations, we obtain the following residuals (corresponding to the Euler-Lagrange equations of Eq. \eqref{eq::lagrangian-fluidDD}) along with the additional restrictions on fields  $\vec{\beta}^{\text{mom}}_{n+1}$ and  $\beta^{\text{mass}}_{n+1}$ as extra boundary conditions $\vec{\beta}^{\text{mom}}_{n+1} = \vec{0}$ on $\partial \Omega_u$ and $\beta^{\text{mass}}_{n+1} = 0$ on $\partial \Omega_p$:
\begin{align}
\begin{split}\label{eq::res-u-FluidDD} 
	\mathcal{R}^{u}_{n+1}   =
	{}&
 			\int_{\Omega}  B \delta \vec{u}_{n+1} \cdot \frac{\partial
 			 		  \epsilon_{\mathrm{vol}}(\vec{u}_{n+1})}{\partial \vec{u}}
 			 		   {\beta}^{\mathrm{mass}}_{n+1} d\Omega 
			-\int_{\Omega} \delta \vec{u}_{n+1} \cdot 
					\grad{\beta}^\text{mom} : 
					\frac{\partial \tensor{\sigma}^
							\prime(\tensor{\epsilon}_{n+1})}
							{\partial \tensor{\epsilon}} :
					\frac{\partial \tensor{\epsilon}(\vec{u}_{n+1})}
							{\partial \vec{u}}
	 				d\Omega = 0,
\end{split}\\
\begin{split}\label{eq::res-p-FluidDD}
   \mathcal{R}^p_{n+1}  =
    {}&
			\int_{\Omega} \grad{\delta p}_{n+1} \cdot 
					\tensor{C_{\mathrm{f}}} \cdot ( \vec{r}_{n+1} - 
					\vec{r}^*_{n+1} ) \Delta t d\Omega
			+\int_{\Omega} \frac{1}{M} \delta p_{n+1}
			 		\beta^{\mathrm{mass}}_{n+1} d\Omega
	\\&
			+\int_{\Omega} B \delta p_{n+1} \grad{{\vec{\beta}}
					^{\mathrm{mom}}_{n+1}} : \tensor{I} d\Omega 
	=0,
\end{split}\\
\begin{split}\label{eq::res-betaMom-FluidDD} 
    \mathcal{R}^{\beta^\text{mom}}_{n+1} = 
    {}&
	    		-\int_{\Omega} \grad{\delta \vec{\beta}^
	    				{\mathrm{mom}}_{n+1}  } : ( 
	    						\frac{\partial \psi(\tensor{\epsilon}_{n+1})}
	    						{\partial \tensor{\epsilon}} - B p_{n+1} \tensor{I})
	    			d\Omega
		   +\int_{\partial \Omega_{\tensor{\sigma}} } \delta \vec{\beta}
		   			^{\mathrm{mom}}_{n+1}  \cdot \bar{\vec{t}}_{n+1}
		   			d\Gamma
	\\&
		   +\int_{\Omega} \delta \vec{\beta}^{\mathrm{mom}}_{n+1}
		    			 \cdot \vec{\gamma}_{n+1} d\Omega = 0,
\end{split}\\
\begin{split}\label{eq::res-betaMass-FluidDD}
    \mathcal{R}^{\beta^\text{mass}}_{n+1}  = 
    {}&
 			\int_{\Omega} \delta {\beta}^{\mathrm{mass}}_{n+1} \left[
 					 \frac{1}{M} (p_{n+1} - p_{n}) + B
 					  ({\epsilon_{\mathrm{vol}}}_{n+1} -
 					  {\epsilon_{\mathrm{vol}}}_n) + s_{n+1} \Delta t \right]
 					  d\Omega
	\\& 
    		   -\int_{\Omega} \grad{\delta {\beta}^{\mathrm{mass}}_{n+1}}
    		   			 \cdot \vec{q}_{n+1} \Delta t d\Omega	    
    		   +\int_{\partial \Omega_{\vec{q}} } \delta {\beta}
    		   			^{\mathrm{mass}}_{n+1} \bar{q}_{n+1} \Delta t d\Gamma
    		   	= 0,
\end{split}\\
    \mathcal{R}^{q}_{n+1} = 
    {}&
    			\int_{\Omega} \delta \vec{q}_{n+1}
    					\cdot \left(
 	   						\tensor{S}_\text{f} \cdot ( \vec{q}_{n+1} -
    						 		\vec{q}^*_{n+1} )
    						 	-   \Delta t \grad{\beta}^{\text{mass}} \right)
    					 d\Omega = 0.
	\label{eq::res-q-FluidDD}
\end{align}
We further reduce number of independent fields (and equations) by the local (point-wise) satisfaction of Eq. \eqref{eq::res-q-FluidDD} via:
\begin{equation}
\vec{q}_{n+1} =\vec{q}^*_{n+1} + \Delta t \tensor{S}_{\mathrm{f}}^{-1} \cdot \grad{\beta^\text{mass}_{n+1}} \ \mathrm{in} \ \Omega.
\label{eq::darcy-velo-fluidDD}
\end{equation}

As mentioned in the previous section, the next step in the fixed-point method is to solve the local minimization problem. According to Eq. \eqref{eq::fluidDD-prob-def}, the local minimization for this hybrid scheme reads as follow,
\begin{equation}
{\psElemFluid}^*_{n+1}(\bar{\vec{x}}_{a}) = \argmin_{{\psElemFluid}^*_{n+1}   \in \dataSetFluid_{n+1}} (\distFluid({\psElemFluid}^*_{n+1}, \psElemFluid_{n+1} (\bar{\vec{x}}_{a}))^{2},\  \mathrm{given} \ \psElemFluid_{n+1}(\bar{\vec{x}_{a}})  \in \bar{\mathcal{C}}^{\mathrm{coupled}}_{n+1}, 
\end{equation}
where $\bar{\vec{x}}_{a} \in \Omega^h$ is an integration point in $\Omega^h$.
 We will discuss how to solve the local sub-problem defined at each integration point in Sec. \ref{sec:localminimization}.

Notice that in the data-driven formulation there are three types of computational costs: data availability (gathering data), global optimization (solving a system of equations), local optimization (searching inside a database). 
The local optimization part is an NP-hard problem, and its computational cost grows exponentially by increasing the database size. Therefore, if the number of unknowns (degree of freedoms) is not so high while the database required for data-driven schemes is large, local optimization is the dominant source of computational cost.
As a result, the hybrid formulation could be more efficient than the fully data-driven counterpart
 if the solid behavior can be accurately captured by a constitutive law with an acceptable standard deviation from the ground-truth. 
We will discuss the computational issues concerning the local optimization step in Sec. \ref{sec:localminimization}.

\subsection{Option 3: Hybrid data-driven poroelasticity 2 (data-driven solid + mode-based fluid solver)} \label{sec:hybrid2}

For comparison purposes,  we present another hybrid formulation that includes Darcy's law as the fluid constitutive model, whereas the solid deformation is predicted directly from a collection of data points. 
We start with the fully data-driven problem defined in Eq. \eqref{eq::fullyDD-prob-def} and adjust it to be compatible with this hybrid formulation that considers just solid part as data-driven. To this end, we add Darcy's equation as an additional constraint in the set $\mathcal{C}^{\mathrm{mass}}_{n+1}$ defined for the fully data-driven formulation. Darcy's law reads as follows:
\begin{equation}
\vec{q}_{n+1} = -  \frac{1}{\mu^{\mathrm{f}}}  \tensor{k} (\grad{p}_{n+1} + \vec{\gamma}^{\mathrm{f}}_{n+1} ),
\label{eq::darcy-equation}
\end{equation}
where $\mu^{\mathrm{f}}$ and $\tensor{k}$ are dynamic viscosity of pore fluid and intrinsic permeability tensor with unit $[\mathrm{Length}^2]$, respectively. Besides, the contribution of the fluid-related term in the norm defined in Eq. \eqref{eq::norm-fully} should be excluded since only solid constitutive behavior is data-driven. Therefore, the problem definition in this hybrid formulation is as follows:
\begin{equation}
\bar{\vec{z}}^{\mathrm{sf}}_{n+1}  = \underset{\hat{\mathcal{C}}^{\mathrm{coupled}}_{n+1}}{\arg} \left\{  
\min_{{\psElemSolid}^*_{n+1}  \in \dataSetSolid_{n+1}}
\min_{\psElemFully_{n+1}  \in \hat{\mathcal{C}}^{\mathrm{coupled}}_{n+1}} 
\norm{{\psElemSolid}^*_{n+1}  - \psElemSolid_{n+1} }_{\psSolid}^{2}
\right\}, 
\label{eq::solidDD-prob-def}
\end{equation}
where $\hat{\mathcal{C}}^{\mathrm{coupled}}_{n+1}=\mathcal{C}^{\mathrm{momentum}}_{n+1} \cap \hat{\mathcal{C}}^{\mathrm{mass}}_{n+1}$ and $\hat{\mathcal{C}}^{\mathrm{mass}}_{n+1}$ is almost the same as $\mathcal{C}^{\mathrm{mass}}_{n+1}$ defined for fully data-driven
formulation except it has an additional constraint from the hydraulic constitutive law (Darcy's law). Recall that ${\psElemSolid}^*_{n+1}$ encodes the solid-related variables of $\psElemFully_{n+1}$, i.e., strain and effective stress. The norm $\norm{\cdot}_{\psSolid}$ defined over the solid phase space $\psSolid$ is the same as Eq. \eqref{eq::norm-solid}. Similarly, we solve the above double-minimization by the fixed-point method consisting global and local steps. 

According to Eq. \eqref{eq::solidDD-prob-def}, the total objective function in the global minimization reads as:
\begin{equation}
\mathcal{L}^\text{HYB2}_{\mathrm{tot}} (\psElemFully_{n+1}, \mathcal{B}_{n+1}; {\psElemSolid}^*_{n+1})= 
\mathcal{L}^\text{HYB2}_{\mathrm{loss}} 
+ \mathcal{L}^\text{HYB2}_{\mathrm{momentum}}
+ \mathcal{L}^\text{HYB2}_{\mathrm{mass}},
\label{eq::lagrangian-solidDD}
\end{equation}
where $\mathcal{L}^\text{HYB2}_{\mathrm{mass}}$ is almost the same as  $\mathcal{L}^\text{DD}_{\mathrm{mass}}$ defined in Eq. \eqref{eq::mass-term-lagrangian-fullyDD} with the only difference that the Darcy's velocity term is replaced by Eq. \eqref{eq::darcy-equation}. The term $\mathcal{L}^\text{HYB2}_{\mathrm{momentum}}$ is exactly the same as $ \mathcal{L}^\text{DD}_{\mathrm{momentum}}$ defined in Eq. \eqref{eq::mom-term-lagrangian-fullyDD} since nothing related to the balance of linear-momentum is changed. The original objective function $\mathcal{L}^\text{HYB2}_{\mathrm{loss}}$ in this hybrid formulation is the norm defined over solid phase space Eq. \eqref{eq::norm-solid} as follows:
\begin{equation}
\mathcal{L}^\text{HYB2}_{\mathrm{loss}} = \int_{\Omega} (\distSolid({\psElemSolid}^*_{n+1}, \psElemSolid_{n+1}) )^{2} \; d\Omega,
\end{equation}
where solid distance function $\distSolid(\cdot)$ is defined in Eq. \eqref{eq::dist-solid}.

Recall that, here, the Darcy velocity $\vec{q}_{n+1}$ is not an independent field in Eq. \eqref{eq::lagrangian-solidDD} which is not the case in Eq. \eqref{eq::lagrangian-fullyDD}. It relates to the pressure gradient field via the Darcy's law. Taking the first variation of Eq. \eqref{eq::lagrangian-solidDD} after applying the divergence theorem leads to:
\begin{align*}
\delta \mathcal{L}^\text{HYB2}_{\text{tot}} =& 
\delta \mathcal{L}^\text{HYB2}_{\mathrm{loss}} 
+ \delta \mathcal{L}^\text{HYB2}_{\mathrm{momentum}} 
+ \delta \mathcal{L}^\text{HYB2}_{\mathrm{mass}} 
\\
=& 
\delta \mathcal{L}^\text{HYB2}_{u} 
+ \delta \mathcal{L}^\text{HYB2}_{p}
+ \delta \mathcal{L}^\text{HYB2}_{\beta^{\text{mom}}}
+ \delta \mathcal{L}^\text{HYB2}_{\beta^{\text{mass}}}
+\delta \mathcal{L}^\text{HYB2}_{\sigma^\prime}
+ \delta \mathcal{L}^\text{HYB2}_{\beta^{\sigma}}
+ \delta \mathcal{L}^\text{HYB2}_{\beta^{q}}  = 0,
\end{align*}
where:
\begin{align*}
    \delta \mathcal{L}^\text{HYB2}_{p} = 
    {}&
    			\int_{\Omega} 
    			 \grad{\delta p} \cdot \tensor{k} \cdot \grad{\beta}
    			 		^{\text{mass}} \frac{\Delta t}{\mu^{\text{f}}} d\Omega
    			-\int_{\partial \Omega} 
    			 \grad{\delta p} \cdot \tensor{k} \cdot \beta
    			 		^{\text{mass}} \vec{n} 
    			 		\frac{\Delta t}{\mu^{\text{f}}} d\Omega
    			-\int_{\partial \Omega_q} 
    			 \grad{\delta p} \cdot \tensor{k} \cdot \beta^q \vec{n} 
    			 		\frac{1}{\mu^{\text{f}}} d\Omega
	\\&
		   +\int_{\Omega} \frac{1}{M} \delta p_{n+1}
		   			\beta^{\mathrm{mass}}_{n+1} d\Omega
		   + \int_{\Omega} B \delta p_{n+1} \grad{{\vec{\beta}}
		   			^{\mathrm{mom}}_{n+1}} : \tensor{I} d\Omega
		   -\int_{\partial \Omega_{\sigma}} \delta p_{n+1} B \vec{\beta}
		   			^{\sigma}_{n+1}  \cdot \vec{n} d\Gamma
	\\&
		   -\int_{\partial \Omega} \delta p_{n+1} B \vec{\beta}
		   			^{\text{mom}}_{n+1}  \cdot \vec{n} d\Gamma,
\end{align*}
and other terms remain similar to the fully data-driven formulation, i.e., 
$\delta \mathcal{L}^\text{HYB2}_{u} = 
\delta \mathcal{L}^\text{DD}_{u}$, 
$\delta \mathcal{L}^\text{HYB2}_{\beta^{\text{mom}}} = 
\delta \mathcal{L}^\text{DD}_{\beta^{\text{mom}}}$, 
$\delta \mathcal{L}^\text{HYB2}_{\beta^{\text{mass}}}=
\delta \mathcal{L}^\text{DD}_{\beta^{\text{mass}}}$,
$\delta \mathcal{L}^\text{HYB2}_{\sigma^\prime}=
\delta \mathcal{L}^\text{DD}_{\sigma^\prime}$ 
$\delta \mathcal{L}^\text{HYB2}_{\beta^{\sigma}} =
\delta \mathcal{L}^\text{DD}_{\beta^{\sigma}}$, and
$\delta \mathcal{L}^\text{HYB2}_{\beta^{q}} =
\delta \mathcal{L}^\text{DD}_{\beta^{q}}$.

Similar to the fully data driven formulation, we first reduce the number of unknown fields by setting $\vec{\beta}^{\sigma}_{n+1} = - \vec{\beta}^{\mathrm{mom}}_{n+1}$ defined on boundary $\partial \Omega_{\sigma}$ and $\beta^{q}_{n+1} = -\Delta t \beta^{\mathrm{mass}}_{n+1}$ defined on boundary $\partial \Omega_{q}$. After some mathematical manipulations, we obtain the following residuals (corresponding to Euler-Lagrange equations) along with the additional restrictions on fields  $\vec{\beta}^{\text{mom}}_{n+1}$ and  $\beta^{\text{mass}}_{n+1}$ as extra boundary conditions $\vec{\beta}^{\text{mom}}_{n+1} = \vec{0}$ on $\partial \Omega_u$ and $\beta^{\text{mass}}_{n+1} = 0$ on $\partial \Omega_p$:
\begin{align}
\begin{split}\label{eq::res-u-solidDD} 
	\mathcal{R}^{u}_{n+1}   =
	{}&
			\int_{\Omega} \delta \vec{u}_{n+1} \cdot \frac{\partial
					 \tensor{\epsilon}(\vec{u}_{n+1})}{\partial \vec{u}} :
					  \mathbb{C}_{\mathrm{s}} : (\tensor{\epsilon}_{n+1} -
					  \tensor{\epsilon}^*_{n+1}) d\Omega
 			+\int_{\Omega}  B \delta \vec{u}_{n+1} \cdot \frac{\partial
 			 		  \epsilon_{\mathrm{vol}}(\vec{u}_{n+1})}{\partial \vec{u}}
 			 		   {\beta}^{\mathrm{mass}}_{n+1} d\Omega = 0,
\end{split}\\
\begin{split}\label{eq::res-p-solidDD}
   \mathcal{R}^p_{n+1}  =
    {}&
   		 	\int_{\Omega} \frac{1}{\mu^{\mathrm{f}}} \grad{\delta p}_{n+1}
   		 			 \cdot \tensor{k} \cdot  \grad{{\beta}^
   		 			 {\mathrm{mass}}_{n+1}} \Delta t d\Omega
			+\int_{\Omega} \frac{1}{M} \delta p_{n+1}
			 		\beta^{\mathrm{mass}}_{n+1} d\Omega
	\\&
			+\int_{\Omega} B \delta p_{n+1} \grad{{\vec{\beta}}
					^{\mathrm{mom}}_{n+1}} : \tensor{I} d\Omega 
	=0,
\end{split}\\
\begin{split}\label{eq::res-betaMom-solidDD} 
    \mathcal{R}^{\beta^\text{mom}}_{n+1} = 
    {}&
	    		-\int_{\Omega} \grad{\delta \vec{\beta}^
	    				{\mathrm{mom}}_{n+1}  } : ( \tensor{\sigma}^
	    				\prime_{n+1} - B p_{n+1} \tensor{I}) d\Omega
		   +\int_{\partial \Omega_{\tensor{\sigma}} } \delta \vec{\beta}
		   			^{\mathrm{mom}}_{n+1}  \cdot \bar{\vec{t}}_{n+1}
		   			d\Gamma
		   +\int_{\Omega} \delta \vec{\beta}^{\mathrm{mom}}_{n+1}
		    			 \cdot \vec{\gamma}_{n+1} d\Omega = 0,
\end{split}\\
\begin{split}\label{eq::res-betaMass-solidDD}
    \mathcal{R}^{\beta^\text{mass}}_{n+1}  = 
    {}&
 			\int_{\Omega} \delta {\beta}^{\mathrm{mass}}_{n+1} \left[
 					 \frac{1}{M} (p_{n+1} - p_{n}) + B
 					  ({\epsilon_{\mathrm{vol}}}_{n+1} -
 					  {\epsilon_{\mathrm{vol}}}_n) + s_{n+1} \Delta t \right]
 					  d\Omega
	\\& 
    		   -\int_{\Omega} \grad{\delta {\beta}^{\mathrm{mass}}_{n+1}}
    		   			 \cdot \vec{q}_{n+1} \Delta t d\Omega	    
    		   +\int_{\partial \Omega_{\vec{q}} } \delta {\beta}
    		   			^{\mathrm{mass}}_{n+1} \bar{q}_{n+1} \Delta t d\Gamma
    		   	= 0,
\end{split}\\
    \mathcal{R}^{\sigma^\prime}_{n+1} = 
    {}&
			\int_{\Omega} \delta \tensor{\sigma^\prime}_{n+1} :
	 				\left( \mathbb{S}_{\mathrm{s}} :
	 					(\tensor{\sigma^\prime}_{n+1} 
	 					-\tensor{\sigma^\prime}^*_{n+1})
	 				- \grad{\vec{\beta}^{\text{mom}}_{n+1}} \right) d\Omega
	 		= 0.
	\label{eq::res-sigmaPrime-solidDD}
\end{align}
We further reduce number of independent fields (and equations) by the local (point-wise) satisfaction of Eq. \eqref{eq::res-sigmaPrime-solidDD} via:
\begin{equation}
\tensor{\sigma}^\prime_{n+1} =  {\tensor{\sigma}^\prime}^*_{n+1} + \mathbb{S}_{\mathrm{s}}^{-1} : \grad{\vec{\beta}}^{\mathrm{mom}}_{n+1}  \ \mathrm{in} \ \Omega.
\label{eq::eff-stress-solidDD}
\end{equation}

The next step in the fixed-point method after solving the above coupled equations is to solve the local minimization problem. According to Eq. \eqref{eq::solidDD-prob-def}, the local minimization for this hybrid scheme reads as follow,
\begin{equation}
{\psElemSolid}^*_{n+1}(\bar{\vec{x}}_{a}) = \argmin_{{\psElemSolid}^*_{n+1}   \in \dataSetSolid_{n+1}} (\distSolid({\psElemSolid}^*_{n+1}, {\psElemSolid}_{n+1} (\bar{\vec{x}}_{a})))^{2},\  \mathrm{given} \ {\psElemSolid}_{n+1}(\bar{\vec{x_{a}}}) \in \hat{\mathcal{C}}^{\mathrm{coupled}}_{n+1}, 
\end{equation}
where $\bar{\vec{x}}_{a} \in \Omega^h$ is an integration point in the  domain $\Omega^h$, i.e., Gauss quadrature point. We will discuss how to solve this local problem in Sec. \ref{sec:localminimization}.

\section{Local minimization: physical response projection onto material space}
\label{sec:localminimization}
This section describes the procedure to project the hydro-mechanical responses obtained 
in the global minimization step onto the points that belong to the material phase space for each integration point.
The data-driven solver searches for the nearest neighbor point to a hydro-mechanical response inside the material phase space (database). Since the local minimization problems for the three formulations defined in Sections \ref{sec:fullyDD}, \ref{sec:hybrid1} and \ref{sec:hybrid2} 
are similar (except for the distance measure and the basis of the material space), the local minimization description is generic.

In this research, the nearest neighbor search (NNS) is used to facilitate this minimization step. 
In Subsection \ref{sec:nns}, we discuss alternative schemes for the fast NNS and explain why the k-d tree is an appropriate choice. 
We then show how the energy metric can be rewritten as a Euclidean metric through 
a linear projection. 
This mapping provides a straightforward way of using available open-source packages for the fast NNS since these packages are generally developed and optimized for Minkowski's metric functions.

\subsection{Nearest Neighbor Search} \label{sec:nns}
Here, we first describe the Nearest Neighbor Search (NNS), 
which enables us to locate the optimal data points in the phase space $(\tensor{\epsilon}, \tensor{\sigma})$ and 
$(\grad p, \vec{q})$ equipped with different distance metrics.  We also provide an overview and compare different NNS approaches and justify our choice of using the k-d tree data structure for the data-driven poroelasticity problem.

Assume $P$ is a set of $n$ points $\vec{p}_i  \in \mathbb{R}^k $ with $k$ dimensions, i.e., $P=\{  \vec{p}_1, \vec{p}_2,\ldots, \vec{p}_n  \}$. Nearest Neighbor Search (NNS) in a metric space is an optimization problem that determines the closest point $\vec{p}_i \in P$ to a query point $\vec{q} \in \mathbb{R}^k$ where $P$ and $q$ belong to a metric space $M$ equipped with a well-defined distance function $d \colon M \times M \to \mathbb{R}^+$ \citep{muja2014scalable, Shakhnarovich2006Nearest}, i.e., 
\begin{equation}
\mathrm{NNS}(\vec{q}, P; M) = \argmin_{\vec{p}_i \in P} \ d(\vec{q}, \vec{p}_i),
\end{equation}
where a distance function must possess the following properties:
\begin{align}
&
d(\vec{q},\vec{p}_i) = 0 \Leftrightarrow \vec{q}  \equiv \vec{p}_i,
\label{eq::metric-cond-zero}
\\&
d(\vec{q},\vec{p}_i) = d(\vec{p}_i, \vec{q}),
\label{eq::metric-cond-sym}
\\&
d(\vec{q},\vec{p}_j) \le d(\vec{q},\vec{p}_i) + d(\vec{p}_i, \vec{p}_j).
\label{eq::tri-ineq}
\end{align}
In our application, for example, point $\vec{p}_i$ encodes components of strain-stress pair. In three-dimensional cases, $k=12$ since both strain and stress are symmetric. 
A Brute-force algorithm that compares every element in the set is easy to implement but is inefficient for large sets $P$ with a huge amount of queries, due to the resultant linear query time $O(nk)$. There are algorithms developed to make the search fast enough to be 
close to a logarithmic query time for a variety of applications and circumstances \citep{malkov2018efficient, Shakhnarovich2006Nearest, muja2014scalable}.
A fast algorithm for NNS relies heavily on leveraging efficient data structures.
However, more efficient data structures could also be more complex and less memory-efficient, known as time-space trade-offs \citep{andoni2017optimal}. For instance,  if one intends to query the distance between two points of a set of $n$ points, a naive approach is to store $n$ points in an array with the space complexity $O(n)$ and each time computes the distance of two points with the time complexity $O(n)$ (access time complexity). However, the distance between all points can be tabulated with the space complexity $O(n^2)$ in the pre-processing step, and the query time becomes $O(1)$ in this case.
Consequently, an efficient algorithm for the search must strike aw balance among speed, memory efficiency, robustness, and accuracy. 

For any specific application number of data points in $P$, dimensionality of points, metric structure, and number of query points are among the first most important factors needed to be considered for choosing an efficient algorithm (cf. \citet{Shakhnarovich2006Nearest}). 
For example, brute-force is the most efficient approach when data size is small \citep{Shakhnarovich2006Nearest}. As the number of query points or size of data set $P$ is growing the need for other efficient NNS increases.

There exist two types of fast algorithms: exact NNS and approximate NNS (ANNS). Exact NNS are based on the hierarchical space partitioning \citep{Friedman1977Algorithm, Maneewongvatana1999Analysis}. There are many variants of exact NNS, each of which proposes a different approach to split high dimensional space. Tree data structures are recognized as the efficient data structures for these type of algorithms, and k-d tree \citep{bentley1975multidimensional, Friedman1977Algorithm}, vp-tree \citep{yianilos1993data}, R-tree \citep{beckmann1990r}, and X-tree \citep{berchtold1996x} are among popular ones.  Exact NNS is in general fast algorithms loose their efficiency in dealing with high dimensional spaces, known as the curse of dimensionality. Roughly speaking, sub-linear or logarithmic time complexity of fast algorithms increase as $n$ exceeds $2^k$ limit for high dimensional spaces \citep{indyk2004nearest, rajani2015parallel}. Note that in computer science $k=20$ is still not considered as a high dimensional space \citep{indyk2004nearest, rajani2015parallel} whereas in mechanics the space of strain-stress pairs with dimensionality 12 for 3D geometry is assessed as a high dimensional space. On the other side, it is proven that speed can be significantly improved by ANNS comparing with exact NNS if we slightly sacrifice the accuracy \citep{muja2014scalable}. This means ANNS finds a point which is in the $\epsilon$-ball of the exact closest point.  Algorithms based on hashing \citep{andoni2006near}, randomized tree \citep{silpa2008optimised}, and hierarchical graphs \citep{malkov2018efficient} are widely used for ANNS.

One efficient data structure suitable for low dimensional points ($k<20$) is the $k$-dimensional tree known as k-d tree \citep{bentley1975multidimensional, rajani2015parallel}. Bentley's k-d tree as the standard k-d tree is a binary tree that recursively partitions the $\mathbb{R}^k$ space along some orthogonal hyperplanes into at most $O(\log n)$ hypercubes. Basically, each hyperplane is chosen to be orthogonal to a data axis and bisects the axis in a way that two resulting partitions have almost the same size. In the constructed tree, eventually, a maximum predefined number of data points are stored at each leaf node, and other nodes including root save information of each hyperplane. Once the tree structure is built, for a query point the algorithm traverses the tree from the root, and based on the underlying metric structure and hyperplane configuration narrows down the search into one half of the sub-tree until arrives at a leaf node. At the leaf node, it switches to the brute force distance calculation to find the closest point among the points inside the leaf node. What we intended to describe is a big picture of the idea behind k-d tree family of algorithms, and interested readers should refer to \citep{bentley1975multidimensional, Friedman1977Algorithm, Maneewongvatana1999Analysis} for more details. Note that Ball tree is suggested as an efficient alternative for high dimensional spaces $k>20$ \citep{rajani2015parallel, virtanen2020scipy} which is not studied in this research.

\begin{figure}[h]
 \centering
\includegraphics[width=0.7\textwidth]{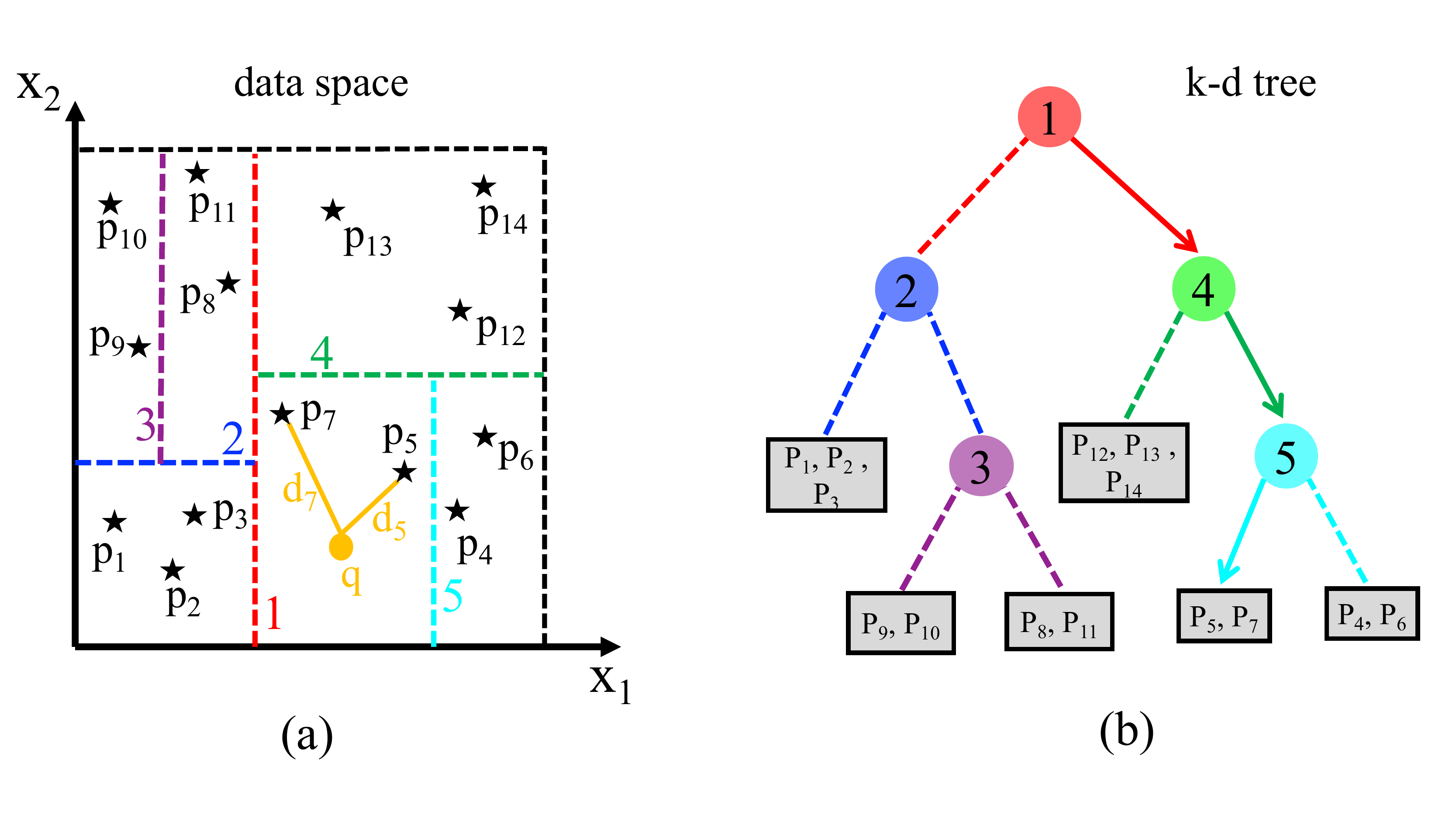}
  \caption{(a) partitioned 2D data space by some orthogonal lines; (b) equivalent representation of (a) by a k-d tree data structure.\label{fig::kd-tree-info}}
\end{figure}

\Fig \ref{fig::kd-tree-info} illustrates the process of constructing a k-d tree for a data set $P=\{ \vec{p}_1, \ldots, \vec{p}_{14} \}$ with $k=2$ and querying a new point $\vec{q}$. In \fig \ref{fig::kd-tree-info}(a), line 1 (red dashed line) which plays the role of a hyperplane in 2D data space partitions the data set into two rectangles, hypercubes in 2D space, with seven points in each. Next, line 2 splits the left rectangle constructed by line 1 into two smaller rectangles where the bottom one has three points, and the top one has four points. This process is continuing until we end up with some small rectangles which include at most three points, the maximum number of data points in leaf nodes. In \fig \ref{fig::kd-tree-info}(b), we show the constructed k-d tree that describes \fig \ref{fig::kd-tree-info}(a). Geometrical information of each hyperplane is encoded hierarchically from root node towards a level before leaf nodes, e.g., lines 1-5 in \fig \ref{fig::kd-tree-info}(a) are encoded in nodes 1-5 in \fig \ref{fig::kd-tree-info}(b). Leaf nodes store those data points inside each final hypercube. Note that k-d tree is constructed in an off-line manner. For an online query $\vec{q}$ shown as an orange circle in \fig \ref{fig::kd-tree-info}(a), the algorithm traverses the tree from the root node numbered 1 in \fig \ref{fig::kd-tree-info}(b) and at each node it determines which sub-tree holds the closest point to $\vec{q}$, based on the metric structure and hyperplane configuration, until it reaches to one of the leaf nodes. Solid arrows in \fig \ref{fig::kd-tree-info}(b) declare the path from the root to leaf node having the closest point to $\vec{q}$, $1 \to 4 \to 5 \to \mathrm{leaf}$. At the founded leaf node the algorithm calculates the distance function $d_i = d(\vec{q}, \vec{p}_i)$ to find the nearest point to $\vec{q}$, distances are shown in orange solid lines in \fig \ref{fig::kd-tree-info}(a). Note that the maximum number of points at each leaf node is a hyper-parameter and should not be chosen too large or too small. In this work, we set it equal to 10. Larger values make height of tree shorter and so faster traverse, but it has more cost during the distance calculation over all leaf nodes. 

Minkowski's $l_p$ distance metrics are among widely used distance functions in several open-source packages SciPy \citep{virtanen2020scipy}, FLANN \citep{muja2009fast}, Faiss \citep{johnson2019billion}, and Annoy \citep{bernhardsson2013annoy}, due to the well-documented theoretical and practical performance. It is noteworthy that more complex distance functions may increase time complexity for the tree construction and query. For instance, \citet{Friedman1977Algorithm} empirically shows that the simplest case $l_p=\infty$, i.e., maximum coordinate distance, outperforms Euclidean distance $l_p=2$.

Herein, we do not intend to develop a data-structure effective for the energy-like distance functions Eqs. \eqref{eq::dist-fully}, \eqref{eq::dist-fluid}, and \eqref{eq::dist-solid} in the data-driven formulations. Instead, we will introduce an isomorphism that maps the energy-like metric to Euclidean metric, so all the data-structures established for Euclidean metric can be easily utilized in our application.

\begin{remark}
Bentley's k-d tree has average height $O(\log n)$, average construction time $O(k n\log n)$, average query time $O(\log n)$ for Minkowski's $l_p$ metric \citep{Friedman1977Algorithm, Maneewongvatana1999Analysis}. The query complexity in worst case scenario is $O(kn^{1-1/k})$ \citep{lee1977worst, maneewongvatana1999s}.
\end{remark}

\begin{remark}
In this work, we utilize k-d tree data structure implemented in class \texttt{scipy.spatial.cKDTree} of open source package SciPy \citep{virtanen2020scipy}.
\end{remark}

\subsection{local metric minimization}\label{sec:local-metric-minimization}
Here, we describe how to perform the local minimization efficiently to utilize techniques available in the NNS literature. First, we focus on the fully data-driven case and demonstrate 
how the NNS may simplify the local distance calculations. 
Then, we will formulate the solid and fluid local distance calculation in a unified and compact formulation.

Recall the local minimization problem corresponding to the distance function in Eq. \eqref{eq::local-dist-fullyDD}. 
Since we assume that the solid and fluid phase spaces are orthogonal to each other, 
the local minimization can be done separately for each phase space, i.e., 
\begin{align}
&
{\psElemSolid}^*(\bar{\vec{x}}_{a}) = \argmin_{ {\psElemSolid}^* \in \dataSetSolid} [\distSolid ( {\psElemSolid}^*, \psElemSolid (\bar{\vec{x}}_{a}) )]^{2}; \ \mathrm{given} \ \psElemSolid (\bar{\vec{x}}_{a}),
\\&
{\psElemFluid}^*(\bar{\vec{x}}_{a})  = \argmin_{ {\psElemFluid}^* \in \dataSetFluid} [\distFluid ( {\psElemFluid}^*, \psElemFluid (\bar{\vec{x}}_{a})   )]^{2} ;\ \mathrm{given} \ \psElemFluid (\bar{\vec{x}}_{a}),
\end{align}
where each sub-problem is solved by an independent 
nearest neighbor search inside the corresponding database. 
The designated metric for the fully data-driven formulation reduces the phase space dimensionality by splitting a 
phase space that contains 18 bases (12 for solid and 6 for fluid in 3D cases) 
 into two smaller phase spaces. 
 The advantage of this metric is two-fold. First, the NNS algorithms may perform more efficiently
 in the two lower-dimensional phase space. Second, these two NNS tasks can be executed in parallel to further reduce computational time.

These local minimization problems should be solved for every quadrature point $\bar{\vec{x}}_{a}$ where $a = 1, 2, ..., n_{\text{int}}$ where $n_{\text{int}}$ is the total number of integration points. 
These global-local iterations continue until updated variables during the fixed-point iterations converge or their variations remain within a small tolerance. The data-driven solver may perform several fixed-point iterations within a time step. Hence the efficiency of the local minimization 
is a dominant factor for the speed of the simulations, especially for large-scale problems. 
In the following, we switch to conventional matrix notation in Linear Algebra. Consequently, the summation convention is no longer hold, in this section, unless we specify the other way around.

Note that both terms in solid distance function Eq. \eqref{eq::dist-solid} have the same structure as $\tensor{\tau} : \mathbb{K} : \tensor{\tau}$ where $\tensor{\tau}$ is a symmetric \nth{2} order tensor which is either $\tensor{\tau} = \tensor{\epsilon}-\tensor{\epsilon}^*$ or $\tensor{\tau} = \tensor{\sigma}^\prime-{\tensor{\sigma}^\prime}^*$, and $\mathbb{K}$ is a symmetric positive definite \nth{4} order tensor which is either $\mathbb{C}_{\mathrm{s}}$ or $\mathbb{S}_{\mathrm{s}}$. This energy-like scalar variable can be represented in a full vector-matrix notation:
\begin{equation}
\tensor{\tau} : \mathbb{K} : \tensor{\tau} = \vec{t}^T \tensor{K} \vec{t},
\label{eq::dist-vec-mat}
\end{equation}
where $\vec{t}$ is a column vector that stores information about $\tensor{\tau}$ and $\tensor{K}$ is a symmetric positive definite matrix that encodes information of $\mathbb{K}$. There are multiple ways to obtain such vector-matrix quantities that preserve the left-hand side in Eq. \eqref{eq::dist-vec-mat}, e.g., Kelvin and Voigt notations. \citep{mehrabadi1990eigentensors, itskov2000theory} have shown that $\mathbb{K}$ and $\tensor{K}$ share common spectral characteristics such as eigenvalues and eigenbases in the case of Kelvin and Voigt representations.

Fluid distance function Eq. \eqref{eq::dist-fluid} can be represented in a vector-matrix format without any further action, since $\tensor{C}_{\mathrm{f}}$ and $\tensor{S}_{\mathrm{f}}$ are \nth{2} order tensors which are also matrix. Hence, they have already followed the format in the right-hand side of Eq. \eqref{eq::dist-vec-mat}. In the view of vector-matrix notation both solid and fluid distances have the same format but with different sizes, e.g., in 3D set-up $\vec{t}$ has  $6$ components for the solid distance and $3$ components for the fluid part. For the rest of the discussion in this section, we develop the framework for a general, unified metric as follows: 
\begin{equation}
d(\vec{t}, \vec{r}) = \vec{t}^T \tensor{K} \vec{t} + \vec{r}^T \tensor{M} \vec{r},
\end{equation}
where $\vec{t}$ and $\vec{r}$ are column vectors with size $n$, and $\tensor{K}$ and $\tensor{M}$ are constant symmetric positive definite matrices. For solid distance function, we incorporate Voigt notation to store elements of tensors $\tensor{\epsilon} - \tensor{\epsilon}^*$ and $\tensor{\sigma}^\prime - {\tensor{\sigma}^\prime}^*$ into vectors $\vec{t}$ and $\vec{r}$, respectively, and build matrices $\tensor{K}$ and $\tensor{M}$ equivalent to tensors $\mathbb{C}_{\mathrm{s}}$ and $\mathbb{S}_{\mathrm{s}}$, respectively. For fluid distance function, vectors $\vec{t}$ and $\vec{r}$ are equal to vectors $\grad{p} - {\grad{p}}^*$ and $\vec{q}-\vec{q}^*$, respectively, and matrices $\tensor{K}$ and $\tensor{M}$ are exactly equal to tensors $\tensor{C}_{\mathrm{f}}$ and $\tensor{S}_{\mathrm{f}}$. Since any symmetric positive definite matrix $\tensor{K}$ possess decomposition $\tensor{K} = \tensor{C}^T \tensor{C}$ (e.g., via Cholesky factorization and matrix square root) we can write:

\begin{equation}
d(\vec{t}, \vec{r}) = \vec{t}^T \mat{K} \vec{t} + \vec{r}^T \mat{M} \vec{r} = \vec{t}^T \mat{C}^T \mat{C} \vec{t} + \vec{r}^T \mat{S}^T \mat{S} \vec{r} = \bar{\vec{t}}^T \bar{\vec{t}} + \bar{\vec{r}}^T \bar{\vec{r}},
\end{equation}
where $\bar{\vec{t}} = \mat{C} \vec{t}$ and $\bar{\vec{r}} = \mat{S} \vec{r}$ are projected vectors of $\vec{t}$ and $\vec{r}$, also $\mat{M}  = \mat{S}^T \mat{S}$. The distance function can be further reduced to a more compact form by concatenation of vectors $\bar{\vec{t}}$ and $\bar{\vec{r}}$ into
$\bar{\vec{w}} = \begin{bmatrix}
\begin{tabular}{ c}
$\bar{\vec{r}}$\\
\hline
$\bar{\vec{t}}$
\end{tabular}
\end{bmatrix} \in \mathbb{R}^{2n}$
as follows:
\begin{equation}
\label{eq::l2-mapped}
d(\vec{t}, \vec{r}) = \bar{\vec{t}}^T \bar{\vec{t}} + \bar{\vec{r}}^T \bar{\vec{r}} = \bar{\vec{w}}^T \bar{\vec{w}}.
\end{equation}
The linear projection via the matrices $\tensor{C}$ and $\tensor{S}$ for the corresponding variables is introduced for convenience. 
This simple treatment enables us to adopt open-source libraries 
developed by the NNS community for 
Euclidean space (and in a more general sense Minkowski space) 
with minimal implementation efforts.

\begin{algorithm}[h!]
    \begin{algorithmic}[1]
    \State \textbf{Input:} Database of 
    		$\{ (\tensor{\epsilon}^*_i, {\tensor{\sigma}^\prime}^*_i) \}_{i=1}^{N}$ 
    		strain-effective stress tensor pairs (\nth{2} order symmetric tensors) in 
    		$m = 1, 2, 3$ dimensions, numerical parameters 
    		$\mathbb{C_{\mathrm{s}}}$ and $\mathbb{S_{\mathrm{s}}}$
    		(\nth{4} order super symmetric positive definite tensors)
    \State \textbf{Output:} Tree object $\mathcal{T}^{\mathrm{s}}$, $\mat{C} \in \mathbb{R}^{n \times n}$, $\mat{S} \in \mathbb{R}^{n \times n}$, $\mat{K}^{n \times n}$, and $\mat{M}^{n \times n}$
    \State $\mat{K} \gets \mathrm{Voigt}(\mathbb{C_{\mathrm{s}}})$, $\mat{M} \gets \mathrm{Voigt}(\mathbb{S_{\mathrm{s}}})$ \Comment{Apply Voigt notation (or Kelvin notation)}
     \State $\mat{C} \gets \mathrm{Fact(\mat{K} )}$, $\mat{S} \gets \mathrm{Fact(\mat{M} )}$  \Comment{Matrix factorization, e.g. Cholesky or square root}
     \State $\mat{D} \gets \mat{0}$ \Comment{Zeros matrix of size $N\times n$ where $n=m(m+1)/2$}
    \For{$i = 1:N$} \Comment{Loop over $N$ data points (pairs of strain-stress)}
    		\State $\vec{t} \gets \mathrm{Voigt}(\tensor{\epsilon}^i)$, $\vec{r} \gets \mathrm{Voigt}({\tensor{\sigma}^\prime}^i)$
    		\State $\mat{D}_{i,:} \gets \left[ \left( \mat{C} \vec{t} \right)^T | \left( \mat{S} \vec{r} \right)^T \right]$
	\EndFor\label{alg::data-proj}
     \State $\mathcal{T}^{\mathrm{s}} \gets \mathrm{KDTree(\mat{D})}$ \Comment{Apply the kd-tree algorithm \texttt{scipy.spatial.cKDTree}}
\caption{Off-line kd-tree construction}\label{algo::proj-data}
    \end{algorithmic}
\end{algorithm}

\begin{algorithm}[h!]
    \begin{algorithmic}[1]
    \State \textbf{Input:} Physical strain-effective stress 
    		pair $(\tensor{\epsilon}, \tensor{\sigma}^\prime)$ at 
    		a quadrature point, k-d tree object of solid database $\mathcal{T}^{\mathrm{s}}$
    \State \textbf{Output:} Closest material strain-effective stress
          pair $(\tensor{\epsilon}_i^*, {\tensor{\sigma}_i^\prime}^*)$
          from database, distance function $d_i$ 
     \State $\vec{t} \gets \mathrm{Voigt}(\tensor{\epsilon})$,
     	$\vec{r} \gets \mathrm{Voigt}(\tensor{\sigma}^\prime)$
    	\State $\bar{\vec{w}}^T \gets \left[ \left( \mat{C} \vec{t} \right)^T | \left( \mat{S} \vec{r} \right)^T \right]$ 
    		\Comment{$\mat{C}$ and $\mat{S}$ are stored once before
    		 		the on-line simulation}
    	\State 
    		$d_i, \bar{\vec{w}}_i^* \gets \mathrm{NNS}(\bar{\vec{w}}, \mathcal{T}^{\mathrm{s}})$
    		\Comment{A new query $\bar{\vec{w}}$ for k-d tree nearest 
    			neighbor search}
    	\State $ \bar{\vec{t}}_i^* \gets \bar{\vec{w}}_i^*|_{1:n}$, 
    		$ \bar{\vec{r}}_i^* \gets \bar{\vec{w}}_i^*|_{n+1:2n}$
    		\Comment{Separate strain and stress related components}
    	\State $ \vec{t}_i^* \gets \mat{K}^{-1} \mat{C}^{T}  \bar{\vec{t}}_i^*$,
    		$ \vec{r}_i^* \gets \mat{M}^{-1} \mat{S}^{T}  \bar{\vec{r}}_i^*$ 
    			\Comment{inverse projection onto real configuration}
     \State $\tensor{\epsilon}_i^* \gets \mathrm{Voigt}^{-1}(\vec{t}_i^*)$, ${\tensor{\sigma}_i^\prime}^* \gets \mathrm{Voigt}^{-1}(\vec{r}_i^*)$ \Comment{From vector representation to tensor}
\caption{Local minimization: nearest data point calculation}\label{algo::NNS}
    \end{algorithmic}
\end{algorithm}

We provide Pseudo-codes \ref{algo::proj-data} and \ref{algo::NNS} to summarize this section. In Pseudo-code \ref{algo::proj-data}, we first project database and then construct the kd-tree object for the projected database by utilizing the functionality of\\ \texttt{scipy.spatial.cKDTree}. This task is executed once before the simulation.
 We present this pseudo-code for brevity only for solid database, but the same procedure should be applied for the fluid database. 
The resultant tree object of this pseudo-code stores projected data points in an efficient way for fast nearest neighbor search during the simulation.
 Recall that the global minimization step does not have any projected quantities, so the input for the nearest neighbor search is in the real configuration of databases. In Pseudo-code \ref{algo::NNS}, we first project the new query point, and then we perform the nearest neighbor search.
  Finally, the nearest point should be projected back to the real configuration; otherwise, it cannot be used in the global optimization step. Instead of projecting back, the real database could be stored as well. In this case, the real data point could be tracked by the data index without any extra matrix multiplication due to the inverse projection.
 This latter approach could be faster if the original database is stored in a hash table, but it may also exhibit memory deficiency. Notice that the local minimization step is the execution of Pseudo-code \ref{algo::NNS} at each fixed-point iteration over all quadrature points separately.

\begin{remark}
Symmetric square matrix $\mat{K}_{n \times n}$, with factorization $\mat{K} = \mat{C}^T \mat{C}$ where $\mat{C}_{m \times n}$, is positive definite if and only if $\mat{C}$ has full column rank; see Theorem 7.2.7 in \citet{horn2012matrix}. 
\end{remark}

\begin{remark}
Cholesky factorization and square root matrix are unique for symmetric positive definite matrix; see Corollary 7.2.9 and Theorem 7.2.6 in \citet{horn2012matrix}.
\end{remark}

\begin{remark}
Note that if $\mat{C}$ satisfies $\mat{K} = \mat{C}^{T}\mat{C}$ then for any unitary matrix (rotation) $\mat{Q}$ the new matrix $\mat{Q}\mat{C}$ also factorizes $\mat{K} =(\mat{\mat{Q}\mat{C}})^{T}\mat{\mat{Q}\mat{C}}$. Therefore, symmetric positive definite matrix $\mat{K}$ can be factorized by infinite matrices unless $\mat{K}$ is identity matrix.
\end{remark}

\begin{remark}
Note that the Cholesky factorized matrix $\mat{L}$ and its inverse are lower triangular matrices which can be optimized to reduce operations approximately by half in case of matrix multiplication or summation in comparing with operations involving square root matrix $\mat{K}^{1/2}$ which is a full matrix. Also, this reduction of basic operations reduces the chance of over floating. In this work, we do not compare computational performance of available decomposition schemes for symmetric positive definite matrices, and we use square root factorization.
\end{remark}

\section{Algorithm and numerical implementation} \label{sec:algoDD}
This section intends to bundle all the discussed ingredients for solving poroelasticity with/out constitutive laws. For brevity, the overall algorithm is presented for the fully data-driven framework, but it is kept general and can be used for hybrid formulations.

Algorithm \ref{algo::fullyDD} provides general steps used in the fully data-driven formulation. In this research, we first solve global minimization, so the first fixed-point iteration of the first time step needs an initialization for the distribution of material sates at each quadrature point (see line 5 of Algorithm \ref{algo::fullyDD}). This initialization can be performed by a random assignment of data points from the database. The initial assignment affects the fixed-point method convergence, and different assignments may lead to slightly different final solution \citet{kirchdoerfer2016data}. For nonlinear cases, hybrid formulation 1 with nonlinear model-based constitutive law, such a random initialization may result in non-convergence of the Newton-Raphson algorithm. Hence, in nonlinear cases, a homogeneous assignment with zero values (or residual states) is more meaningful. Notice that there is no need to do more than one Newton-Raphson iteration for the fully data-driven formulation regardless of any hidden non-linearity in the material database since this formulation is always linear, as discussed earlier. Although the material states (line 5 of Algorithm \ref{algo::fullyDD}) can be re-initialized for other time steps we use values from the previous time step.

\begin{algorithm}[h!]
    \begin{algorithmic}[1]
    \State \textbf{Input:} solid database $\{ (\tensor{\epsilon}^*_k, {\tensor{\sigma}^\prime}^*_k) \}_{k=1}^{N}$, fluid database $\{ (\grad{p}^*_l, \vec{q}^*_l ) \}_{l=1}^{M}$, numerical parameters $\mathbb{C_{\mathrm{s}}}$, $\mathbb{S_{\mathrm{s}}}$, $\tensor{C}_{\mathrm{f}}$, and $\tensor{S}_{\mathrm{f}}$
    \State construct k-d tree object $\mathcal{T}^{\mathrm{s}}$ for solid database via Algorithm \ref{algo::proj-data}
    \State construct k-d tree object $\mathcal{T}^{\mathrm{f}}$ for fluid database via a fluid version of Algorithm \ref{algo::proj-data}
    \For{$i = 1:N_{ts}$} \Comment{loop over $N_{ts}$ time steps}
		    \State initialize $(\tensor{\epsilon}^*, {\tensor{\sigma}^\prime}^*)$ and $(\grad{p}^*, \vec{q}^*)$ at each quadrature point from database.  \Comment{required for the first time step $i=1$ and it can be a random assignment.}
    		    \While {$\mathrm{\textit{Fixed-Point not converged}}$}
    		    			\State initialize degree of freedoms corresponding to 
    		    					   fields $\vec{u}_i$,
    		    							  	${\vec{\beta}}^{\mathrm{mom}}_i$,
    		    							  	$p_i$, $\beta^{\mathrm{mass}}_i$.
    		    			\While {$\mathrm{\textit{Newton-Raphson not converged}}$} \Comment{global minimization}
    		    					\State solve linearized Eqs.
    		    					    \eqref{eq::res-u-FullyDD},
    		    						\eqref{eq::res-p-FullyDD},
								\eqref{eq::res-betaMom-FullyDD}, and
    		    						\eqref{eq::res-betaMass-FullyDD}.
    		    					\State update degree of freedoms corresponding to 
    		    							  fields $\vec{u}_i$,
    		    							  			${\vec{\beta}}^{\mathrm{mom}}_i$,
    		    							  			$p_i$, $\beta^{\mathrm{mass}}_i$.
		    		    \EndWhile
		    		    	\For{$j = 1:N_{qp}$} \Comment{loop over $N_{qp}$ 
		    		    												quadrature points}
		    		    			\State update physical stress by 
		    		    				Eq. \eqref{eq::eff-stress-dd}.
		    		    			\State local minimization for solid part via Algorithm 
		    		    				\ref{algo::NNS}.
		    		    			\State update physical Darcy's velocity by 
		    		    				Eq. \eqref{eq::darcy-velo-dd}.
		    		    			\State local minimization for fluid part via a fluid 
		    		    				version of Algorithm \ref{algo::NNS}.
		    		    	\EndFor
    		    \EndWhile
    		    \State record solution for next time marching.
	\EndFor
\caption{Fully data-driven poroelasticity}
\label{algo::fullyDD}
    \end{algorithmic}
\end{algorithm}

\section{Numerical Examples}
\label{sec:numericalexamples}

Now we examine different aspects of the proposed data-driven formulations through six examples. 
Before presenting the actual poroelasticity problems, in the appendix \ref{kd-tree-study}, we show how effective is the k-d tree search for even a very simple problem. Then we will verify data-driven poroelasticity formulations by solving two well-known problems in the literature of geomechanics and biomechanics with available analytical solutions. Convergence with respect to the amount of data and fixed-point iterations will be studied. Three other problems will be investigated to demonstrate the capability and robustness of formulations in more complex circumstances such as non-linearity of material behavior. In the last two examples, we will discuss the fidelity and availability of data and the effectiveness of hybrid data-driven formulations. In the last example, we showcase the application of the hybrid data-driven formulation in dealing with an extreme condition where fluid flow database possesses a highly oscillatory porosity-permeability behavior. At the same time, the solid skeleton deformation follows a nonlinear porosity and pressure-dependent behavior. In all of the poroelastic examples, we utilize the k-d tree search method unless otherwise is noted. These examples are solved with our in-house data-driven solver developed in the Python programming language. The k-d tree data structure is adopted from \texttt{scipy.spatial.cKDTree} class of the open-source package SciPy \citep{virtanen2020scipy}. Linear systems are solved by the LU factorization scheme implemented in the class  \texttt{scipy.linalg.lu} of the SciPy package. We perform this factorization once for linear cases when material behavior is linear or the fully data-driven scheme is used. Numerical experiments were run on a computer with a Quad-Core Intel Core i5 processor running at 1.4 GHz using 16 GB of RAM.

\subsection{Verification exercise 1: Terzaghi's consolidation problem} 
\label{sec::problem-terzaghi}
In this problem, we verify the three proposed formulations in a two-dimensional computational set-up and compare the accuracy and convergence of the three formulations. 

\begin{figure}[h]
 \centering
\includegraphics[width=0.4\textwidth]{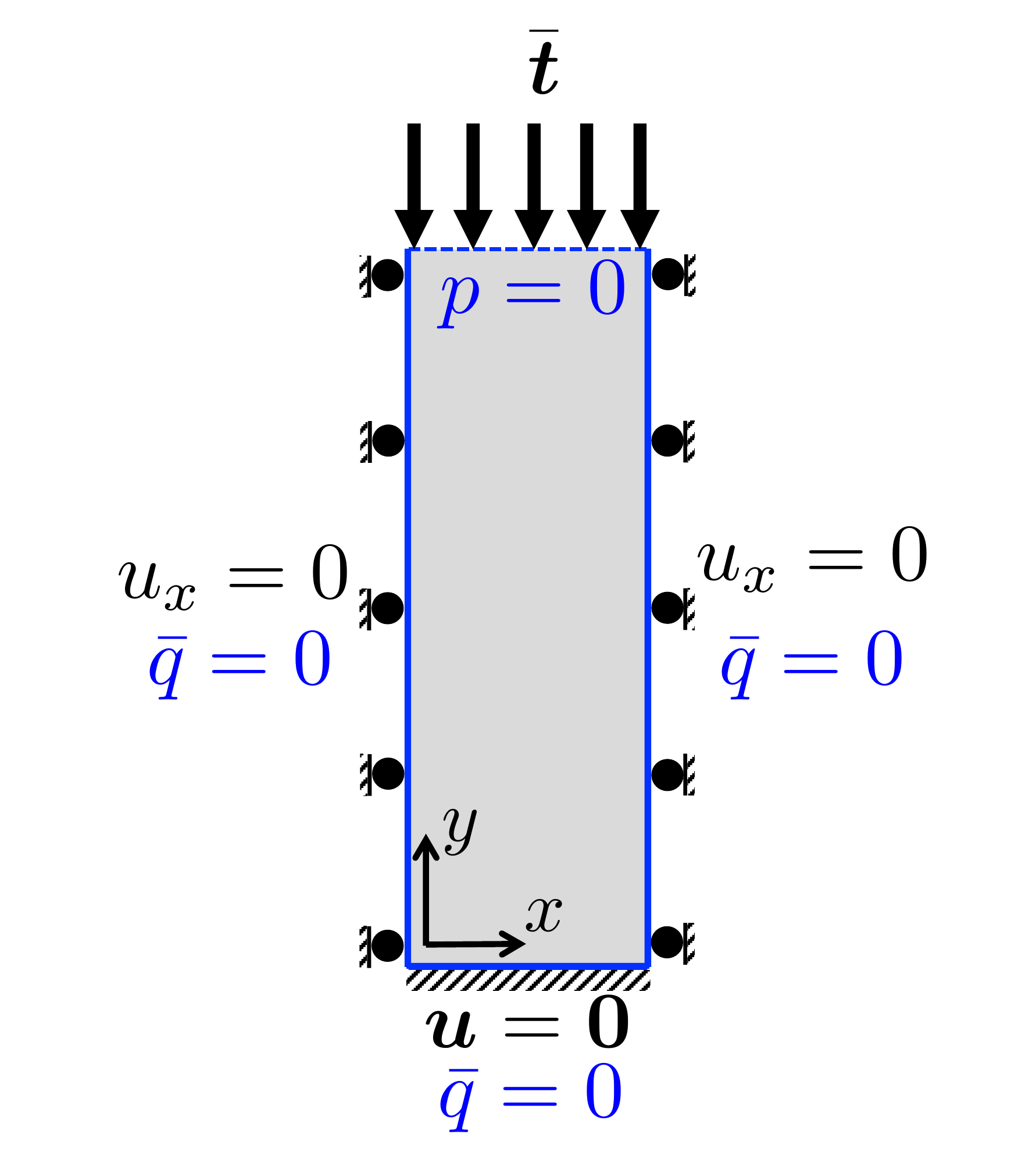}
  \caption{Terzaghi consolidation problem.\label{fig:terzaghi-problem-statement}}
\end{figure}

Terzaghi's consolidation problem has been previously used to verify Finite Element codes for modeling the consolidation process of two-phase porous media \citep{borja1991one, zienkiewicz1999computational, teichtmeister2019aspects, Castelletto2015, kim2011stability, white2008stabilized, korsawe2006finite, sun2014modeling, wei2016stabilized, wang2016semi}. The analytical solution is derived for the one-dimensional problem. However, we solve the problem in a 2D domain (see \fig \ref{fig:terzaghi-problem-statement}) to validate the formulation and implementation for two-dimensional elements.

According to \fig \ref{fig:terzaghi-problem-statement}, symmetry boundary conditions are applied laterally. Traction loads $\bar{\vec{t}} = (0, -0.9) \ \mathrm{GPa}$ are suddenly applied (at $t=0\mathrm{s}$) over the drained boundary and remained constant during the simulation. The bottom boundary is clamped. Initial conditions are at rest, i.e., $p_0=0$ and $u_0=0$. We use $1 \times 20$ quadrilateral elements to mesh a domain of $0.1 \mathrm{m} \times 1 \mathrm{m}$ ($\mathrm{width} \times \mathrm{height}$). Numerical integration is performed by the 4-point Gaussian quadrature rule. Time increment is set to $\Delta t=0.1\mathrm{s}$, and simulations end at $t_{\mathrm{end}} = 10 \mathrm{s}$. We will use material parameters listed in Table \ref{tab:terzaghi-params} to generate artificial data needed for each data-driven scheme.

We define the following error measures to study the convergence of each scheme:
\begin{align}
	&
	\mathrm{Err}_{t}(*^{\mathrm{DD}}_{t}, *^{\mathrm{ref}}_{t})  = \frac{  \int_{\Omega} |*^{\mathrm{DD}}_t - *^{\mathrm{ref}}_t| d\Omega}{  \int_{\Omega} |*^{\mathrm{ref}}_t| d\Omega},
	\label{eq::err-space}
	\\&
	\mathrm{Err}(*^{\mathrm{DD}}, *^{\mathrm{ref}})  = \frac{\Delta t}{t_{\mathrm{end}}} \sum_{t=0}^{t_{\mathrm{end}}} \mathrm{Err}_{t}(*^{\mathrm{DD}}_{t}, *^{\mathrm{ref}}_{t}),
	\label{eq::err-space-time}
\end{align}
where $*$ stands for fields including $p$, $\partial p/ \partial y
$, $q_y$, $u_y$, $\epsilon_{yy}$, and $\sigma_{yy}$, subscript $t$ indicates solution at specific time, and superscripts $\mathrm{DD}$ refers to one of the data-driven schemes. Functions $\mathrm{Err}_{t}$ and $\mathrm{Err}$ calculate spatial error at time $t$ and total error over space-time, respectively. Reference solution $*^{\mathrm{ref}}$ corresponds to the solution obtained from analytical expression or conventional model-based u-p finite element. In the forthcoming results, normalized pressure, displacement and height are defined as $p/ \bar{t}_y$, $u_y/H$, and $y/H$, respectively.

\begin{table}[h]
\centering
\caption{Material parameters for Terzaghi problem \label{tab:terzaghi-params}}
\small%
\begin{tabular}{lll}
\hline 
Physical parameter & Unit & Value\tabularnewline
\hline 
\hline 
Young's modulus ($E$) & GPa & $70$\tabularnewline
Poisson's ratio ($\nu$) & - & $0$\tabularnewline
Plane condition & - & plane strain\tabularnewline
Intrinsic permeability ($k$) & $\mathrm{m}^2$ & $3.0612\times10^{-9}$\tabularnewline
Fluid dynamics viscosity ($\mu$) & Pa.s & $0.001$\tabularnewline
Biot coefficient ($B$) & - & $1$\tabularnewline
Biot modulus  ($M$) & GPa & $266.667$\tabularnewline
\hline 
\end{tabular}
\end{table}

\subsubsection{Hybrid formulation 1: data-driven fluid phase}
Hooke's law in plane strain is used as the constitutive law for the hybrid formulation, used in the first term of Eq. \eqref{eq::res-betaMom-FluidDD}, but the flow part relies on data. To compare data-driven solution with model-based approach, we generate a database with $N$ equidistant points sampled from linear Darcy's law by setting $\partial p / \partial x=0$ and $-8.6 \ \mathrm{GPa} /\mathrm{m}  \le  \partial p  / \partial y \le  4.3 \ \mathrm{GPa} /\mathrm{m}$. In this way, we expect to recover the model-based solution within the limit of infinite data points. According to the problem's one-dimensional nature, pressure gradient and fluid velocity are zero in the $x$ direction. Hence, we can generate a database without including any variations in the $x$ direction, the same idea used in \citep{kirchdoerfer2016data}. This simplification helps to generate smaller databases for this particular problem. In one case, we will show that even if we include some variations of $\partial p / \partial x$ into the database, the solver will find the closest data points with $ \partial p / \partial x \to 0$ as final solutions. The prior knowledge about model-based FEM solution (i.e., conventional u-p formulation with known constitutive laws) indicates $ -4.3 \ \mathrm{GPa} / \mathrm{m} \le \frac{\partial p}{\partial y}|^{\mathrm{FEM}} \le 0$. Hence, we have sufficient out of solution candidates in the database to testify the accuracy of data-driven solvers. Also, to make the situation difficult for the solver, fixed-point iteration starts with initial homogeneous assignment $ \partial p / \partial y = 4.3\ \mathrm{GPa} / \mathrm{m}$ at every quadrature point which is completely out of the solution space. The same configuration will be used for the fluid part of the fully data-driven formulation, as well.

The effect of numerical parameters in metric functions, see Eqs. \eqref{eq::dist-fully}, \eqref{eq::dist-solid}, and \eqref{eq::dist-fluid}, on the overall performance of data-driven scheme is not negligible and needs a separate study to find a systematic way of fine-tuning of these values \citep{leygue2018data}. Based on a suggestion in \citep{he2020physics}, we assume optimal values for $\mathcal{C}_{\mathrm{s}}$ and $\tensor{C}_{\mathrm{f}}$ are equal to elasticity and permeability tensors, respectively. Although this is not a fair decision, since it means we have already known the underlying structure in data, it helps us compare the performance of different data-driven formulations by freezing artifacts of these numerical parameters, assuming their numerical parameters are optimal.

\subsubsection{Hybrid formulation 2: data-driven solid phase}
In this case, the Darcy's law serves as the constitutive law in the hybrid formulation 2 (see Sec. \ref{sec:hybrid2}). Due to the one-dimensional characteristic, we sample $N$ equidistant points for $-0.026 \le \epsilon_{yy} \le 0.013$ and set $\epsilon_{xx}= \epsilon_{xy}=0$ in the database. Note that the prior knowledge shows $-0.013 \le \epsilon_{yy}^{\mathrm{FEM}} \le -0.0027$ base on the model-based FEM solution. To testify the robustness of the formulation, we force the fixed-point iteration starts with an initially homogeneous data assignment $\epsilon_{yy}=0.013$ which is completely out of the solution space, instead of random assignment.

\subsubsection{Fully data-driven formulation: data-driven fluid and solid phases}
We employ the formulation described in Sec. \ref{sec:fullyDD} which requires two separate data sets for solid and fluid parts. These data sets are generated as described above within the same ranges and conditions for pressure gradient and strain tensor. Note that, in the fully data-driven scheme, when we say $N$ data points, it means we have two separate data sets with $N$ points in each for solid and fluid databases. To compare convergence, the initial assignment for the fixed-point iteration in the fully data-driven scheme is the same as the hybrid simulations.

\subsubsection{Discussion}
\begin{figure}[h]
 \centering
 \subfigure[]
{\includegraphics[width=0.45\textwidth]{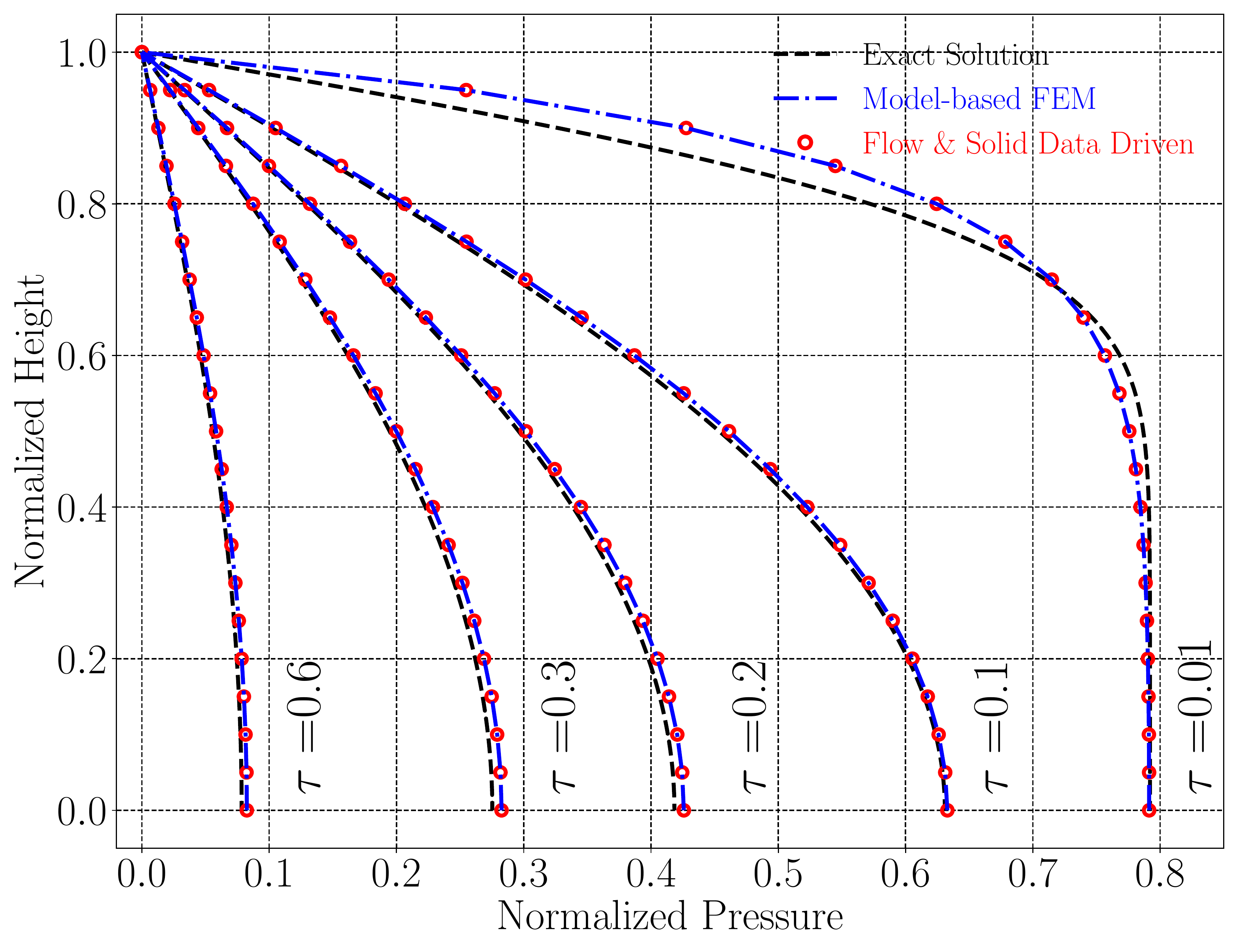}}
\hspace{0.01\textwidth}
 \subfigure[]
{\includegraphics[width=0.45\textwidth]{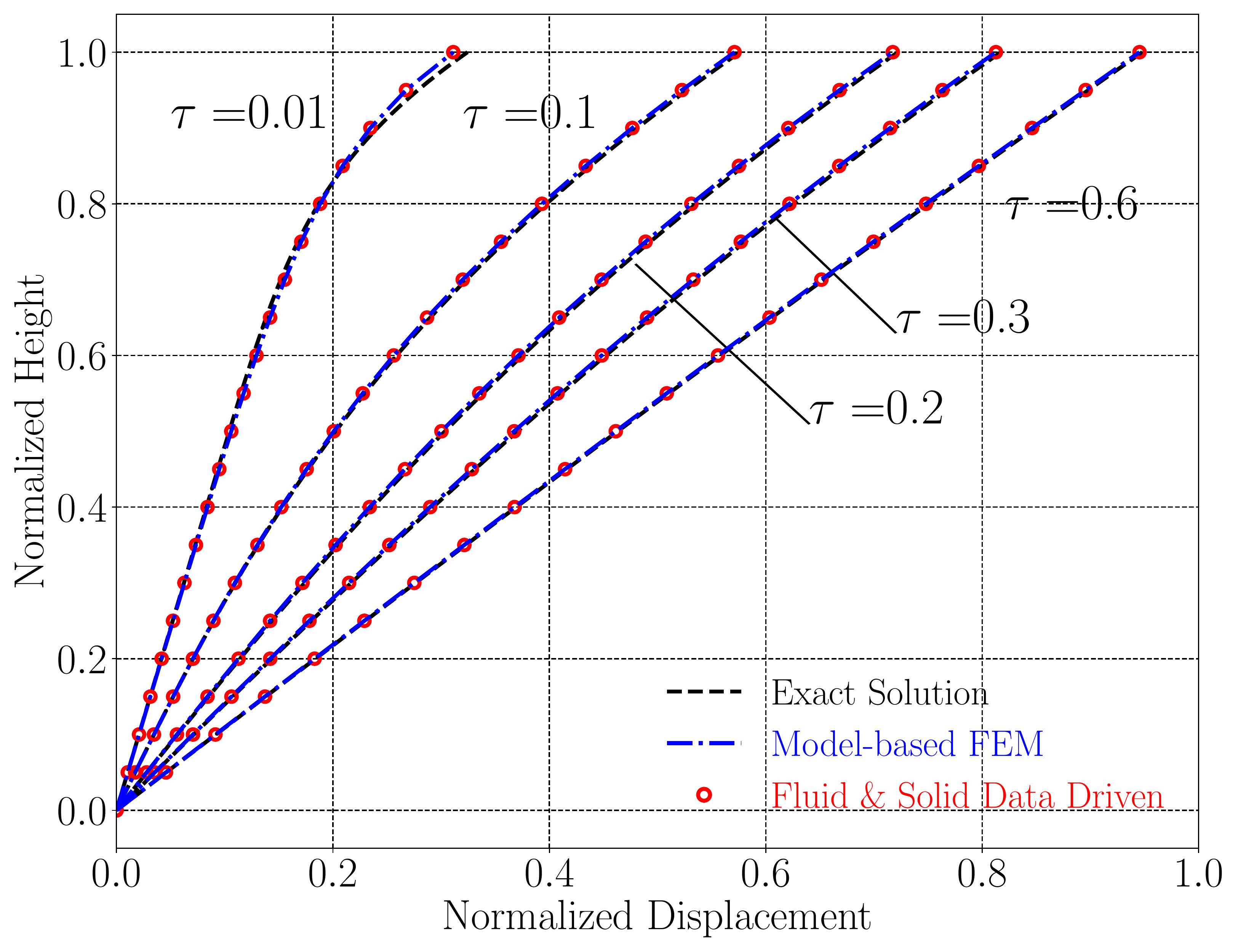}}
\hspace{0.01\textwidth}
  \caption{A comparison between results of the fully data-driven formulation, conventional model-based formulation, and exact solution. Normalized time is $\tau= t/t_{\mathrm{end}}$. 
   \label{fig::comp-pess-disp-fullyDD-FEM-exact}}
\end{figure}

\Fig \ref{fig::comp-pess-disp-fullyDD-FEM-exact} compares profiles of pressure and displacement along the sample height at various time steps. Two data sets each with $N=16385$ data points is used for the fully data-driven simulation shown by red circles in \fig \ref{fig::comp-pess-disp-fullyDD-FEM-exact}.
There is a good agreement between results obtained by the fully data-driven formulation and exact solutions; see \ref{appx::terzaghi} for analytical expressions. Moreover, the error between fully data-driven and conventional model-based FEM results is almost negligible. We observed similar trends for hybrid formulations as well which are not included in this paper for the sake of brevity.

\begin{figure}[h!]
 \centering
\includegraphics[width=0.3\textwidth]{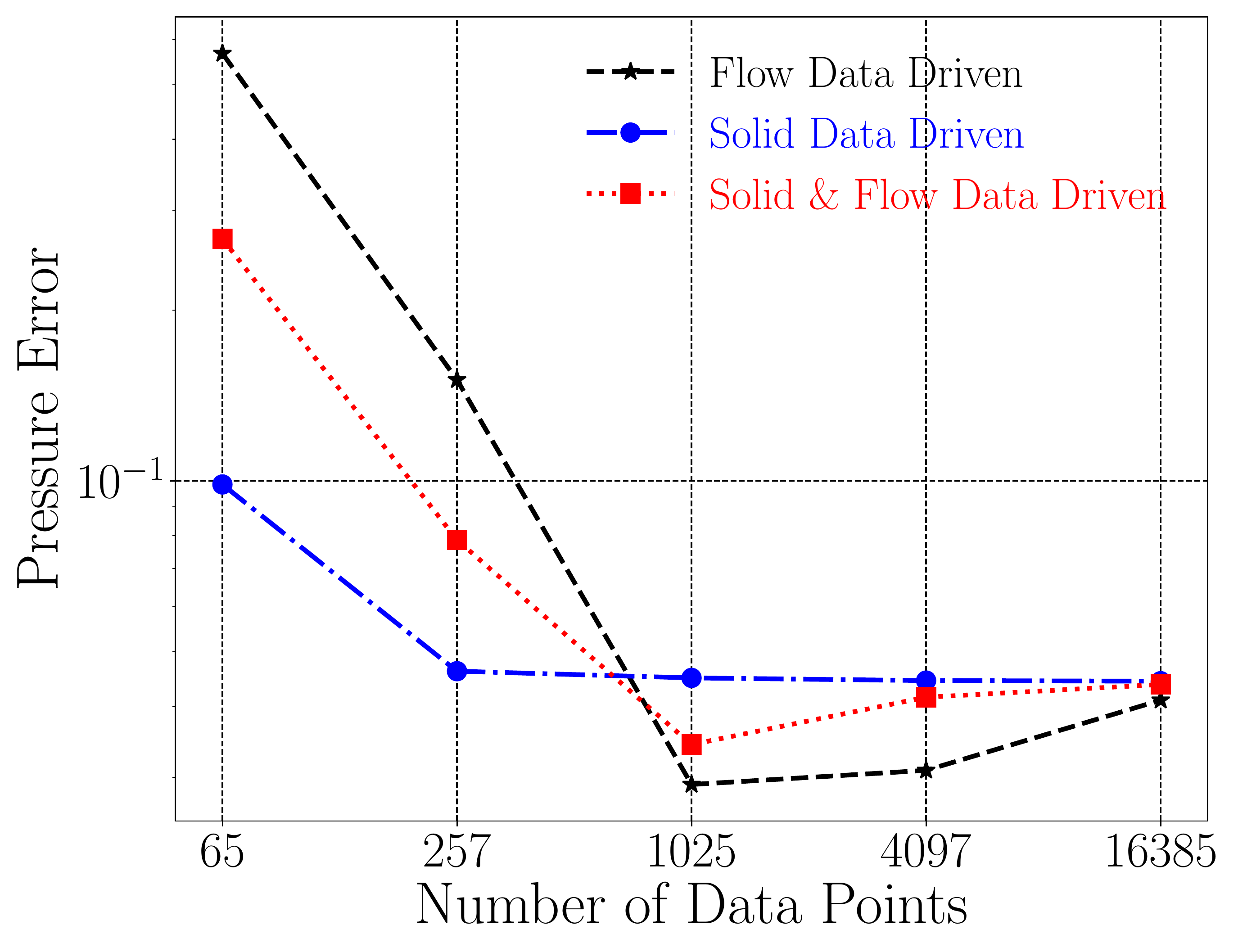}
\hspace{0.01\textwidth}
\includegraphics[width=0.3\textwidth]{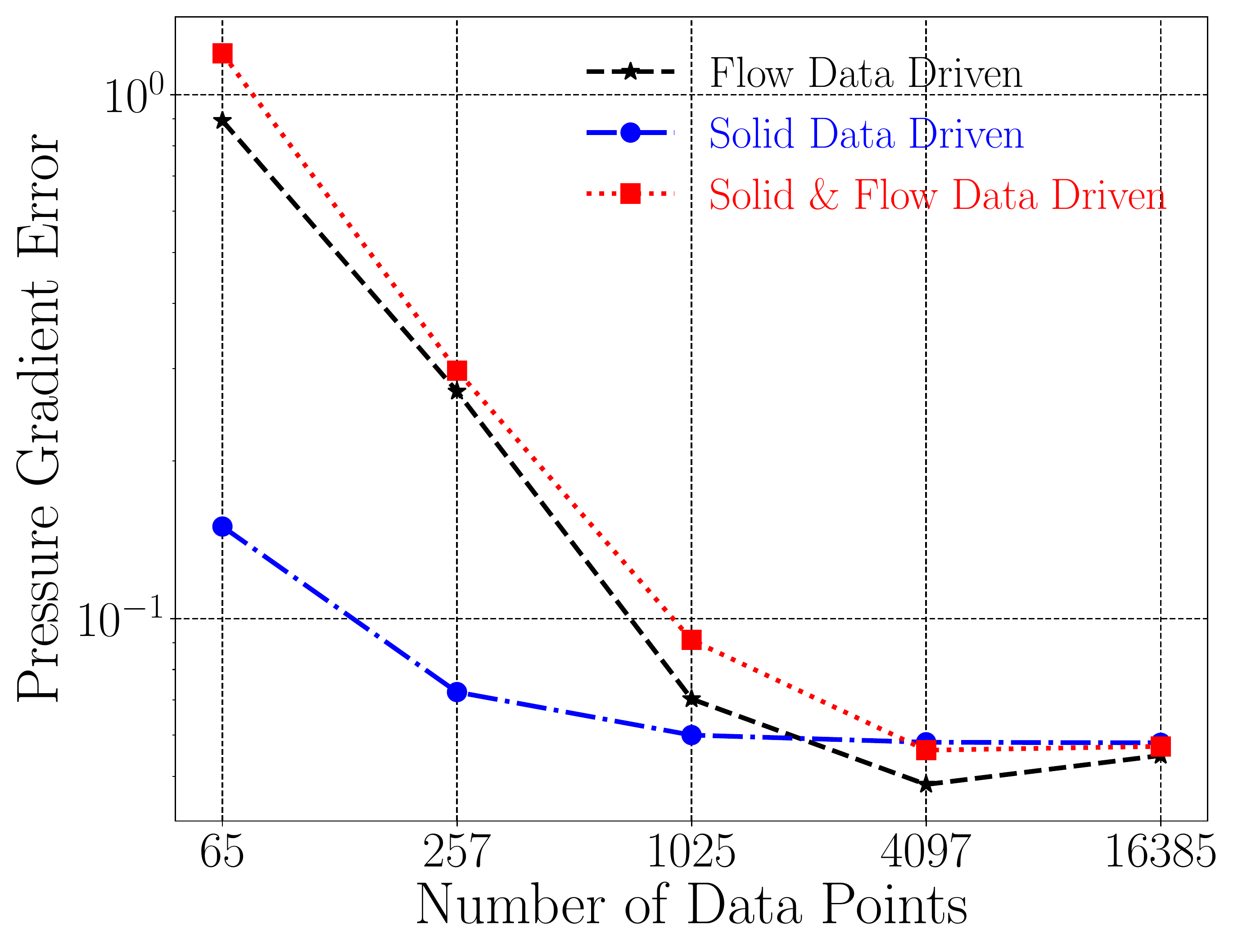}
\hspace{0.01\textwidth}
\includegraphics[width=0.3\textwidth]{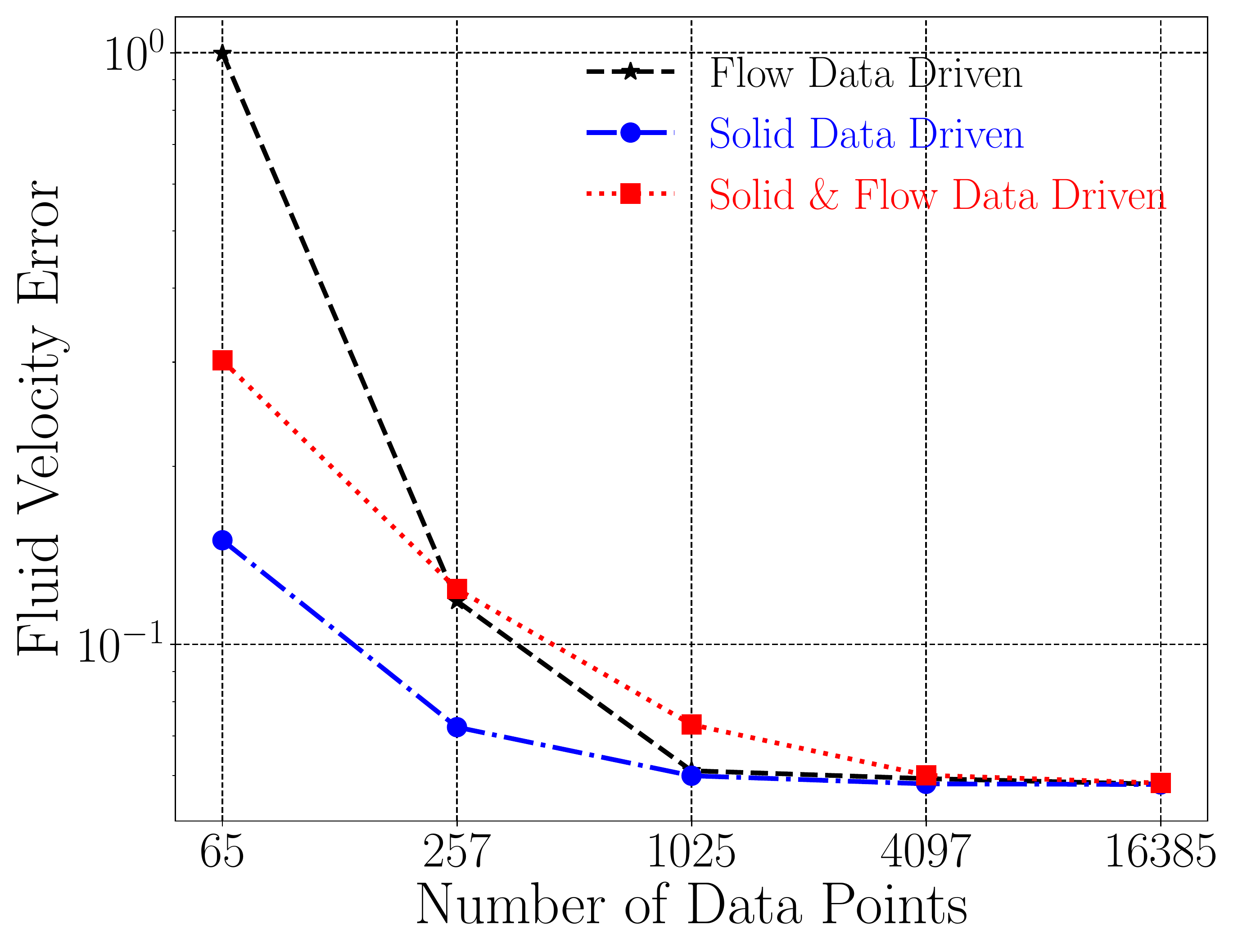}\\
\includegraphics[width=0.3\textwidth]{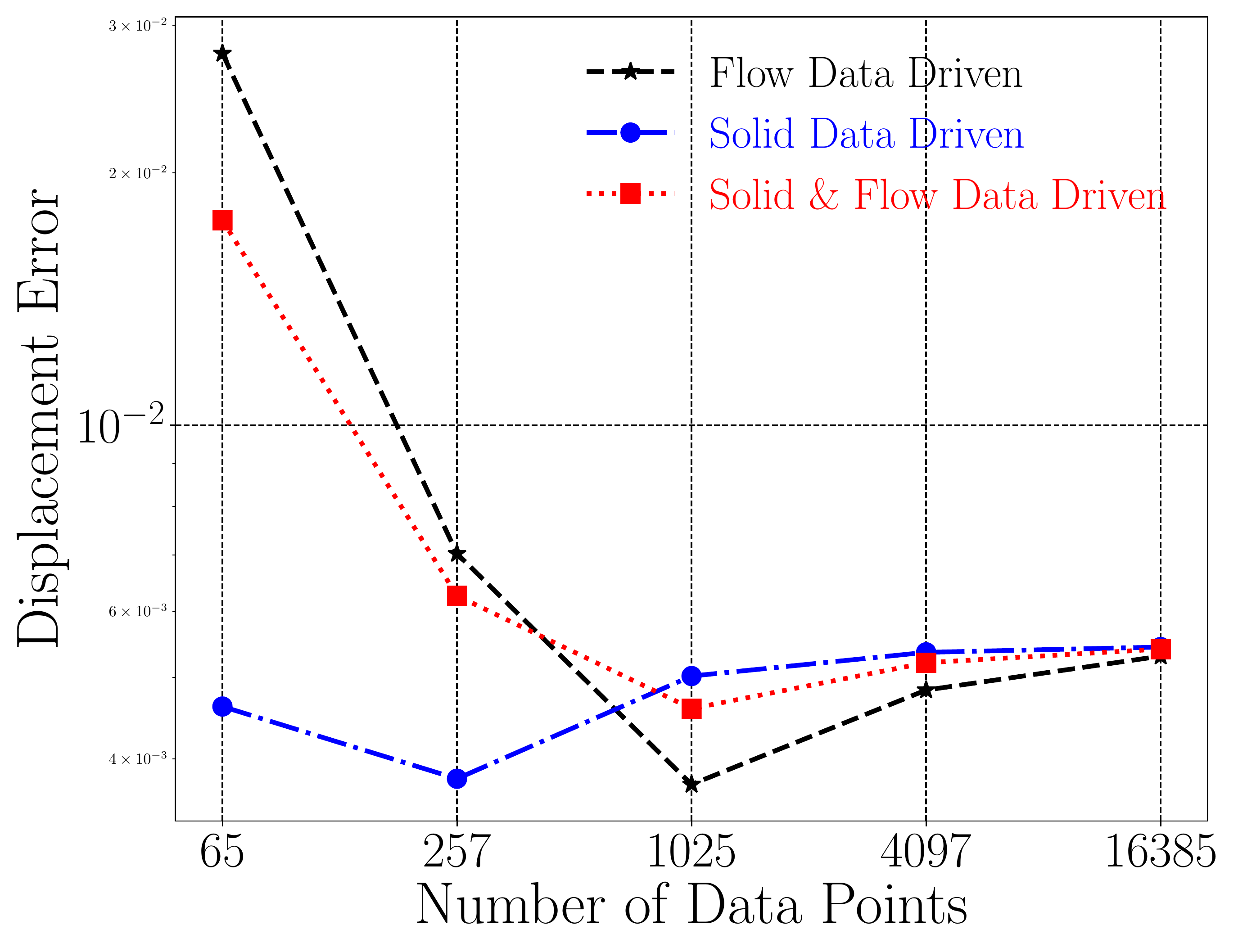}
\hspace{0.01\textwidth}
\includegraphics[width=0.3\textwidth]{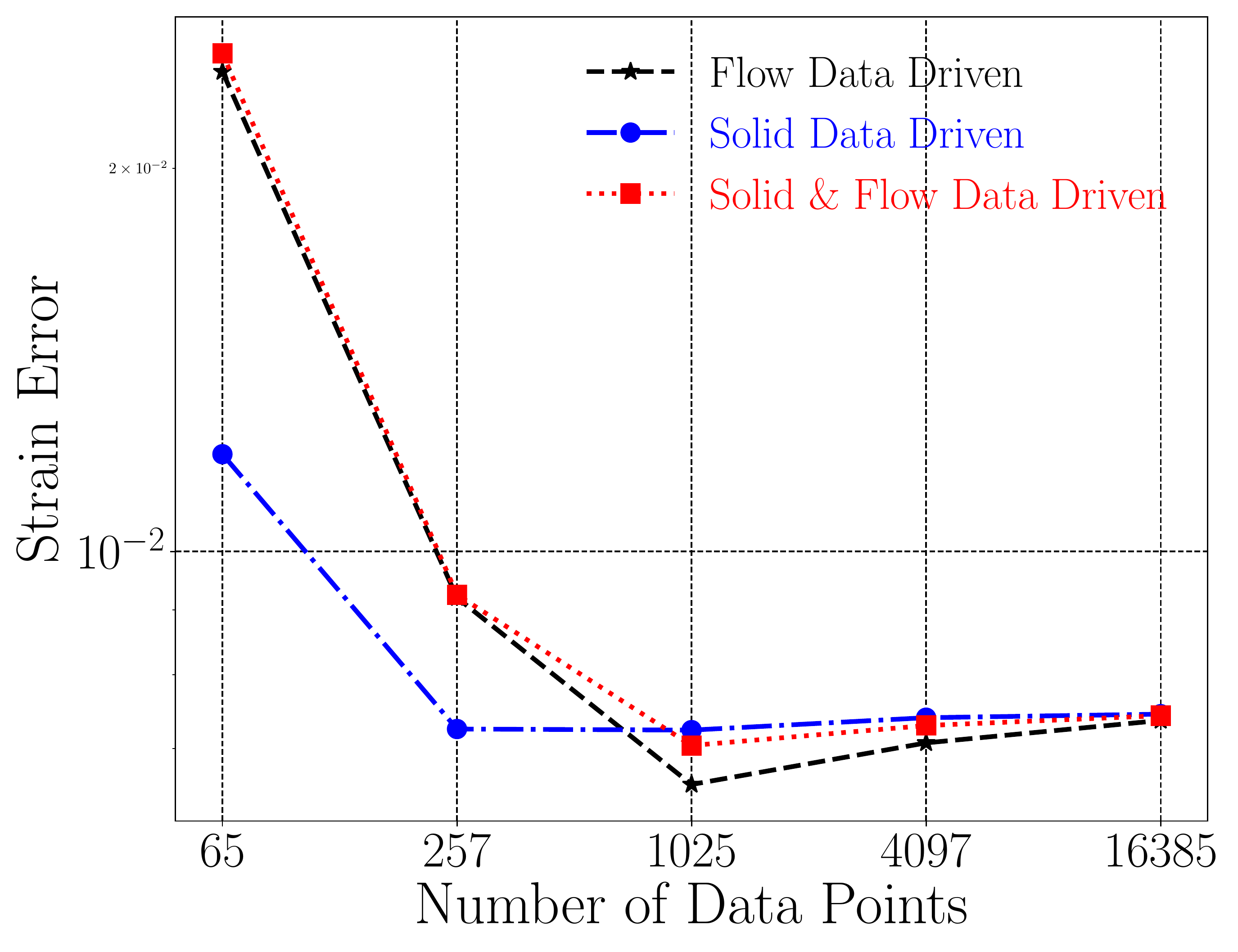}
\hspace{0.01\textwidth}
\includegraphics[width=0.3\textwidth]{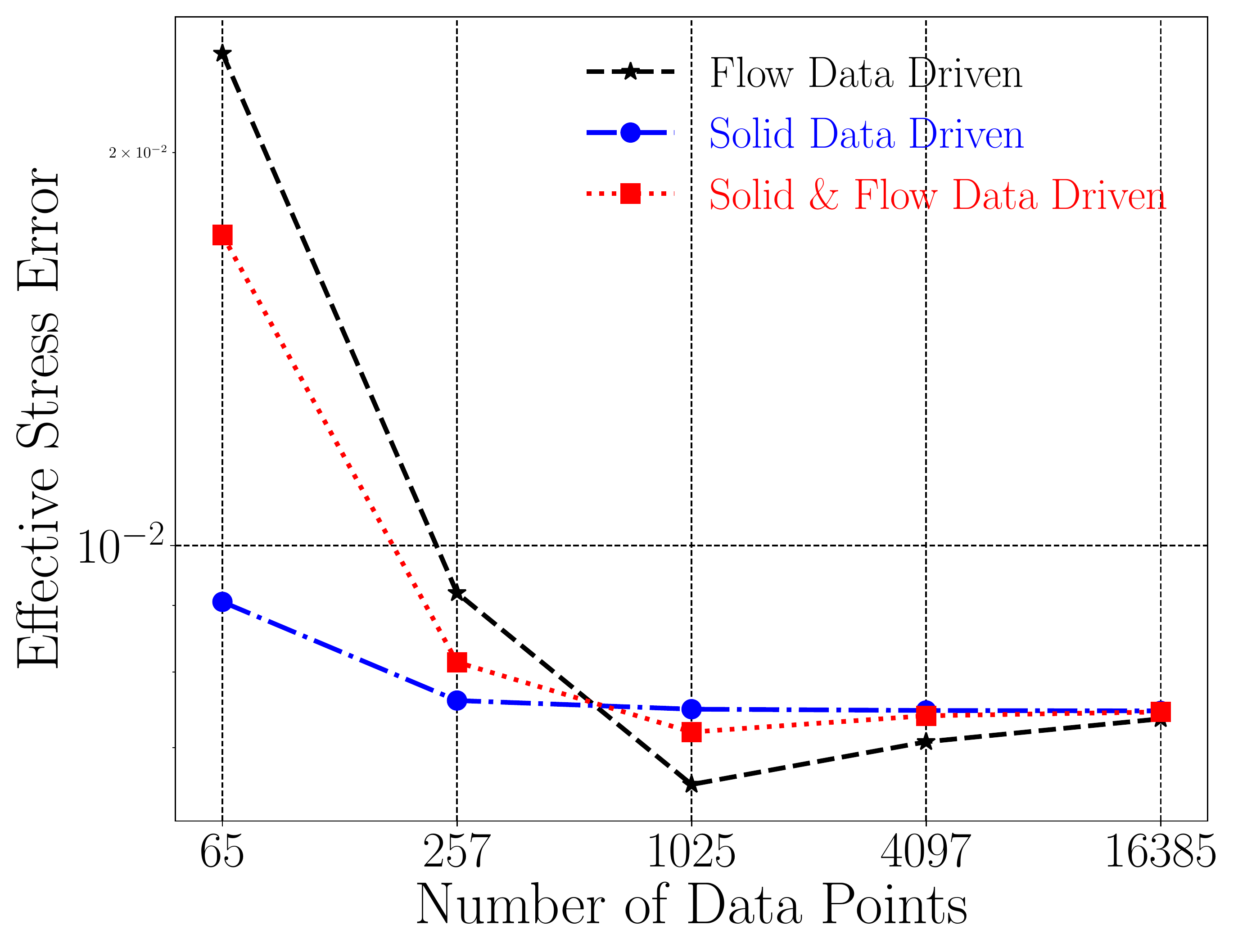}
  \caption{Total space-time error with respect to exact solution, see Eq. \eqref{eq::err-space-time}.
   \label{fig::terzaghi-spec-time-errExact-all}}
\end{figure}

Total space-time errors Eq. \eqref{eq::err-space-time} with respect to analytical solutions are plotted for different field quantities in \fig \ref{fig::terzaghi-spec-time-errExact-all}. Error is decreasing at initial stages of data refinement, but later it starts to grow with a milder slop. Eventually, errors for all the data-driven formulations converge to the same value for $N=16385$. Since there exist interacting sources of errors associated with spatial and temporal discretizations and data distribution, we cannot draw a precise conclusion about the error behavior at the intermediate stages. However, this observation suggests that all data-driven formulations converge to the same solution provided sufficient, large data.

\begin{figure}[h!]
 \centering
\includegraphics[width=0.3\textwidth]{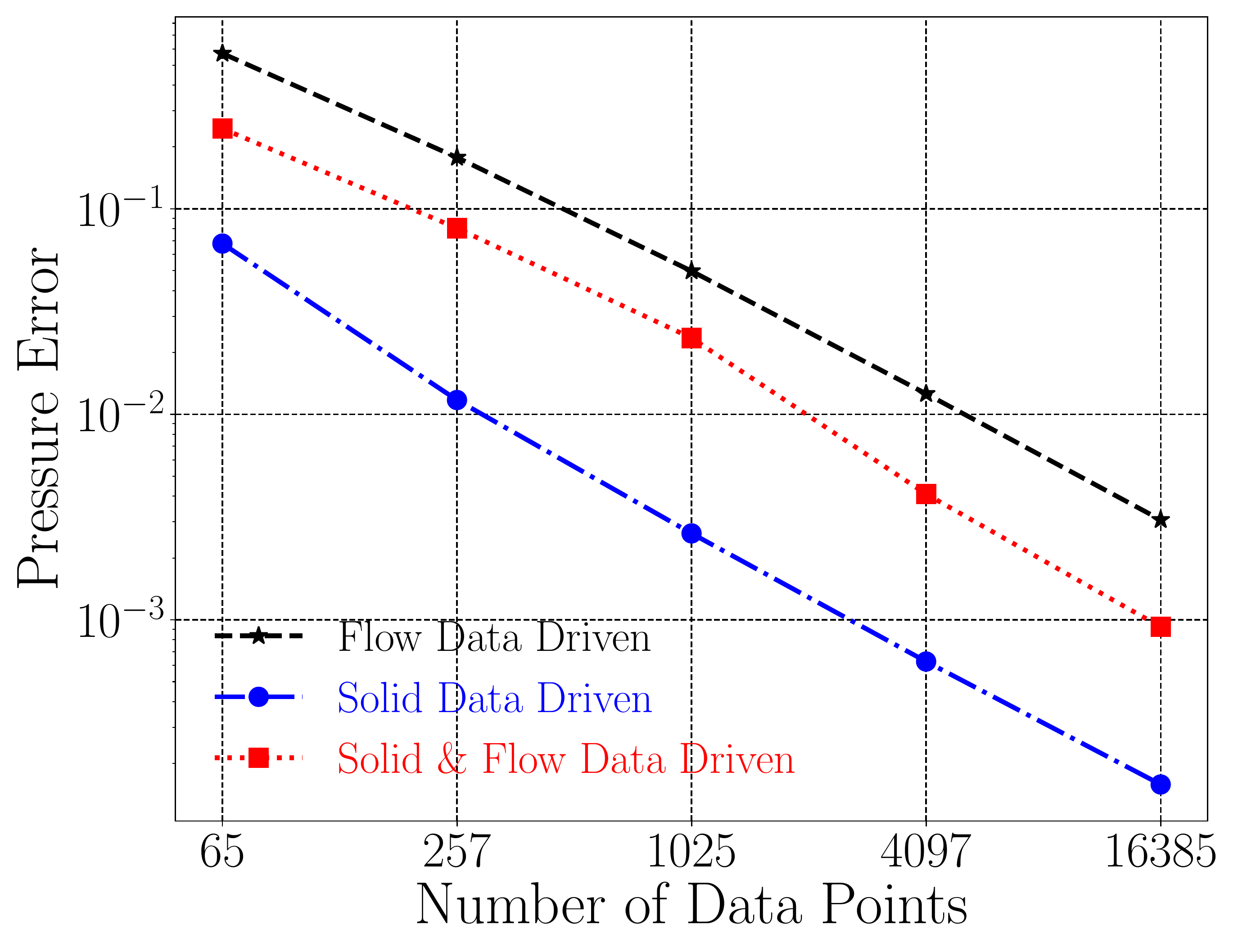}
\hspace{0.01\textwidth}
\includegraphics[width=0.3\textwidth]{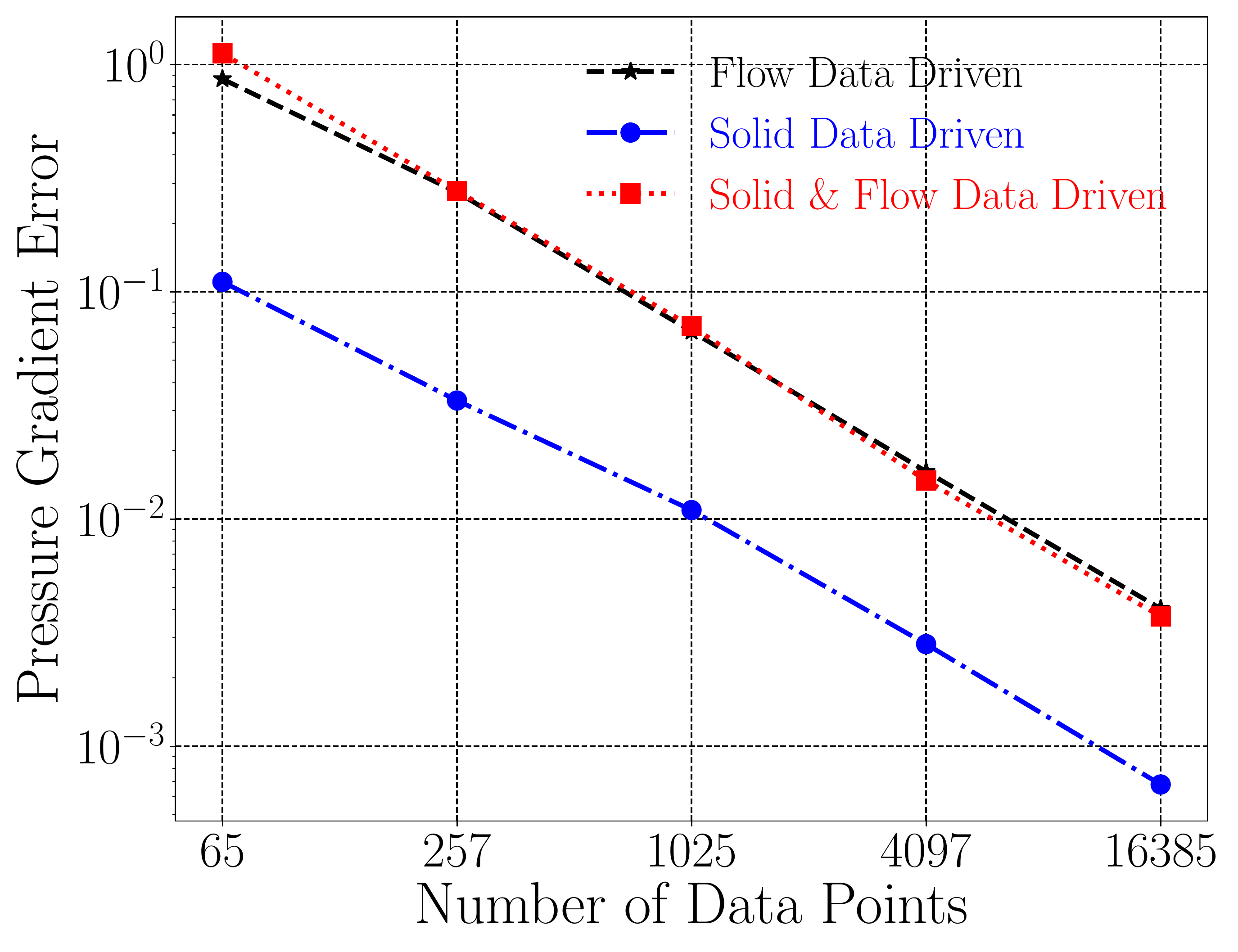}
\hspace{0.01\textwidth}
\includegraphics[width=0.3\textwidth]{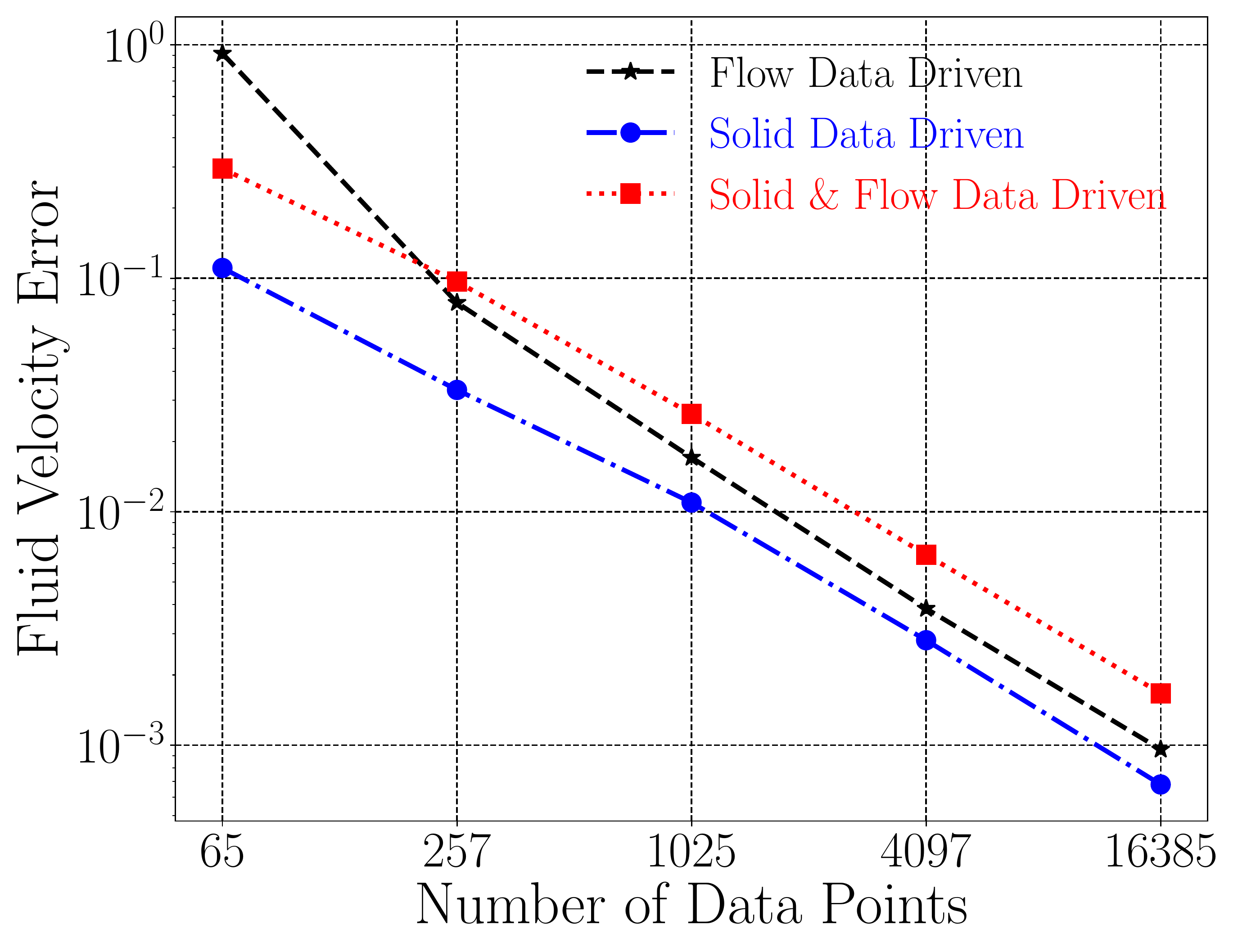}\\
\includegraphics[width=0.3\textwidth]{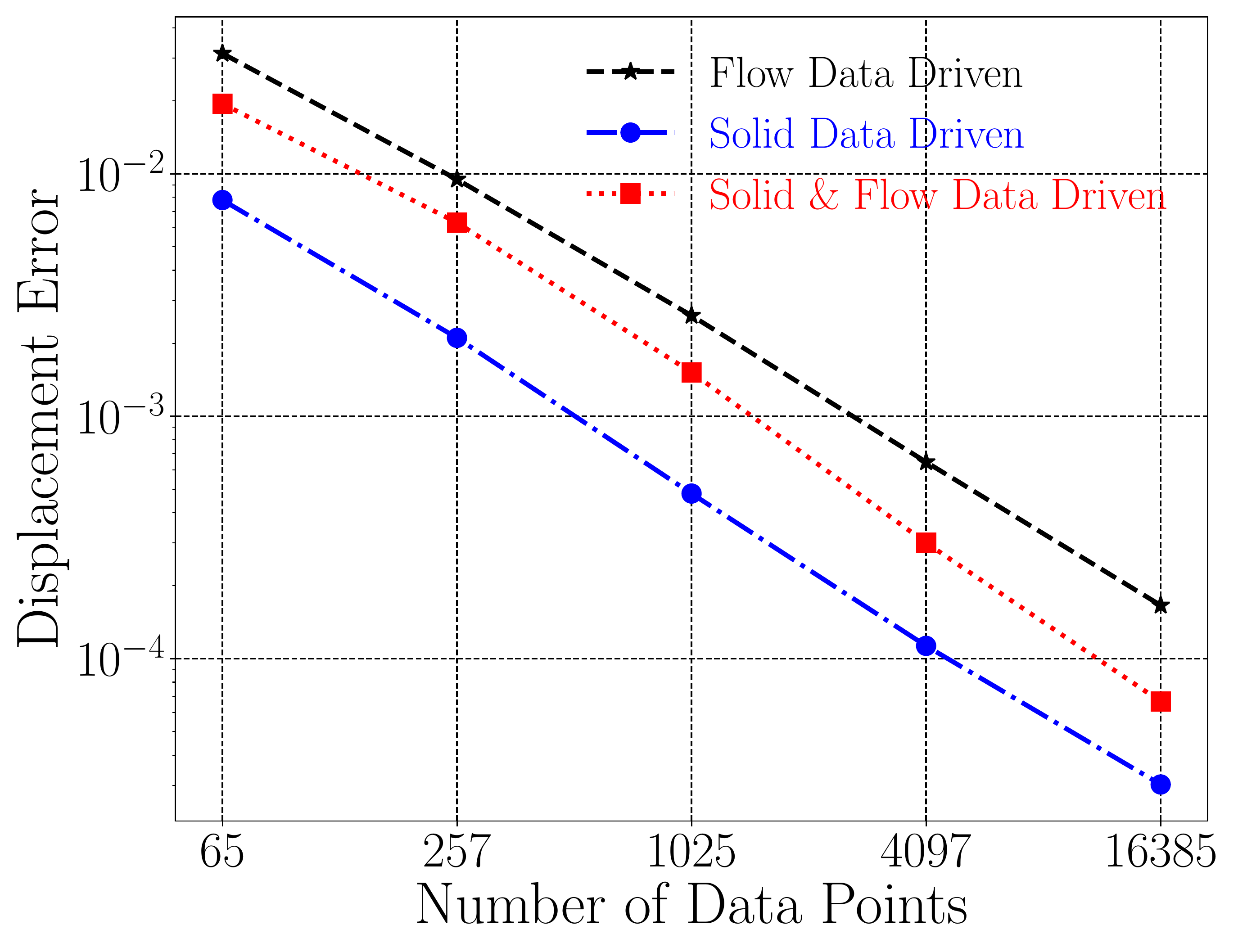}
\hspace{0.01\textwidth}
\includegraphics[width=0.3\textwidth]{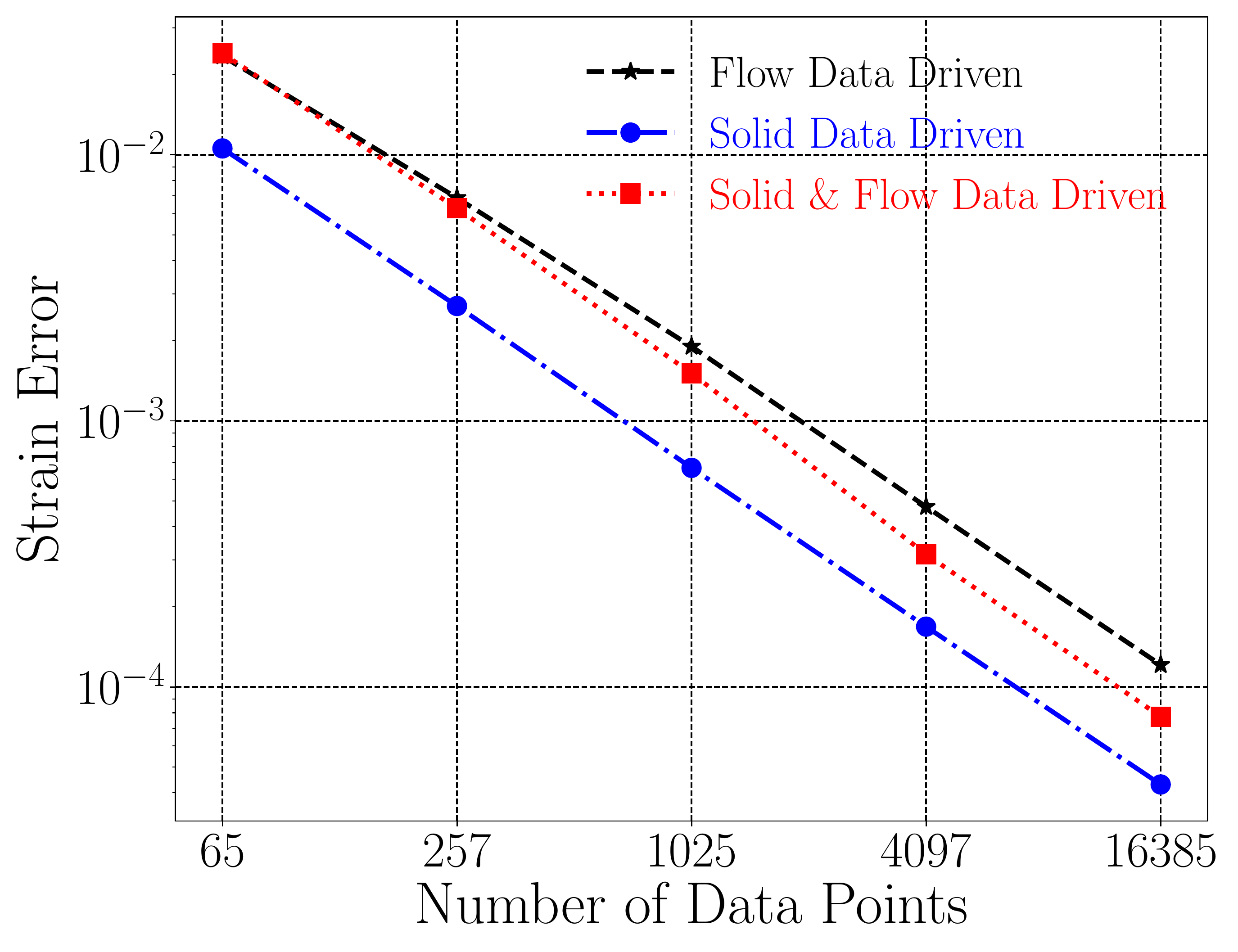}
\hspace{0.01\textwidth}
\includegraphics[width=0.3\textwidth]{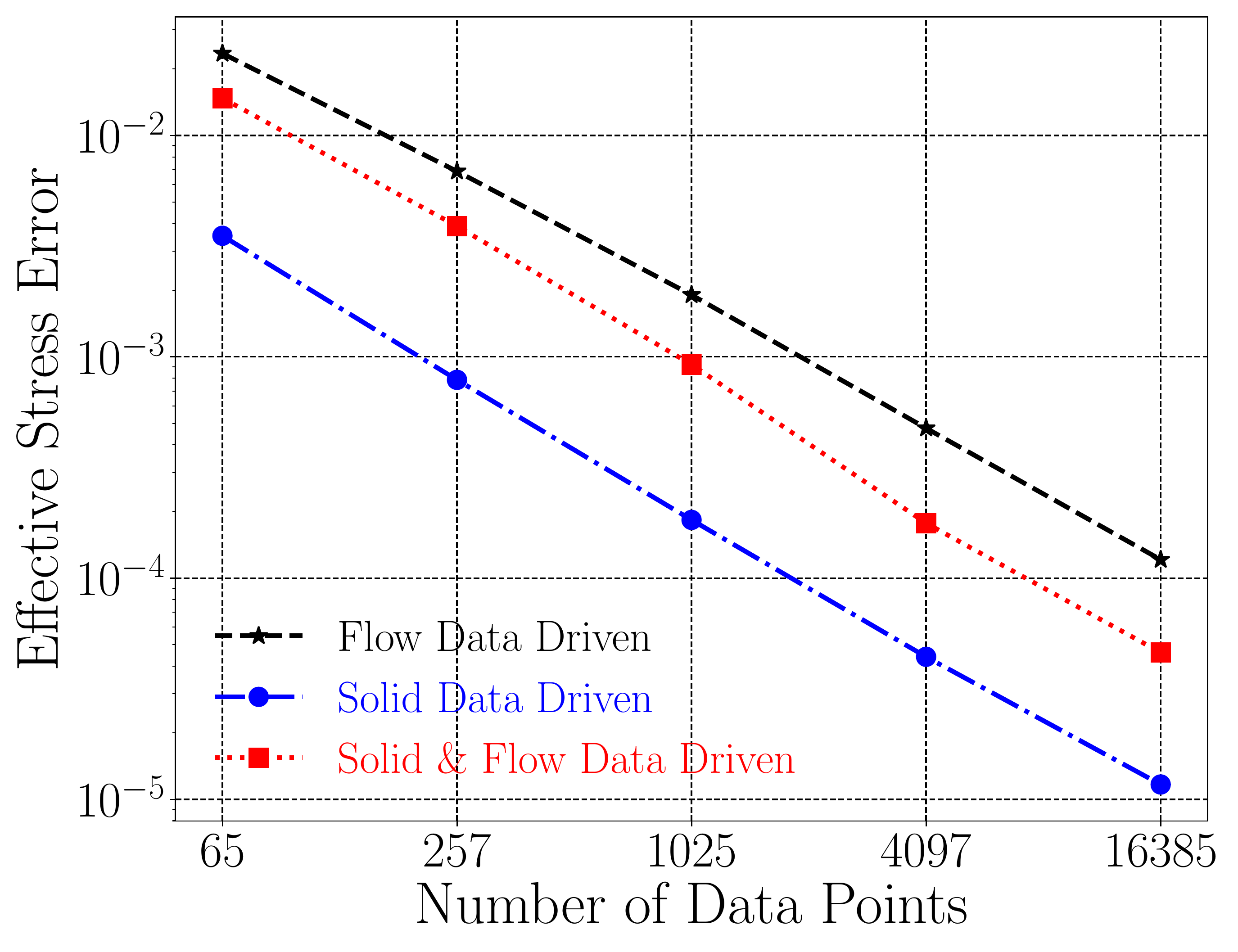}
  \caption{Total space-time error with respect to conventional model-based FEM solution, see Eq. \eqref{eq::err-space-time}.
   \label{fig::terzaghi-space-time-errorFEM-all}}
\end{figure}

\Fig \ref{fig::terzaghi-space-time-errorFEM-all} reports total space-time errors with respect to conventional model-based finite element solution. It is clear that increasing data intensity consistently decreases errors for all the data-driven formulations. In another word, data-driven formulations can converge to the model-based solution within the limit of infinite data availability. 

\begin{figure}[h]
 \centering
 \subfigure[]
{\includegraphics[width=0.45\textwidth]{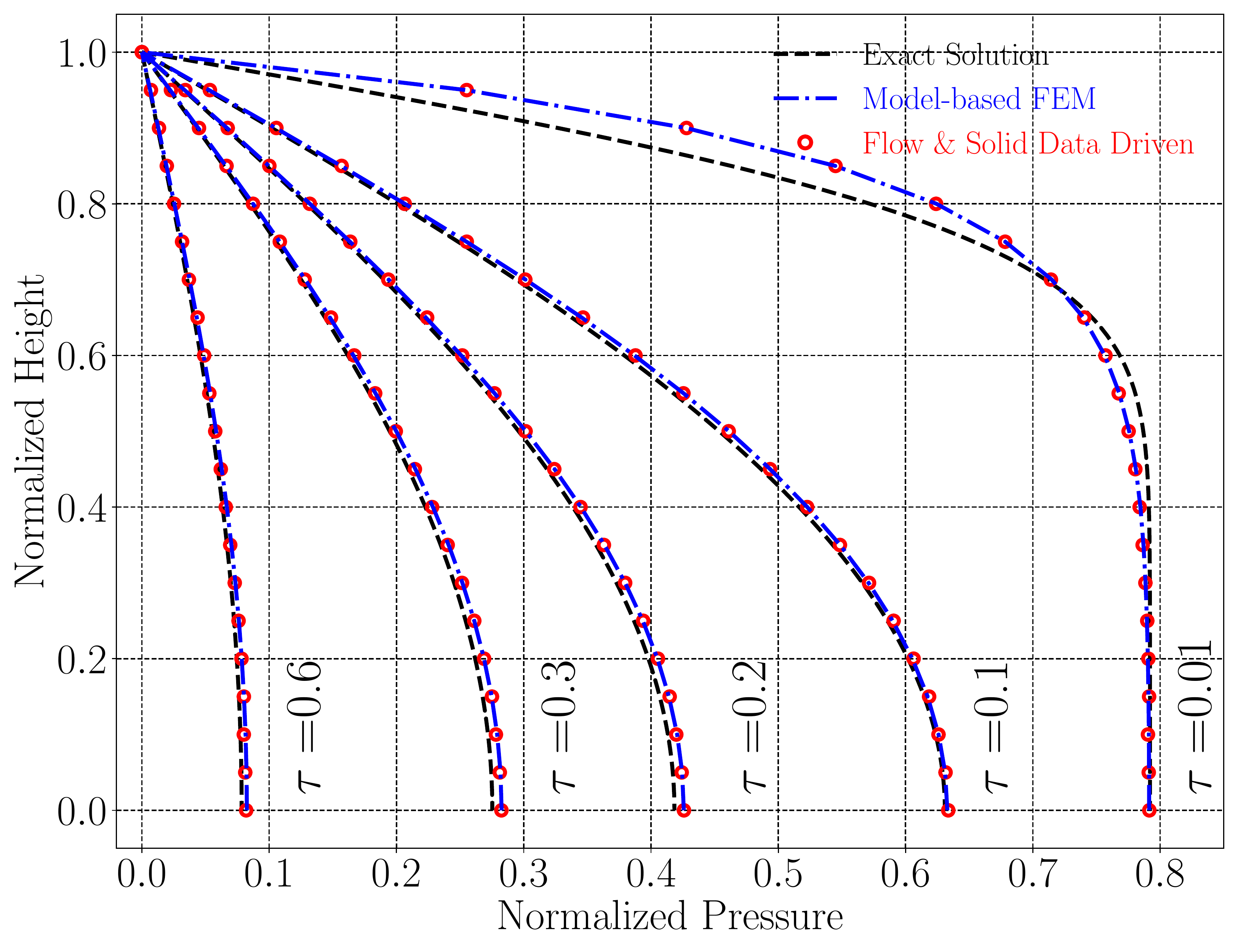}}
\hspace{0.01\textwidth}
 \subfigure[]
{\includegraphics[width=0.45\textwidth]{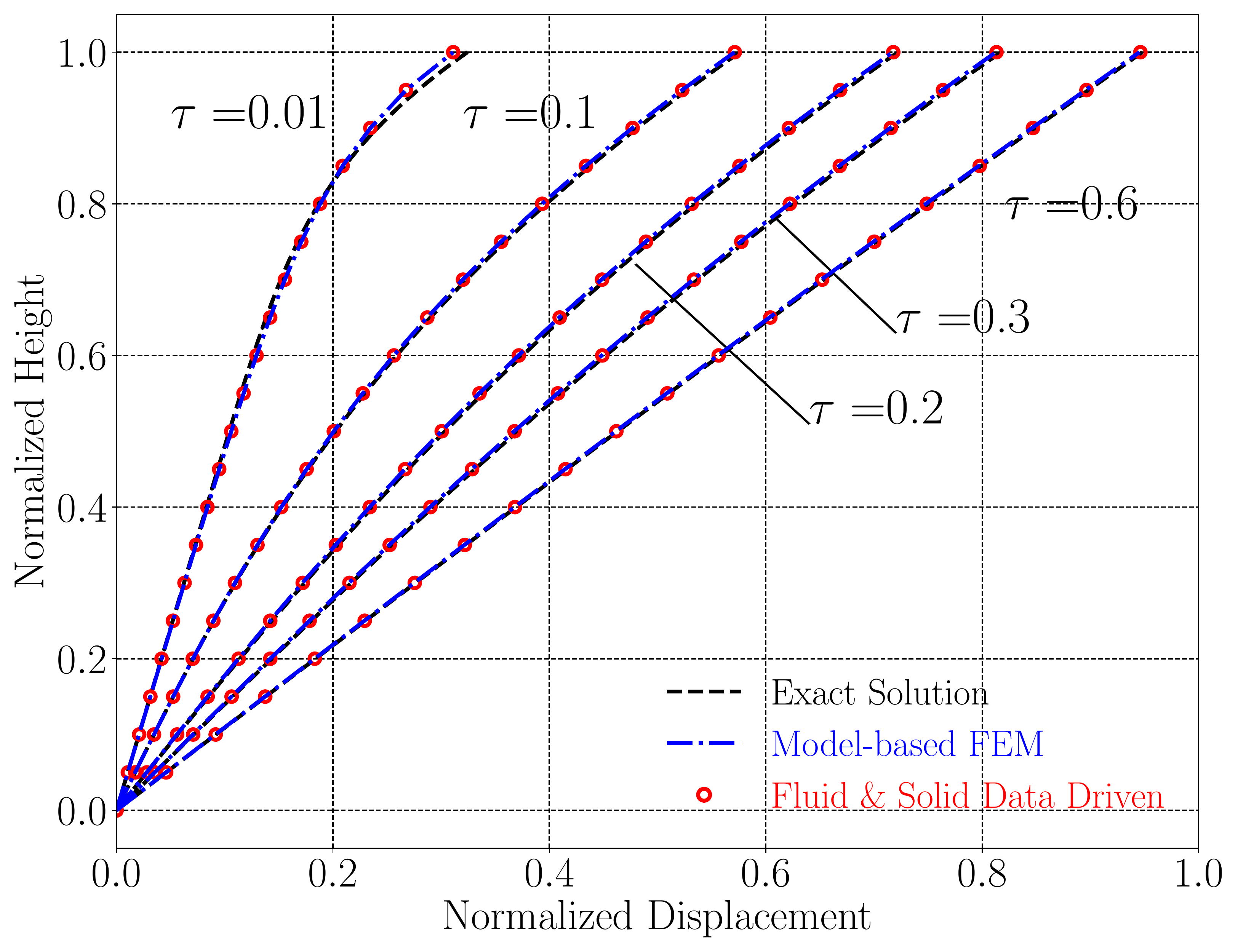}}
\hspace{0.01\textwidth}
  \caption{A comparison between results of the fully data-driven formulation, conventional model-based formulation, and exact solution with a multi-dimensional data set. Normalized time is $\tau= t/t_{\mathrm{end}}$. 
   \label{fig::comp-pess-disp-fullyDD-FEM-exact-multi-dim-data}}
\end{figure}

To make sure the accuracy is not limited to specialized data sets, we showcase the pressure and displacement profiles for the fully data-driven formulation with more general data sets in \fig \ref{fig::comp-pess-disp-fullyDD-FEM-exact-multi-dim-data}. In this figure, fluid data set is constructed by uniform discretization of data axis $\partial p/ \partial x$ with $65$ points in range $\left[-0.4, 0.4 \right] \mathrm{GPa}$ and axis $\partial p/ \partial y$ with $4096$ points in range $\left[-8.6, 4.3 \right] \mathrm{GPa}$. In solid data set $\epsilon_{xx}$, $\epsilon_{yy}$, and $\epsilon_{xy}$ are uniformly sampled by $65$, $4096$, and $65$ points within intervals $\left[-0.002, 0.002 \right]$,  $\left[-0.026, 0.013 \right]$, and $\left[-0.001, 0.001 \right]$, respectively. In the initial fixed-point iteration, data points are randomly assigned, in contrast to uniform assignment in previous cases. Moreover, numerical parameters for fluid and solid metrics are set to $2.45 \times 10^{-6} \tensor{I} \ \mathrm{m^2/Pa.s}$ and plane strain elasticity tensor with Poisson's ratio $0.1$ and Young's modulus $56 \ \mathrm{GPa}$ respectively, as opposed to previous cases where they were chosen to their potential optimal values. The reason we changed the previous set-ups is to show the robustness of formulation in more general settings. As it is clear form this figure, the results are in a good agreement with the ground truths, similar to previous cases.

\subsection{Verification exercise 2: Stress Relaxation}
In this problem, we validate formulations and implementation for three-dimensional elements. Also, we will study the convergence behavior of the metric minimization part of the algorithm.

\begin{figure}[h]
 \centering
\includegraphics[width=0.3\textwidth]{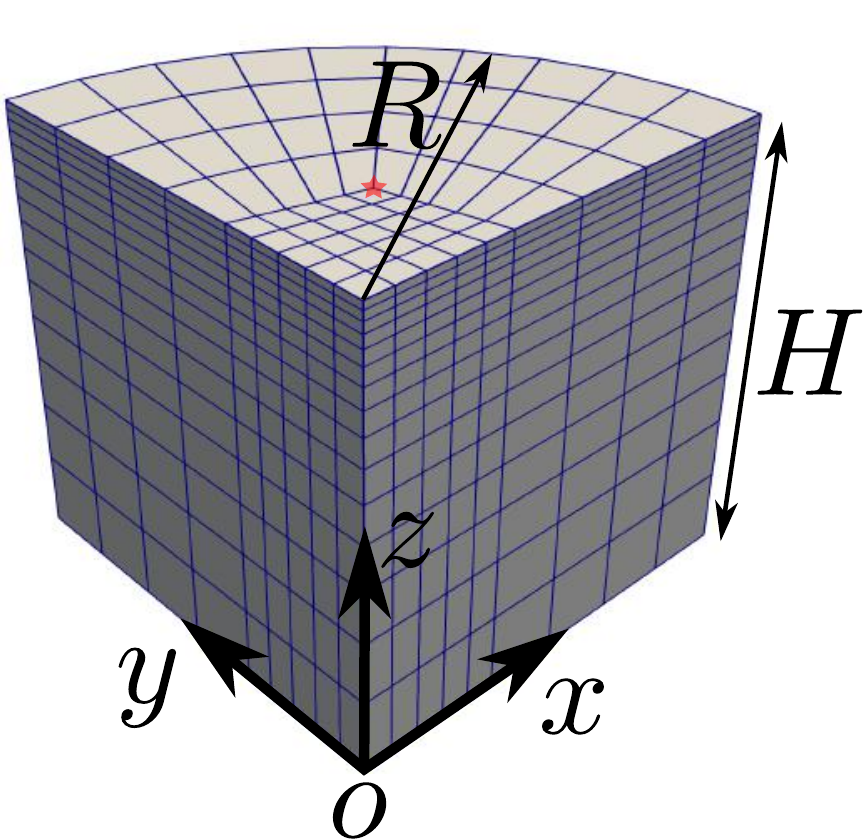}
  \caption{Geometry and mesh for stress relaxation problem with $R=H=10 \mathrm{m}$. Red star point with coordinates $(7.89, 7.89, 10)\mathrm{m}$ shows an observation location for plotting purposes. \label{fig::strs_relax-geom-mesh}}
\end{figure}

\Fig \ref{fig::strs_relax-geom-mesh} depicts geometry and mesh used herein. All external faces of the cylinder are impermeable except the top face. The only free displacement degrees of freedom for boundary faces are in the $z$ direction except the bottom boundary which is completely clamped from movement. Pressure is kept zero at the top boundary, and a compressive displacement with constant rate $\dot{\bar{u}}_z = -0.005 \mathrm{m} / \mathrm{s}$ is applied for $t_{\mathrm{ramp}} = 2s$ and remained constant towards the simulation end $t_{\mathrm{end}} = 10\mathrm{s}$. This problem is known as stress relaxation test which is another benchmark for validation of numerical codes \citep{haider2007application, Sun.Ostien.Salinger:2013}. Material properties are listed in Table \ref{tab:strs-relax-params} which are used for the data generation or conventional model-based FEM. The system is in rest at $t=0$, and time increment is set to $\Delta t = 0.1 \mathrm{s}$. We use trilinear Lagrangian basis functions to discretize all unknown field. Numerical integration is performed by the 8-points Gaussian quadrature rule.

\begin{table}[h]
\centering
\caption{Material parameters for  stress relaxation problem \label{tab:strs-relax-params}}
\small%
\begin{tabular}{lll}
\hline 
Physical parameter & Unit & Value\tabularnewline
\hline 
\hline 
Young's modulus ($E$) & GPa & $100$\tabularnewline
Poisson's ratio ($\nu$) & - & $0.25$\tabularnewline
Intrinsic permeability ($k$) & $\mathrm{m}^2$ & $8.33\times10^{-8}$\tabularnewline
Fluid dynamics viscosity ($\mu$) & Pa.s & $0.001$\tabularnewline
Biot coefficient ($B$) & - & $1$\tabularnewline
Biot modulus  ($M$) & GPa & $2\times10^{10}$\tabularnewline
\hline 
\end{tabular}
\end{table}

We generate fluid data set using equidistant points sampled from Darcy's law with $\partial p / \partial x= \partial p / \partial y =0$ and $-58\  \mathrm{MPa} \le  \partial p / \partial z  \le 0.0005\  \mathrm{MPa} $. The only nonzero component of strain in solid data set is $-0.0024 \le \epsilon_{zz} \le 0$, and data is sampled from Hooke's law. Again, we have used our prior knowledge about the problem's one-dimensional characteristic to reduce the amount of data needed to complete the simulation. For the following results, $1000$ data points are used in fluid data-driven and solid data-driven simulations. The fully data-driven simulation has two separate databases with $1000$ data points in each for solid and fluid parts. Numerical parameters for fluid and solid metric functions are set to $6.66 \times 10^{-5} \tensor{I} \ \mathrm{[m^2/Pa.s]}$ and elasticity tensor with elastic modulus $80\ \mathrm{[GPa]}$ and Poisson's ratio $0.225$, respectively.

\begin{figure}[h]
 \centering
{\includegraphics[width=0.4\textwidth]{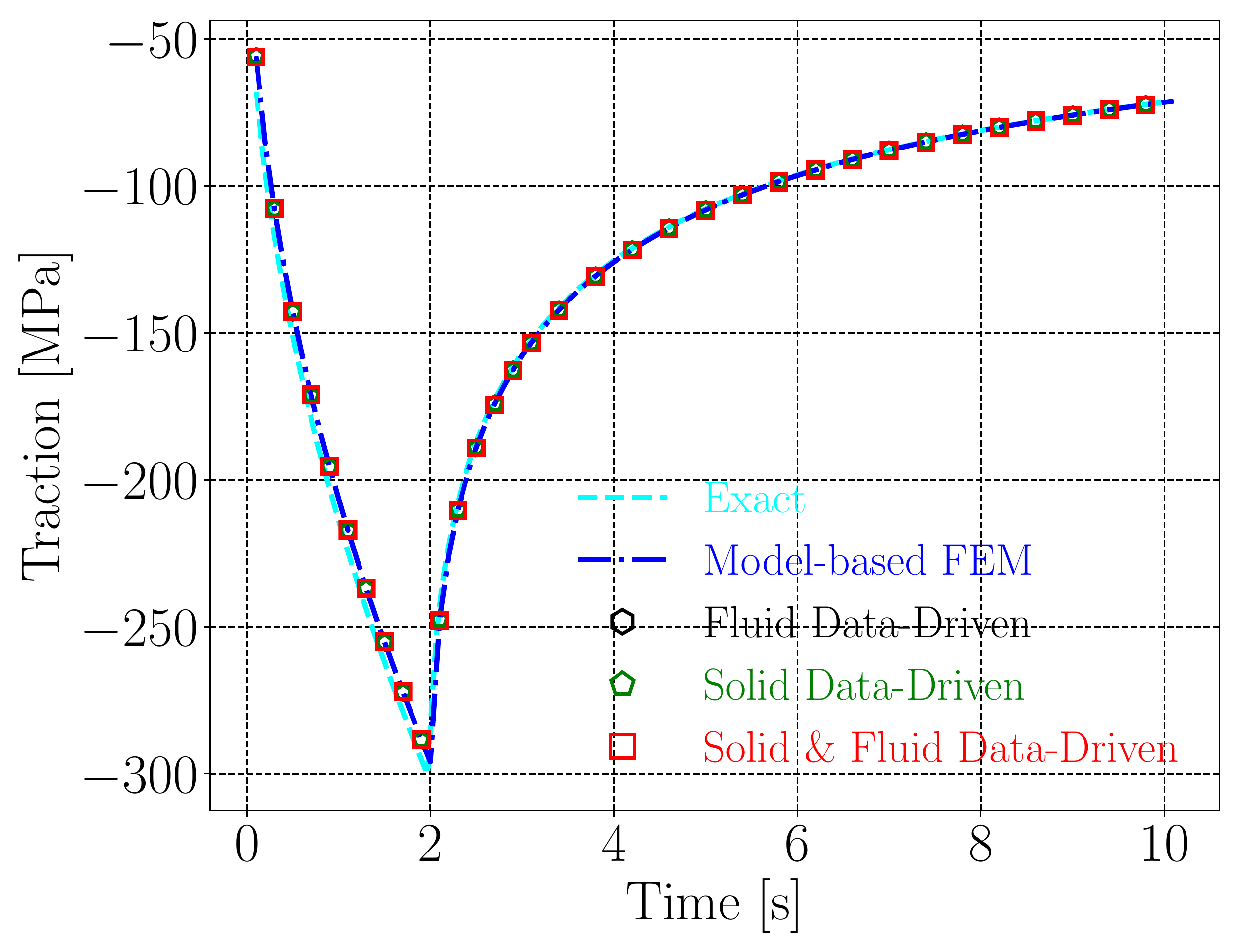}}
  \caption{Normal traction history for different data-driven schemes at the observation point shown in \fig \ref{fig::strs_relax-geom-mesh}.
   \label{fig::strs-relax-top-strs-height-allDD}}
\end{figure}

\begin{figure}[h]
 \centering
{\includegraphics[width=0.4\textwidth]{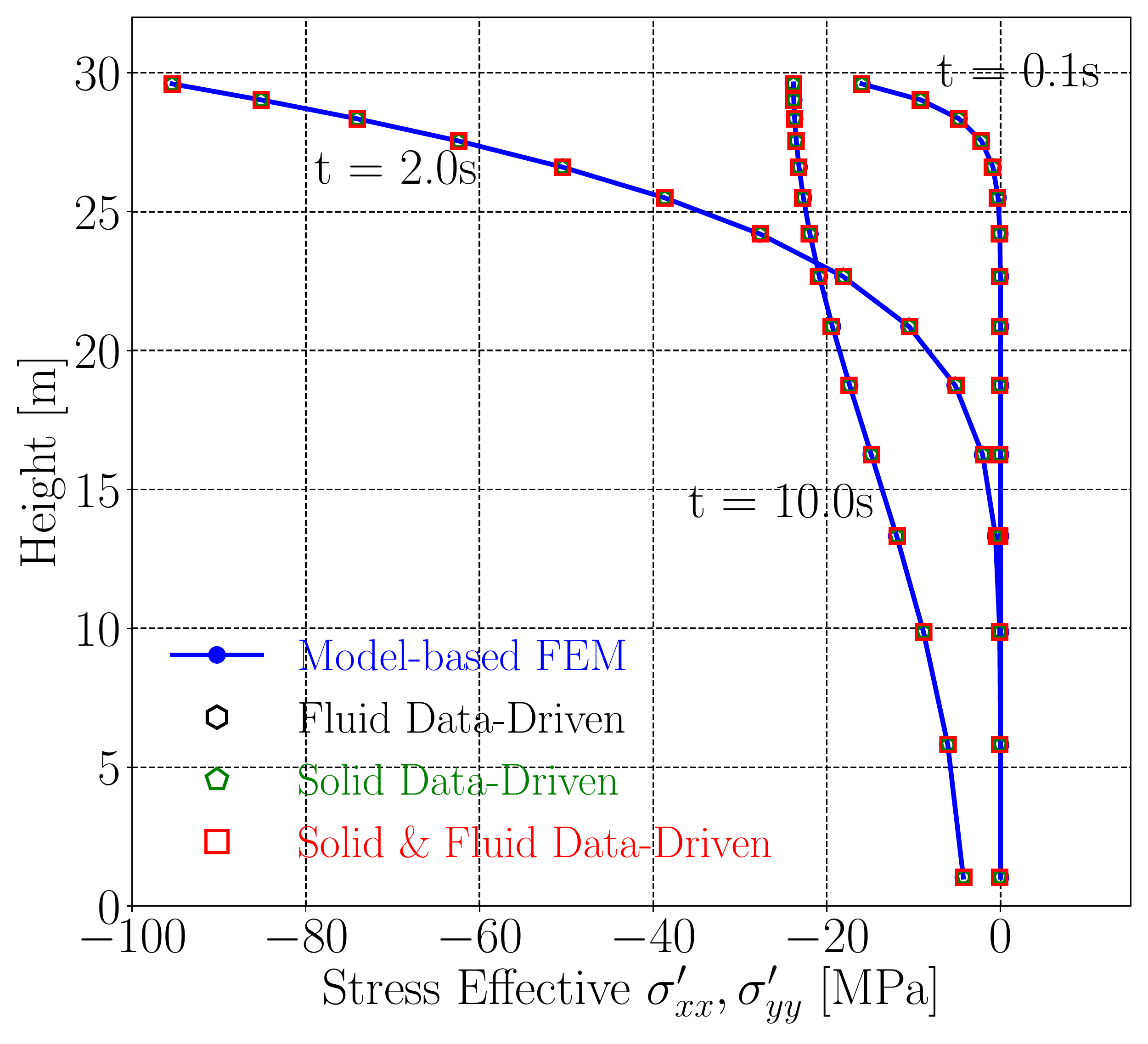}}
  \caption{Lateral effective stress profile along the sample height at different times.
   \label{fig::strs-relax-strsEff-height-allDD}}
\end{figure}

\begin{figure}[h]
 \centering
{\includegraphics[width=0.4\textwidth]{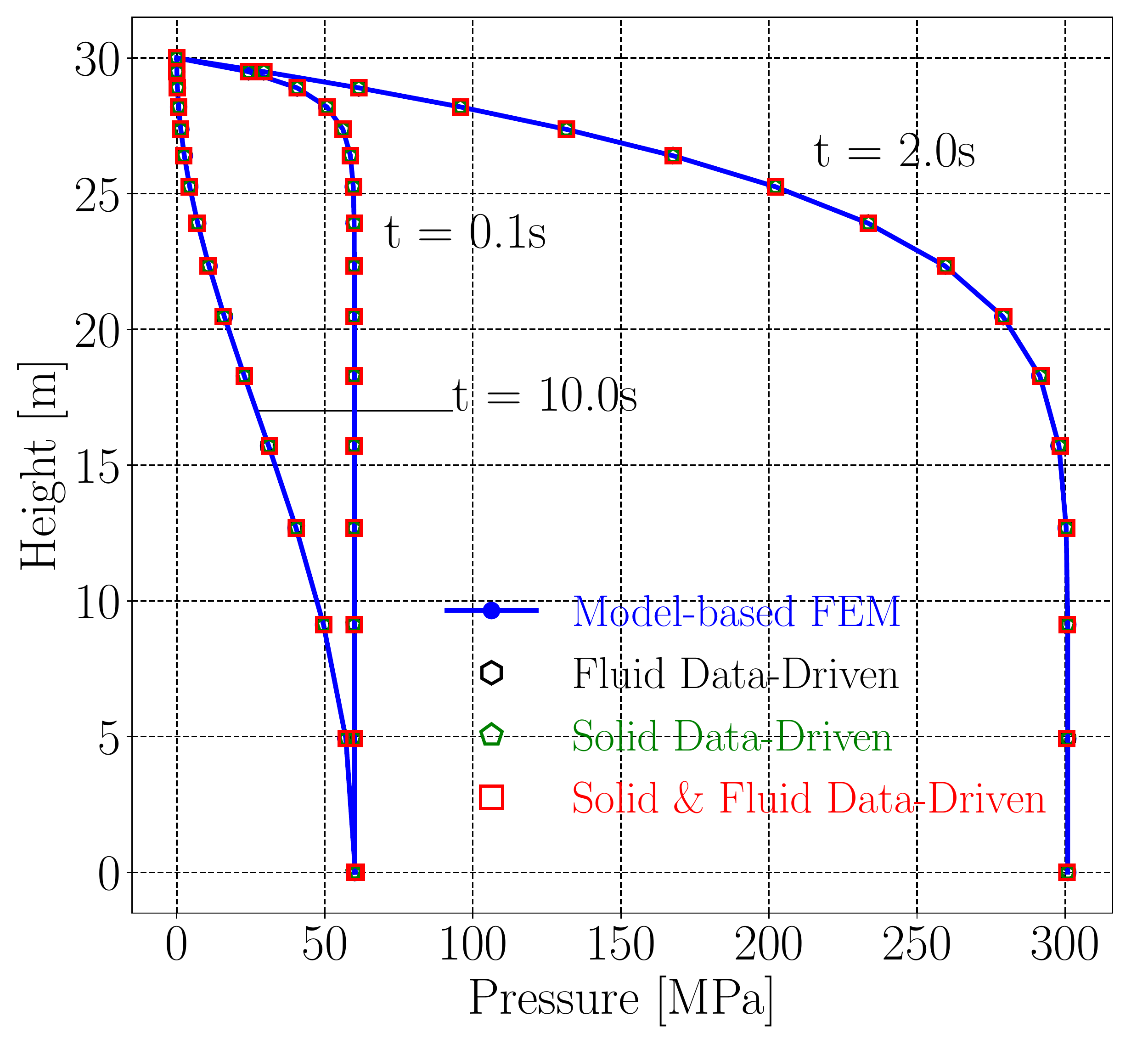}}
  \caption{Pressure profile along the sample height at different times.
   \label{fig::strs-relax-press-height-allDD}}
\end{figure}

As illustrated in \fig \ref{fig::strs-relax-top-strs-height-allDD}, traction history at the top nodal point, colored with red star in \fig \ref{fig::strs_relax-geom-mesh}, is in good agreement with the analytical solution (see Appendix \ref{appx:strs-relax}) for all three data-driven formulations. Additionally, \fig \ref{fig::strs-relax-strsEff-height-allDD} confirms a satisfactory match for the lateral effective stress between model-based response and data-driven formulations. Although there is no constitutive relation in solid and fully data-driven schemes to impose the Poisson effect this phenomenon is captured correctly even with a small database consists of 1000 data points. Note that this observation reveals the capacity of the proposed framework in transferring hidden information from data to modeling without human intervention. \Fig \ref{fig::strs-relax-press-height-allDD} compares pressure predictions at different times along the height of cylinder which indicates solution accuracy for the continuity equation.

\begin{figure}[h]
 \centering
{\includegraphics[width=0.4\textwidth]{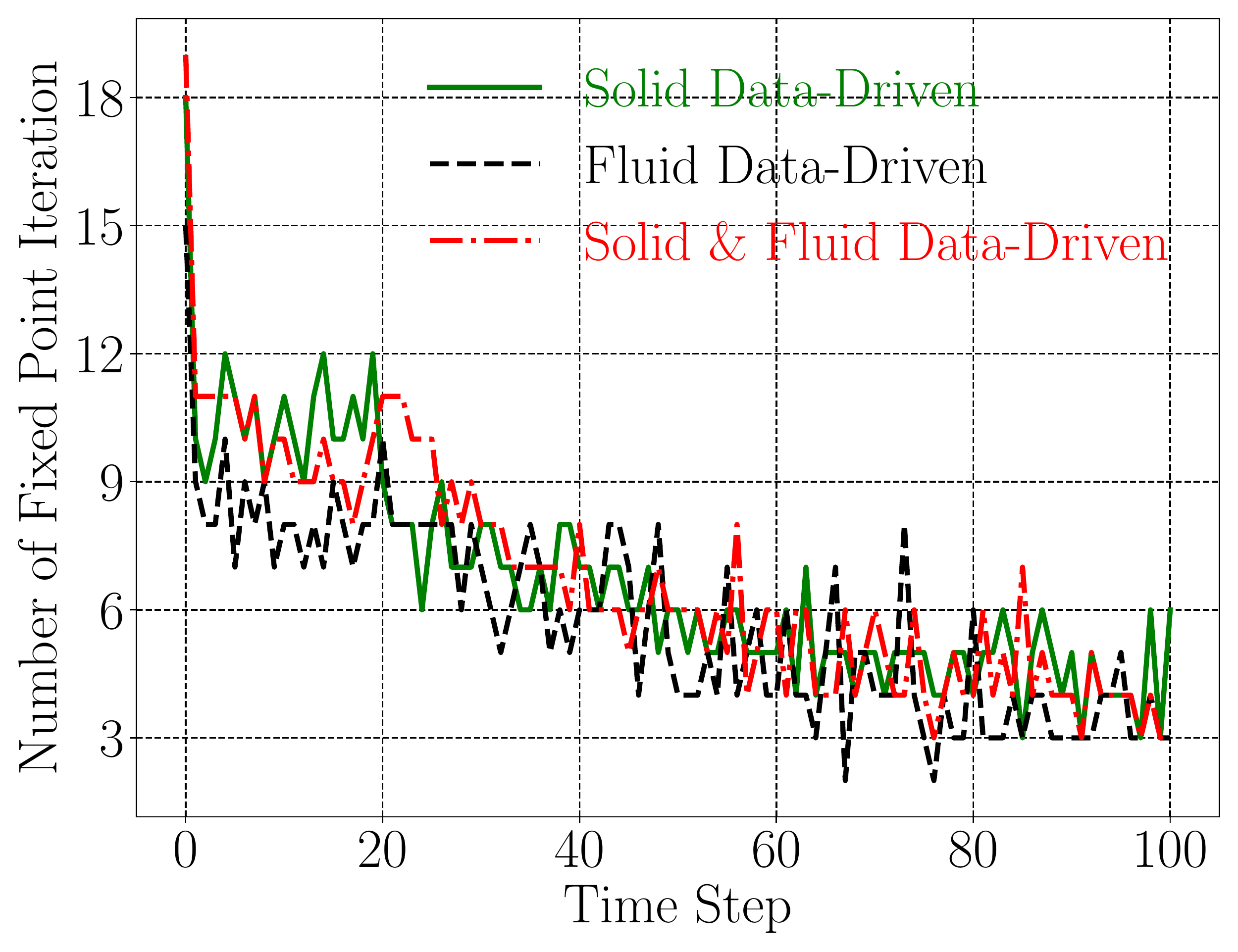}}
  \caption{Number of fixed-point iterations at each time step for all data-driven formulations.
   \label{fig::strs-relax-numFixItr-allDD}}
\end{figure}

\begin{figure}[h]
 \centering
 \subfigure[]
{\includegraphics[width=0.44\textwidth]{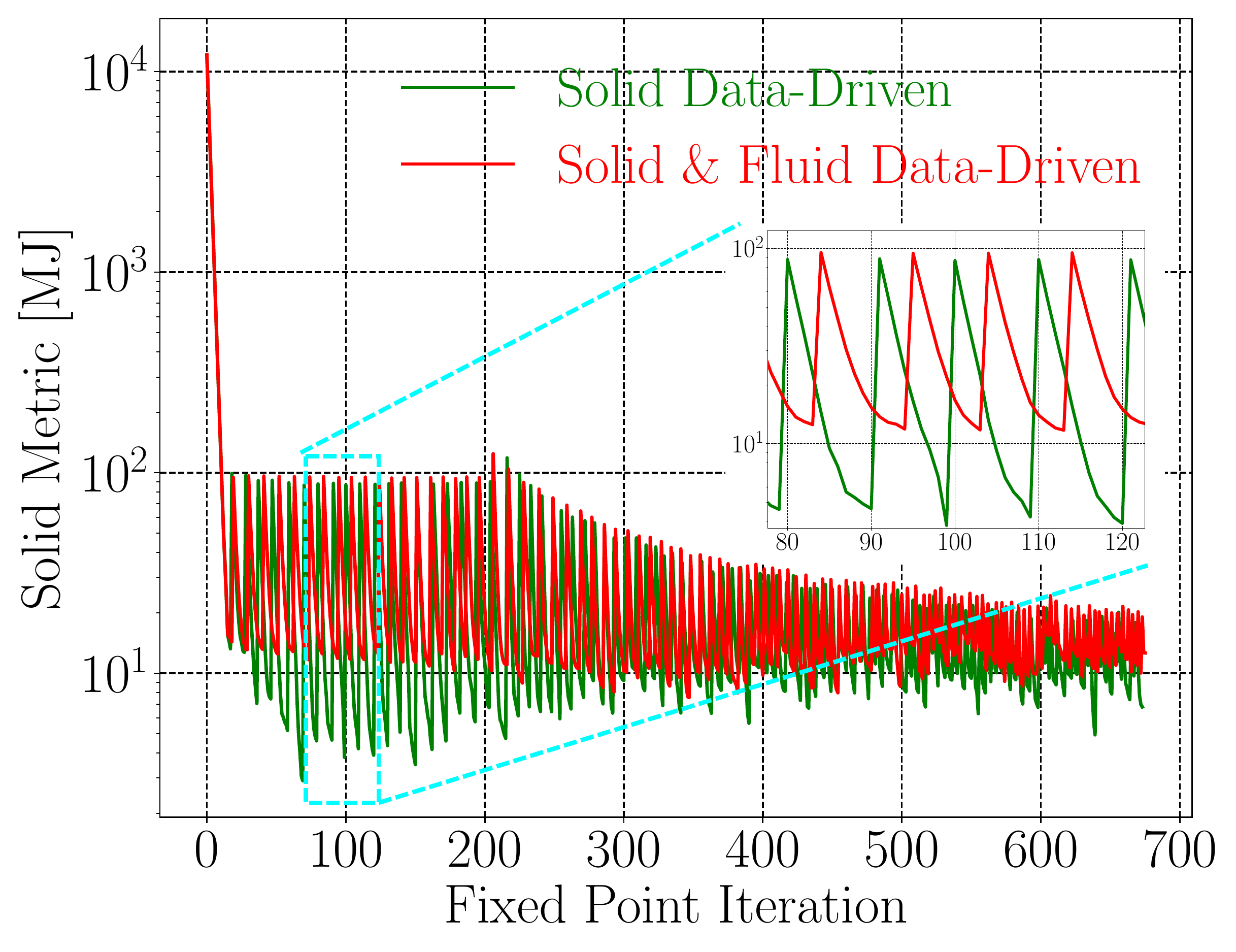}}
\hspace{0.01\textwidth}
 \subfigure[]
{\includegraphics[width=0.44\textwidth]{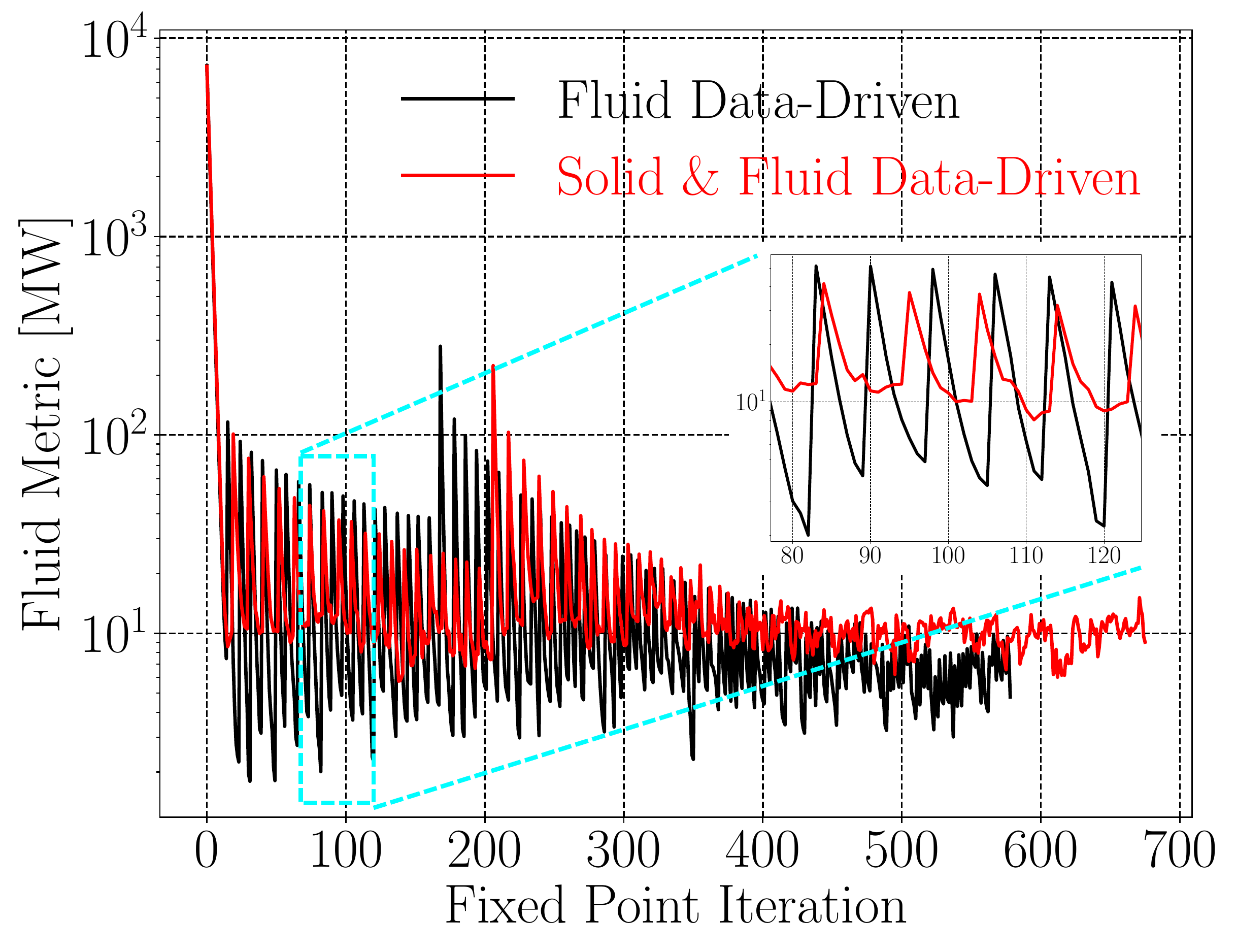}}
  \caption{Comparing the total metric at each fixed-point iteration for different schemes.
 \label{fig::strs-relax-metric-allDD}}
\end{figure}

Initial material assignments for the fixed-point iteration algorithm is random at the first time step. Within each time step, fixed-point iterations continue until the total number of material projections becomes zero. \Fig \ref{fig::strs-relax-numFixItr-allDD} depicts the number of fixed-point iterations at each time step for all data-driven schemes. This figure shows that data-driven formulations behave almost similar, in terms of the number of fixed-point iterations. Total values for metric functions (spatial integration of distance functions over the whole domain) at each fixed-point iteration for entire simulation are plotted in \fig \ref{fig::strs-relax-metric-allDD}. Each of those peaks corresponds to the first iteration of a new time step. Within each time step, we observe a decay behavior in metric function which confirms the convergence performance of fixed-point iteration, see zoom snapshots in \fig \ref{fig::strs-relax-metric-allDD}(a) and \fig \ref{fig::strs-relax-metric-allDD}(b).


\subsection{Two-dimensional Footing}
\begin{figure}[h]
 \centering
\includegraphics[width=0.6\textwidth]{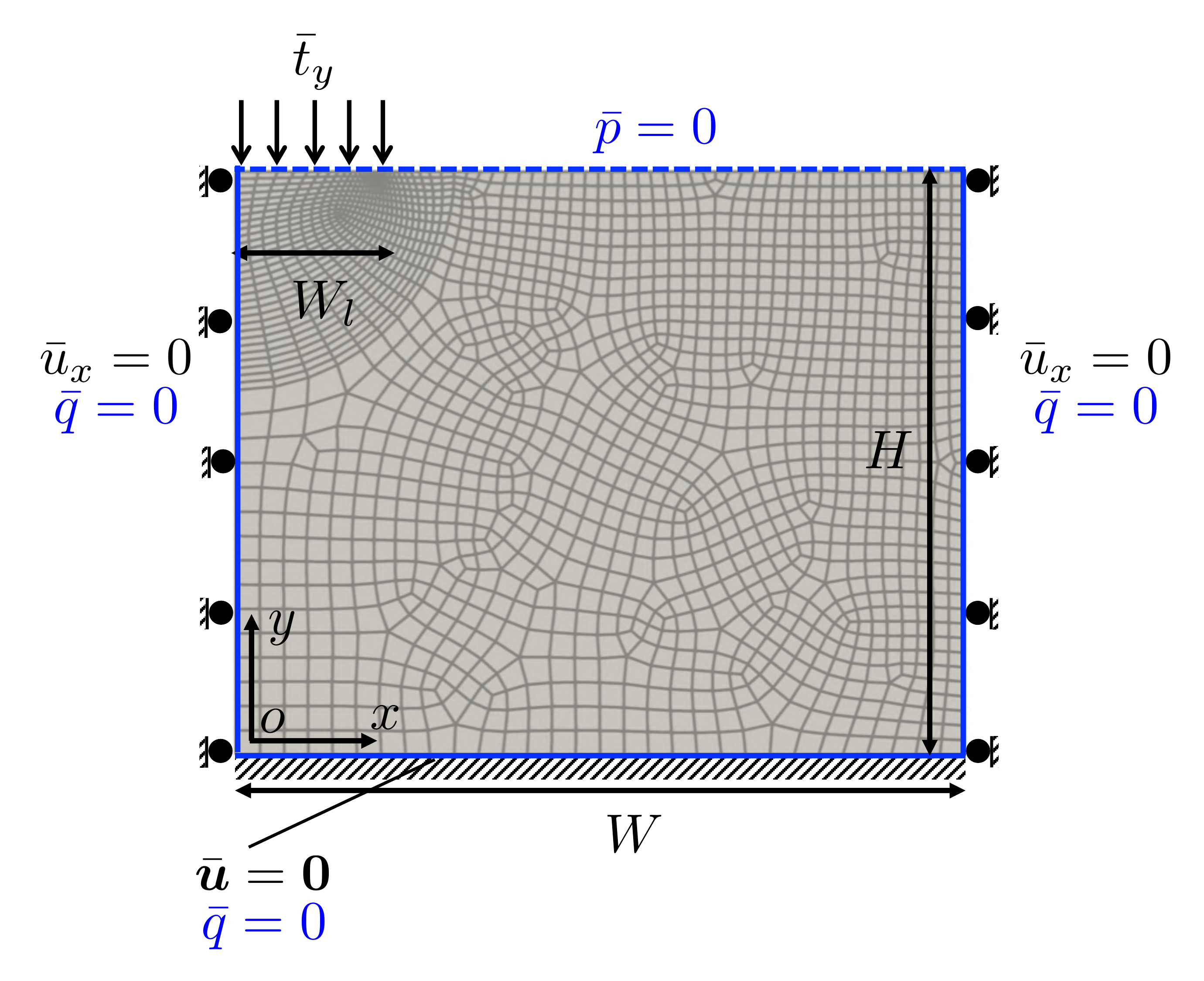}
  \caption{Geometry, mesh, and boundary conditions for the footing problem with $W=5W_l=1.25H=10\mathrm{m}$. \label{fig::footing-geom-mesh}}
\end{figure}

This problem demonstrates the capability of formulation in capturing rather a complex response field with singular characteristics. The boundary condition at the footing edge $(W_l, H)$, see \fig \ref{fig::footing-geom-mesh}, induces a singular response in stress and consequently fluid velocity fields.

\Fig \ref{fig::footing-geom-mesh} depicts boundary conditions, geometry, and mesh used herein. Normal traction over the footing area is applied with the rate $-50 \mathrm{MPa/s}$. We run the simulation for 20 time steps with time increment $\Delta t = 1\mathrm{s}$. Material properties listed in Table \ref{tab:footing-params} are utilized for a synthetic data generation or model-based simulation. We set numerical parameters of metric functions equal to $8\times 10^{-8}\tensor{I} \ \mathrm{m}^2/\mathrm{Pa.s}$ and elasticity tensor calculated for Young's modulus $24\ \mathrm{GPa}$ and Poisson's ratio $\mathrm{0.3}$. The initial data-assignment for the first fixed-point iteration is random. Bilinear Lagrangian basis functions are used for all unknown fields, and the 4-point Gaussian quadrature rule calculates numerical integration.

Fluid constitutive law follows Darcy's law, and synthetic data  is generated for $-864 \mathrm{MPa} \le \partial p / \partial x \le 984 \mathrm{MPa}$ and $-3740 \mathrm{MPa} \le \partial p / \partial y \le 288 \mathrm{MPa}$. To generate solid database, Hooke's law is assumed for the effective stress-strain relations, and solid database is sampled within strain ranges $-5.04\times 10^{-3} \le \epsilon_{xx} \le 14.3\times 10^{-3}$ , $-28.7\times 10^{-3} \le \epsilon_{yy} \le 5.04\times 10^{-3}$, and $-0.72\times 10^{-3} \le \epsilon_{xy} \le 12.1\times 10^{-3}$. In the following results, to generate data, we divide each active dimension of sample space into $N$ equidistant points. For example, database for solid part will have $N^3$ data points in total, since its sample space has only three active dimensions corresponding to symmetric strain tensor components.

\begin{table}[h]
\centering
\caption{Material parameters for Footing problem \label{tab:footing-params}}
\small%
\begin{tabular}{lll}
\hline 
Physical parameter & Unit & Value\tabularnewline
\hline 
\hline 
Young's modulus ($E$) & GPa & $30$\tabularnewline
Poisson's ratio ($\nu$) & - & $0.2$\tabularnewline
Plane condition & - & plane strain\tabularnewline
Intrinsic permeability ($k$) & $\mathrm{m}^2$ & $1\times10^{-10}$\tabularnewline
Fluid dynamics viscosity ($\mu$) & Pa.s & $0.001$\tabularnewline
Biot coefficient ($B$) & - & $1$\tabularnewline
Biot modulus  ($M$) & GPa & $\infty$\tabularnewline
\hline 
\end{tabular}
\end{table}

\begin{figure}[h]
 \centering
\includegraphics[width=1.\textwidth]{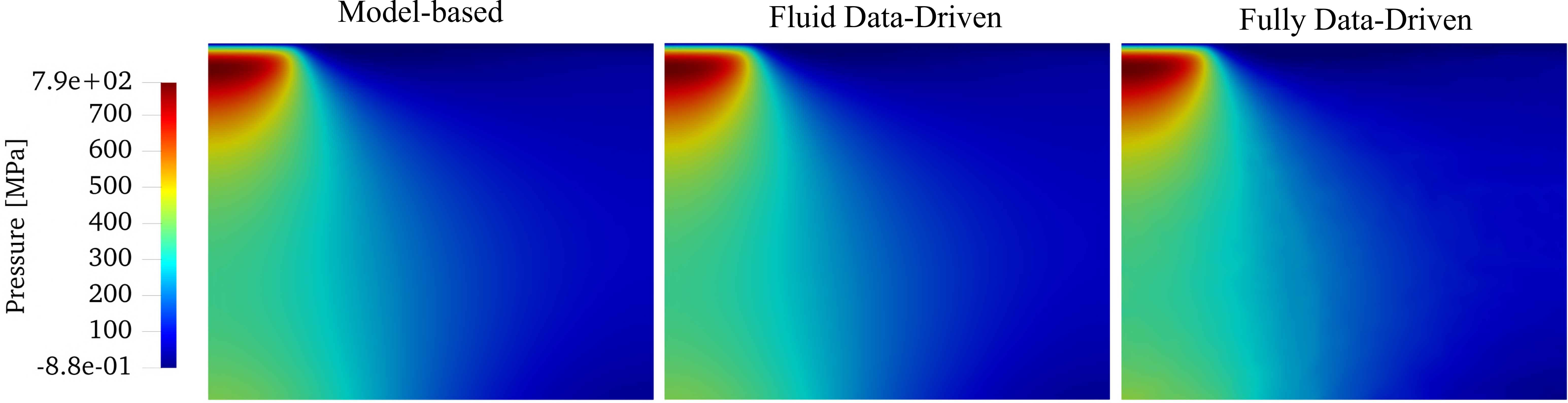}
  \caption{Fluid pressure contour. \label{fig::footing-press}}
\end{figure}

\begin{figure}[h]
 \centering
\includegraphics[width=1.\textwidth]{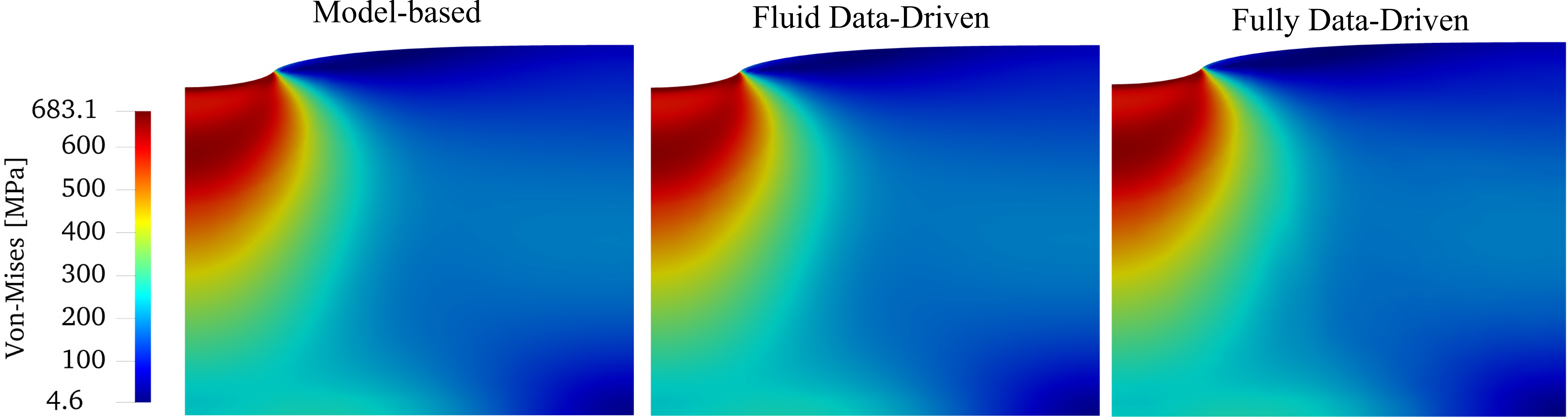}
  \caption{Von Mises stress contour on the deformed geometry. Deformation is magnified by ten times. \label{fig::footing-vonMises}}
\end{figure}

We show pressure and von-Mises total stress contours at end of simulations in \fig \ref{fig::footing-press} and \fig \ref{fig::footing-vonMises}, respectively. The database for fluid data-driven simulation has $401^2$ samples. The fully data-driven simulation has two separate data sets for fluid and solid parts with $401^2$ and $101^3$ samples, respectively. These figures confirm a good agreement between model-based and data-driven solutions.

\subsection{Plate with Hole}
In this problem, we are dealing with stress concentration and non-linearity in constitutive laws. Effective stress and strain relation follows a modified constitutive law proposed in \citep{nguyen2020variational} which is a combination of Saint-Venant and Neo-Hookean material models expressed in the limit of small deformation kinematics:
\begin{equation}
\psi(\tensor{\epsilon}) = \frac{G}{2} \left[  \tr(2\tensor{\epsilon} +\tensor{I}) - 2  -2\ln(1+\tr(\tensor{\epsilon}))  \right]
+ 
 \frac{\lambda}{2} \left[   \ln(1+\tr(\tensor{\epsilon}))    \right]^2
 +
 G \tr(\tensor{\epsilon}^2),
 \label{eq::keip-energy}
\end{equation}
where $\lambda=\kappa - 2G/3$ and $\kappa$ and $G$ are bulk and shear moduli, respectively.
\begin{figure}[h]
 \centering
\includegraphics[width=.5\textwidth]{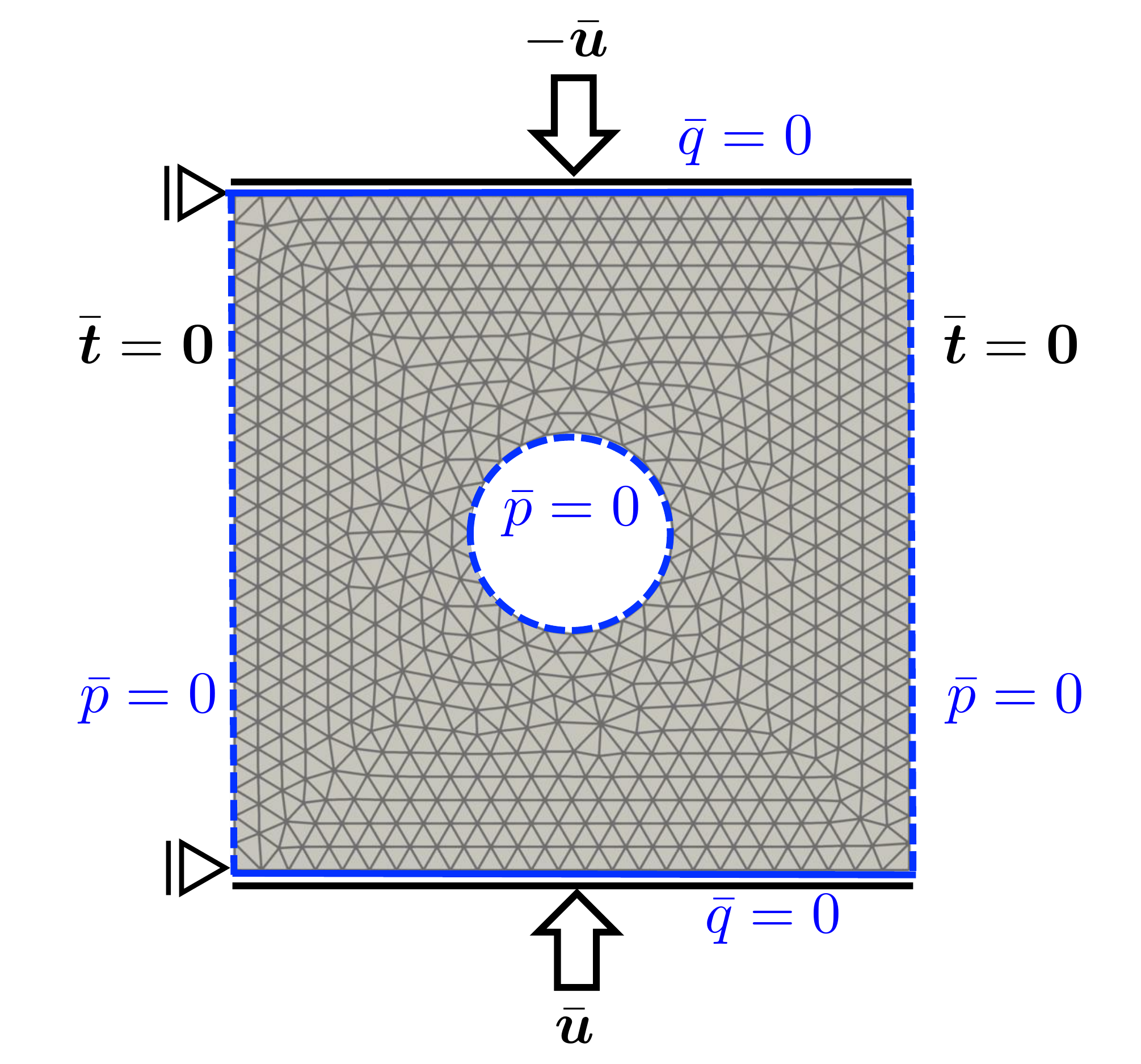}
  \caption{Geometry, mesh, and boundary conditions for the plate problem with a hole. Hole radius is $0.3\mathrm{m}$, and square has length of $2\mathrm{m}$. \label{fig::hole-geom}}
\end{figure}

As shown in \fig \ref{fig::hole-geom}, top and bottom boundaries are rigidly compressed towards each other with rate $0.4 \mathrm{m/s}$. We run the simulation for $t_{\mathrm{end}} =1.2\mathrm{s}$ with time increment $\Delta t= 0.2\mathrm{s}$. The applied deformation goes beyond the infinitesimal strain regime which might be non-physical, but this is intentional to activate severe nonlinear behavior. Material parameters are listed in Table \ref{tab:plate-hole-params}. We discretize all unknown fields with Linear Lagrangian basis functions and use one Gaussian quadrature point to spatially integrate. Data sets for fluid part are synthetically generated using Darcy's law within the ranges $ |\partial p / \partial x| \le 4.1 \mathrm{GPa}$ and $ |\partial p / \partial y| \le 3.7 \mathrm{GPa}$. Solid database is created by sampling strain and effective stress pairs for strains within ranges $0 \le \epsilon_{xx} \le 0.25$, $-0.42 \le \epsilon_{yy} \le 0$, and  $|\epsilon_{xy}| \le 0.135$. Phase spaces for each data set is built by the equidistant sampling along each data axis, same as previous examples. We set Numerical parameters for fluid metric equal to $8\times 10^{-7}\tensor{I} \ \mathrm{m}^2/\mathrm{Pa.s}$ and for solid metric equal to the Hessian tensor of functional Eq. \eqref{eq::keip-energy} at $\epsilon_{xx} = 0.1$, $\epsilon_{yy} = -0.2$, and $\epsilon_{xy} = 0.05$ with the same material properties used for the data generation 

\begin{figure}[h]
 \centering
\includegraphics[width=.9\textwidth]{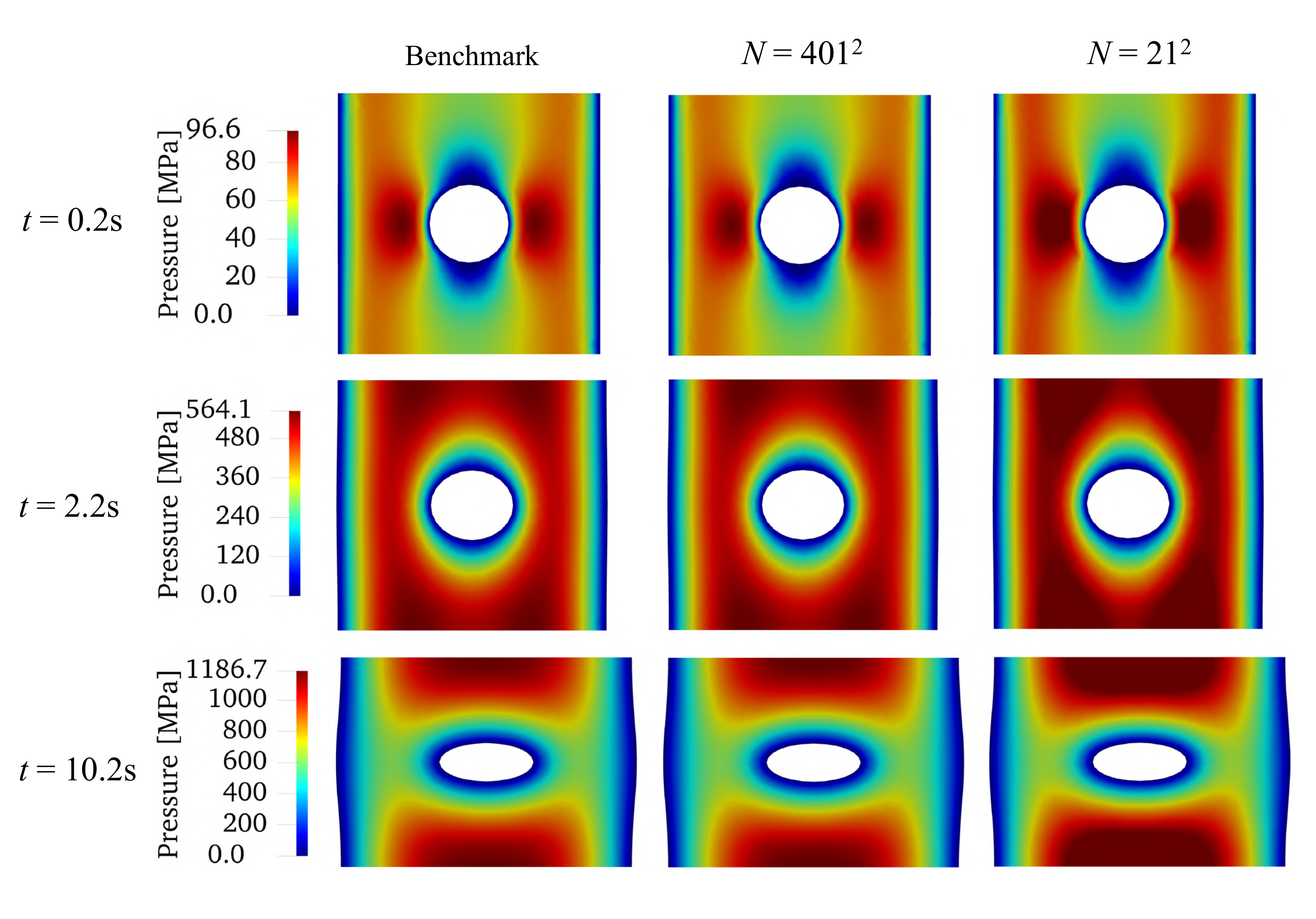}
  \caption{Pressure contour evolution in time for different database sizes used in the fluid data-driven scheme. \label{fig::hole-press-fluidDD}}
\end{figure}
\begin{figure}[h]
 \centering
\includegraphics[width=.9\textwidth]{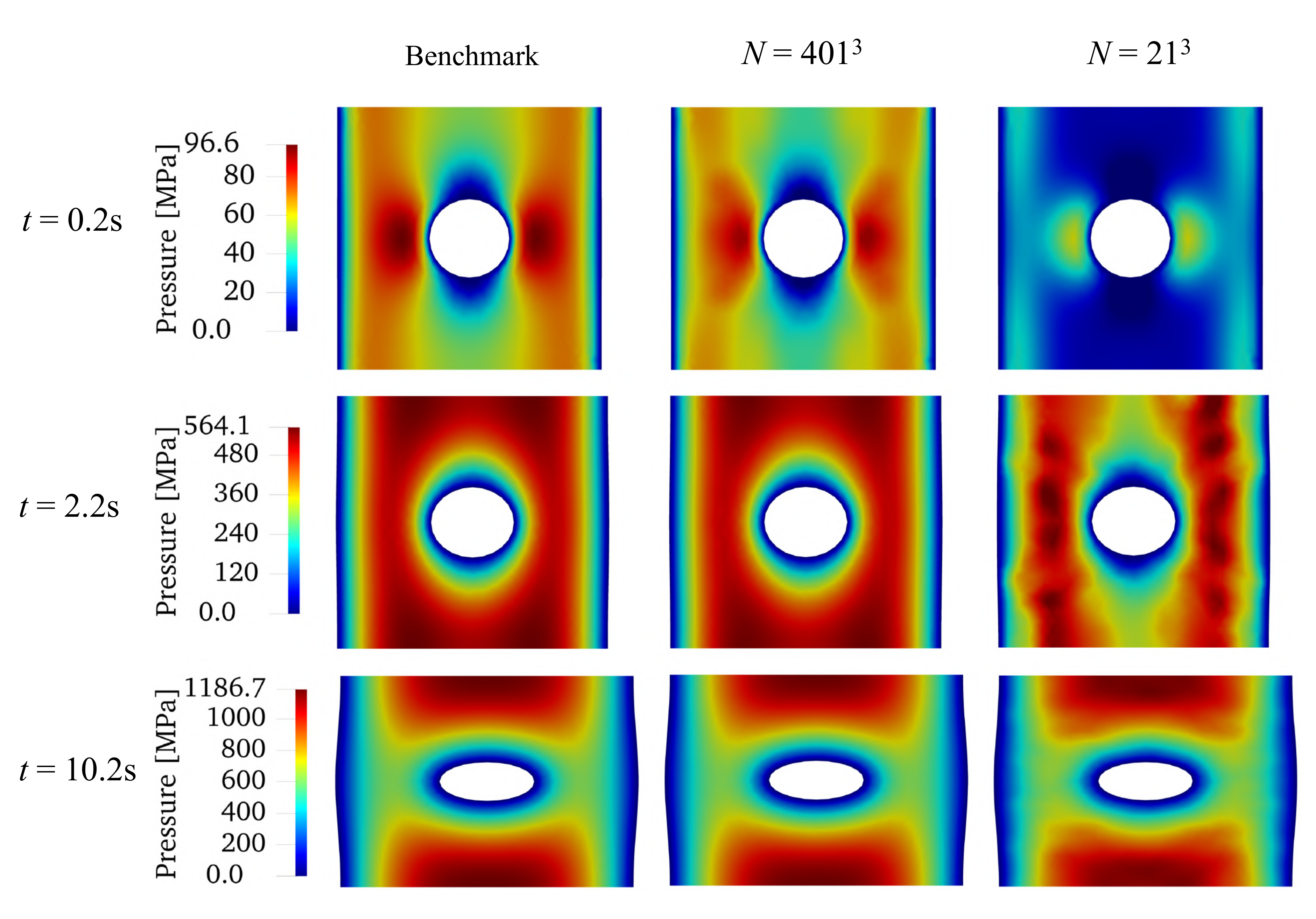}
  \caption{Pressure contour evolution in time for different database sizes used in the solid data-driven scheme. \label{fig::hole-press-solidDD}}
\end{figure}

In \fig \ref{fig::hole-press-fluidDD}, there is a satisfactory agreement between benchmark solution, conventional model-based FEM solution, and results obtained by the fluid data-driven formulation with $N=401^2$ data points for the pressure field. The low-intensity data $N=21^2$ is still able to track the benchmark solution qualitatively. \Fig \ref{fig::hole-press-solidDD} presents the same results for solid data-driven formulation. In this figure, pressure fields at time $t=0.2\mathrm{s}$ are not as good as their counterparts in case of fluid data-driven. First and foremost, comparing results between $N=21^3$ and $N=401^3$ at time $t=0.2\mathrm{s}$ in \fig \ref{fig::hole-press-solidDD} confirms that increasing amount of data improves the accuracy. In other words, case $N=401^3$ still does not have sufficient data coverage to hit the benchmark's response surface. Second, the numerical parameter for solid metric function might not be appropriately chosen which has a considerable effect on the data-driven performance \citep{leygue2018data}. Fine-tuning process for numerical parameters in metric functions is out of the scope of this manuscript. Note that solid behavior in this problem is highly nonlinear, while the fluid constitutive behavior is linear. Also, solid phase-space belongs to a higher dimensional space than fluid phase-space. Therefore, in this particular problem, it is not unexpected to observe that fluid data-driven formulation, which benefits from an explicit model for solid constitutive law, outperforms solid data-driven formulation. In the other side, pressure responses of solid data-driven formulation at times $t=2.2\mathrm{s}, 10.2\mathrm{s}$ for data intensity $N=401^3$ in \fig \ref{fig::hole-press-solidDD} are in good agreement with benchmark.


\begin{table}[h]
\centering
\caption{Material parameters for plate with hole \label{tab:plate-hole-params}}
\small%
\begin{tabular}{lll}
\hline 
Physical parameter & Unit & Value\tabularnewline
\hline 
\hline 
Young's modulus ($E$) & GPa & $30$\tabularnewline
Poisson's ratio ($\nu$) & - & $0.35$\tabularnewline
Plane condition & - & plane strain\tabularnewline
Intrinsic permeability ($k$) & $\mathrm{m}^2$ & $3.0612 \times 10^{-9}$\tabularnewline
Fluid dynamics viscosity ($\mu$) & Pa.s & $0.001$\tabularnewline
Biot coefficient ($B$) & - & $1$\tabularnewline
Biot modulus  ($M$) & GPa & $600$\tabularnewline
\hline 
\end{tabular}
\end{table}

\begin{figure}[h]
 \centering
\includegraphics[width=.4\textwidth]{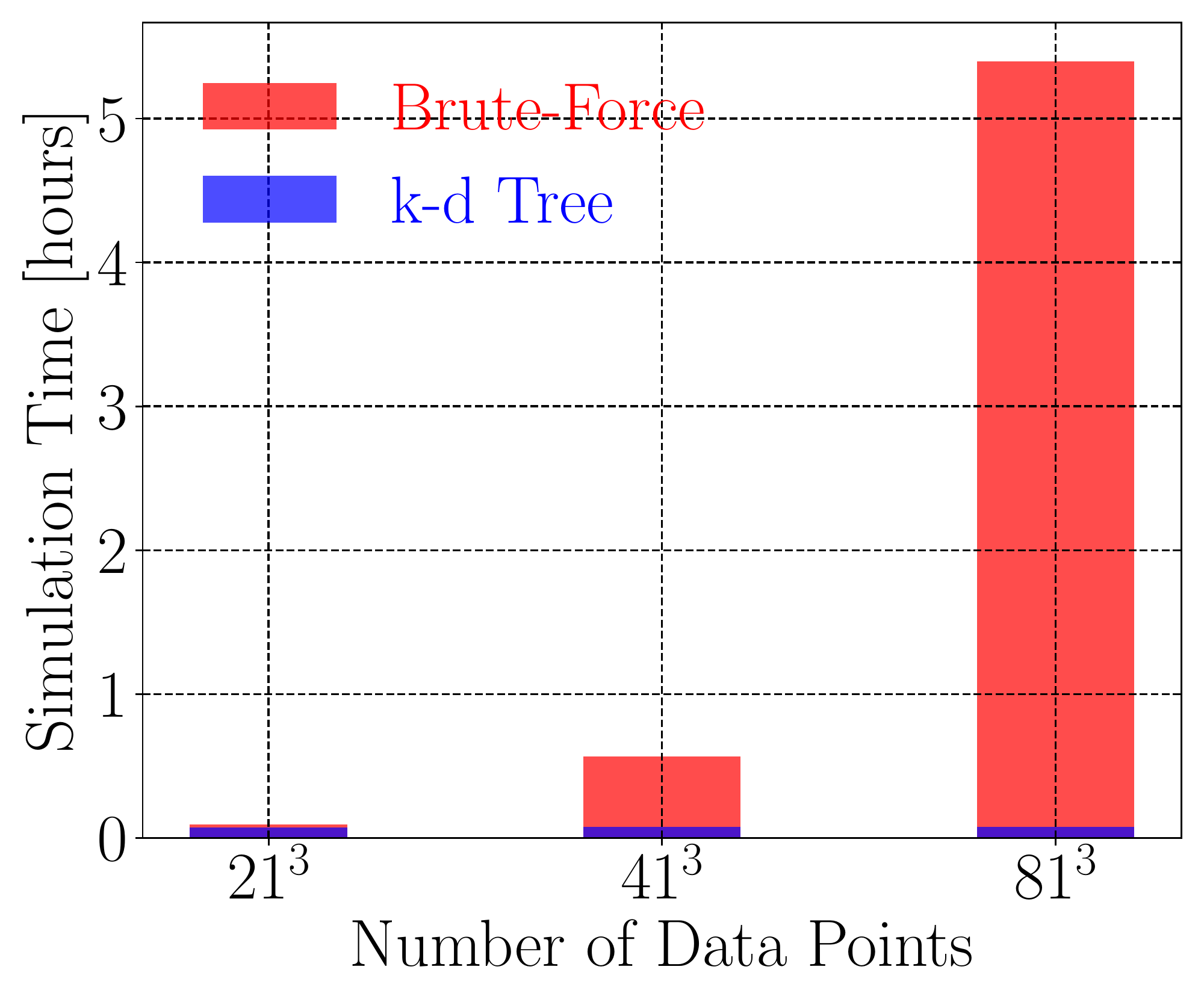}
  \caption{Total simulation time for k-d tree and brute-force approaches. \label{fig::sim-time-hole-plate}}
\end{figure}

We compare the performance of the kd-tree nearest neighbor search with the brute-force approach in \fig \ref{fig::sim-time-hole-plate}. Not surprisingly, due to the logarithmic time complexity of the kd-tree search, the kd-tree outperforms the naive approach, especially for a large database (e.g., in the case $N=81^3$, it reduces a five hours simulation to less than 10 minutes). Note that Newton-Raphson iterations increase the number of NNS requests in nonlinear problems. Consequently, a fast NNS approach may outperform the brute-force one, even for a small database. Recall the conclusion drawn in Appendix \ref{kd-tree-study} where the brute-force scheme may have a slightly better performance for small data sets in dealing with a linear problem.

\subsection{Multifidelity Simulations of Berea Sandstone data}
This problem is designed to demonstrate how the data-driven formulation can be used to handle 
the different fidelities of the solid skeleton elasticity and the hydraulic responses of a porous medium. The difference in fidelity is a well-known issue in poromechanics problems where 
the predictions of the effective permeability are often much less accurate \textbf{and} precise 
than the solid constitutive responses \citep{paterson2005experimental, lock2002predicting, andra2013digital, sun2018prediction}. For instance, a porosity-permeability model capable of 
predicting effective permeability with less than an order of the discrepancy from the
experiment measurement is often considered accurate \citep{bernabe2003permeability, costa2006permeability}. Meanwhile, experimental tests (e.g., uniaxial test and triaxial test) may routinely calibrate
an isotropic linear elasticity model delivering predictions with a few 
percents of errors in the path-independent regime 
\citep{dvorkin1999elasticity, zohdi2001computational, paterson2005experimental, pimienta2015bulk}.

In these highly plausible scenarios, the benefit of replacing the solid constitutive law with the data-driven approach might not outweigh the efficiency of the model-based approach, whereas 
the data-driven approach may still be favorable due to the inevitable high variance of the predictions made by the deterministic porosity-permeability models. 

In this example, we create the fluid database (see \fig \ref{fig::fft_data}(a)) by running 84 FFT simulations on 
microCT images of Berea sandstone (provided by collaborator Prof. Teng-fong Wong) and use the data-driven approach in Sec. \ref{sec:hybrid1} to run simulations without using hydraulic constitutive law. Meanwhile, the solid constitutive response is obtained from the literature  \citet{andra2013digital}.

\begin{figure}[h]
 \centering
{\includegraphics[width=0.7\textwidth]{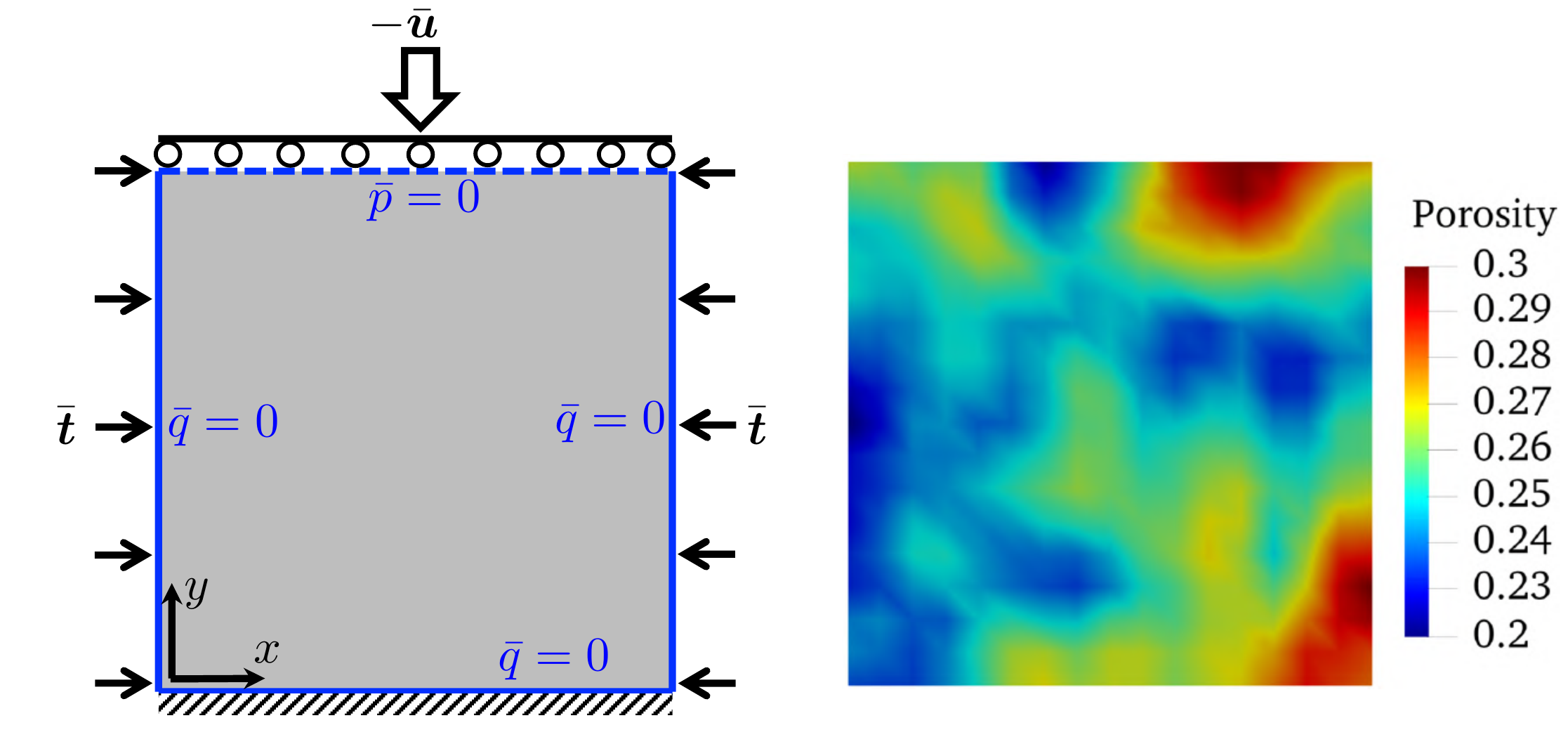}}
  \caption{(a) Geometry and boundary conditions for Berea sandstone and (b) heterogeneous distribution of initial porosity with average 0.25 and standard deviation 0.02.
   \label{fig::berea-prob}}
\end{figure}

The boundary value problem is two-dimensional, and the plane strain condition is assumed, as shown in \fig \ref{fig::berea-prob}(a). Lateral confining pressure of $6416.6\ \mathrm{[MPa]}$ is applied at the beginning of the simulation and kept constant throughout the simulations. A downward displacement is prescribed at the top boundary with the rate $1.2  \ \mathrm{mm/day}$. Time increment is set at $\Delta t = 12\  \mathrm{[hours]}$. The spatial domain is discretized by $15\times 15$ regular quadrilateral elements.

To test the versatility of the data-driven approach in handling boundary value problems with a spatially heterogeneous domain, we employ an open-source code developed by \citep{Baker2011characterization} to generate a realization of initial porosity distribution. 
This random porosity field is element-wise and is generated based on the exponential model with a mean value $0.25$ and a standard deviation $0.02$, shown in \fig \ref{fig::berea-prob}(b), which is in the range of values reported in \citep{andra2013digital}. 
Solid constitutive law is based on the hyperelastic energy functional introduced in chapter 6.4 of \citep{Borja2013} as follows: 
\begin{equation}
\psi(\epsilon_v, \epsilon_s) = -p_0c_r \exp \left( \frac{\epsilon_{v0} - \epsilon_v}{c_r}  \right) - \frac{3}{2} c_{\mu} p_0 \exp \left( \frac{\epsilon_{v0} - \epsilon_v}{c_r} \right) {\epsilon_s}^2 + \psi_0,
\end{equation}
where $p_0$, $c_r$, $c_{mu}$, and $\epsilon_{v0}$ are model parameters, and $\epsilon_v = \tr(\tensor{\epsilon})$ and $\epsilon_s = \sqrt{2/3} \norm{\tensor{\epsilon} - 1/3 \epsilon_v \tensor{I}}$ are volumetric and shear strains, respectively. This model introduces pressure-dependent bulk and shear moduli which is considered necessary for modeling soils and sands \citep{borja1997coupling, andrade2006capturing}. In this model, bulk and shear moduli are related to diagonal terms of Hessian matrix, \nth{2} derivative of energy functional with respect to volumetric and shear strains, with the following relations: 
\begin{align}
&
\kappa(\epsilon_v, \epsilon_s) = -\frac{p_0}{c_r} \left(  1+ \frac{3c_{\mu}}{2c_r} {\epsilon_s}^2  \right)  \exp \left(  \frac{\epsilon_{v0} -\epsilon_v}{c_r}  \right),
\label{eq::bulk_eq}
\\&
\mu(\epsilon_v, \epsilon_s) = -c_{\mu} p_0 \exp \left(  \frac{\epsilon_{v0} -\epsilon_v}{c_r}  \right).
\label{eq::shear_eq}
\end{align}

\begin{figure}[h]
 \centering
 \subfigure[]
{\includegraphics[width=0.33\textwidth]{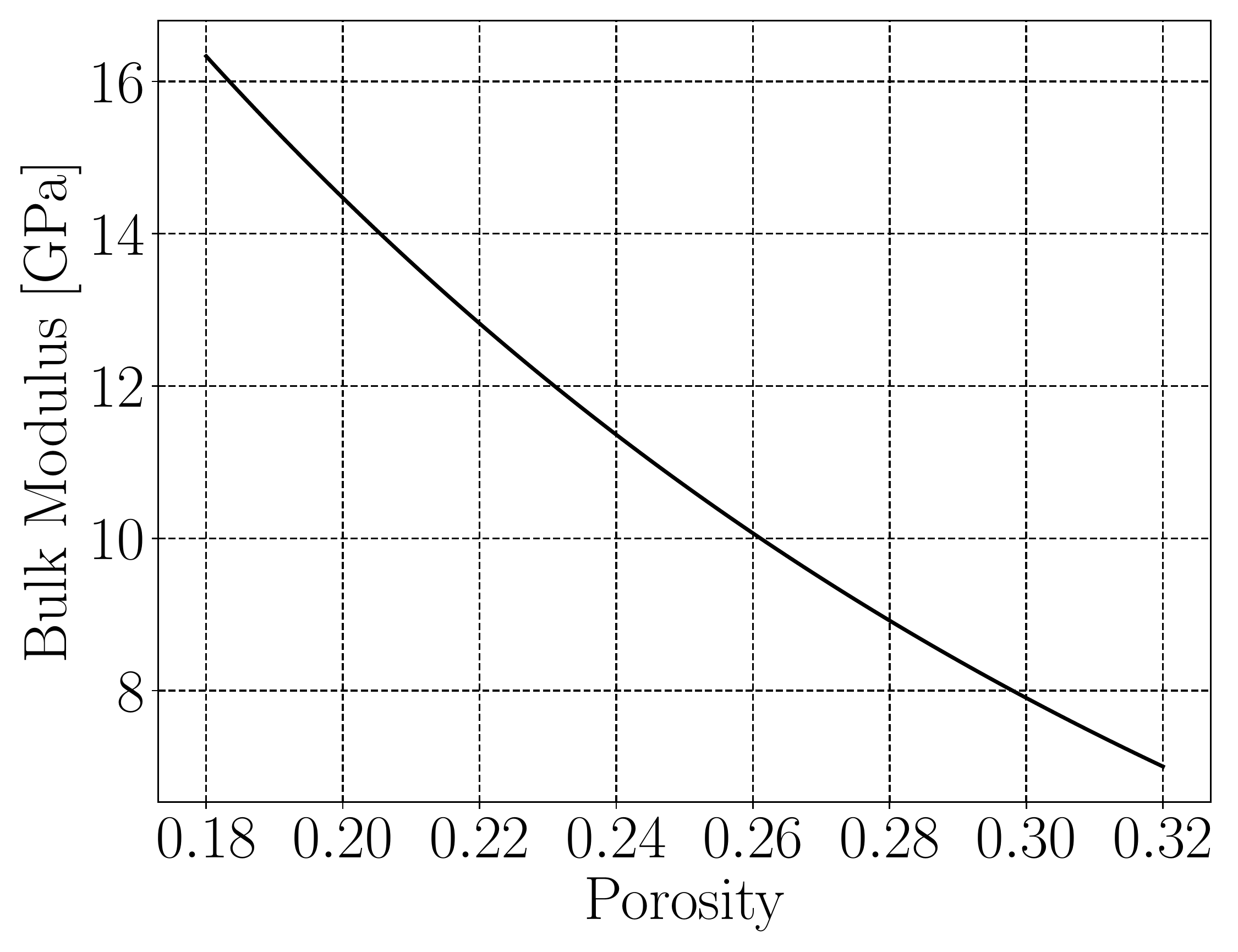}}
\hspace{0.01\textwidth}
 \subfigure[]
{\includegraphics[width=0.33\textwidth]{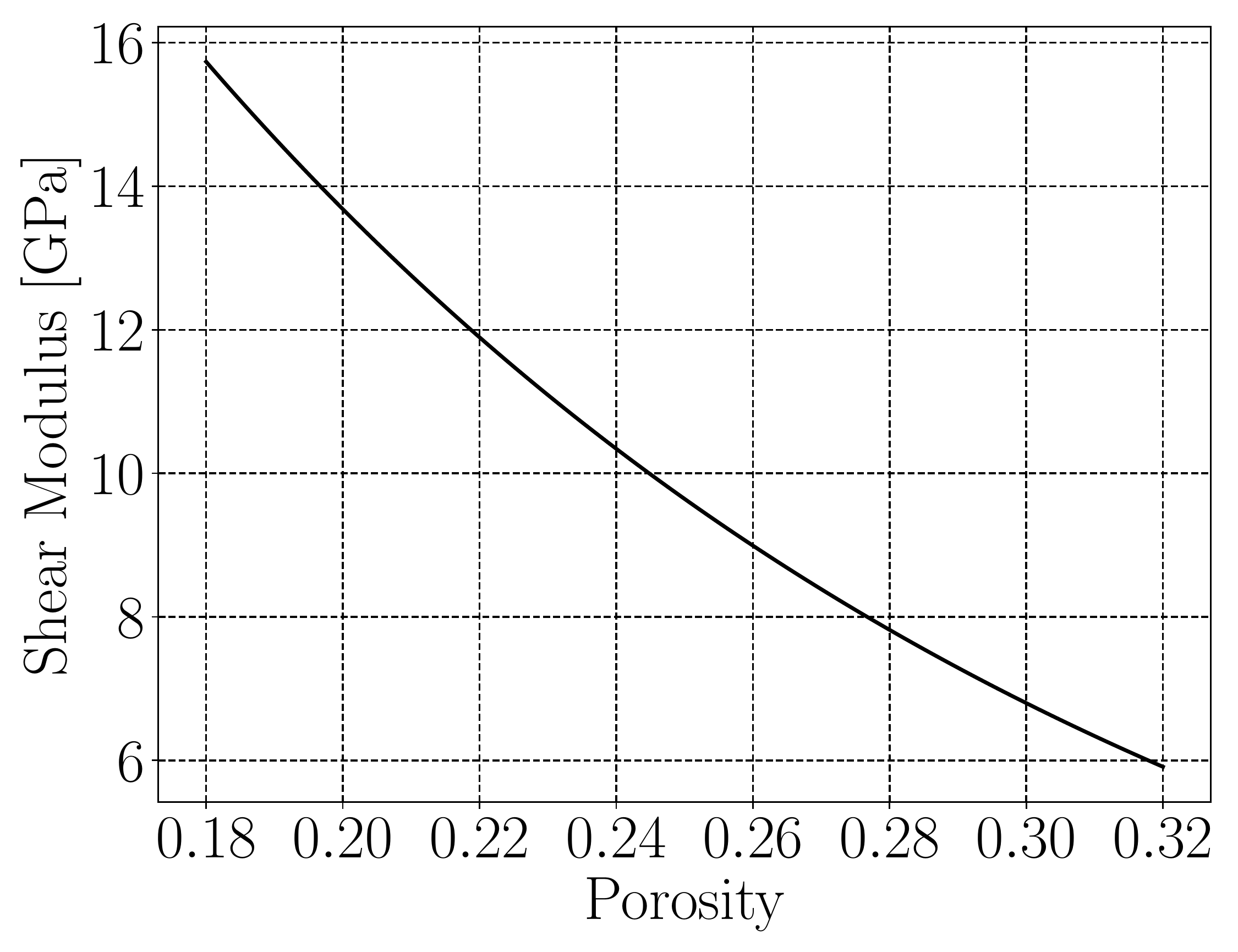}}
\hspace{0.01\textwidth}
  \caption{Bulk and shear moduli for Berea sandstone extracted from \citep{andra2013digital}.
 \label{fig::bulk_shear_base}}
\end{figure}

\begin{figure}[h]
 \centering
{\includegraphics[width=0.9\textwidth]{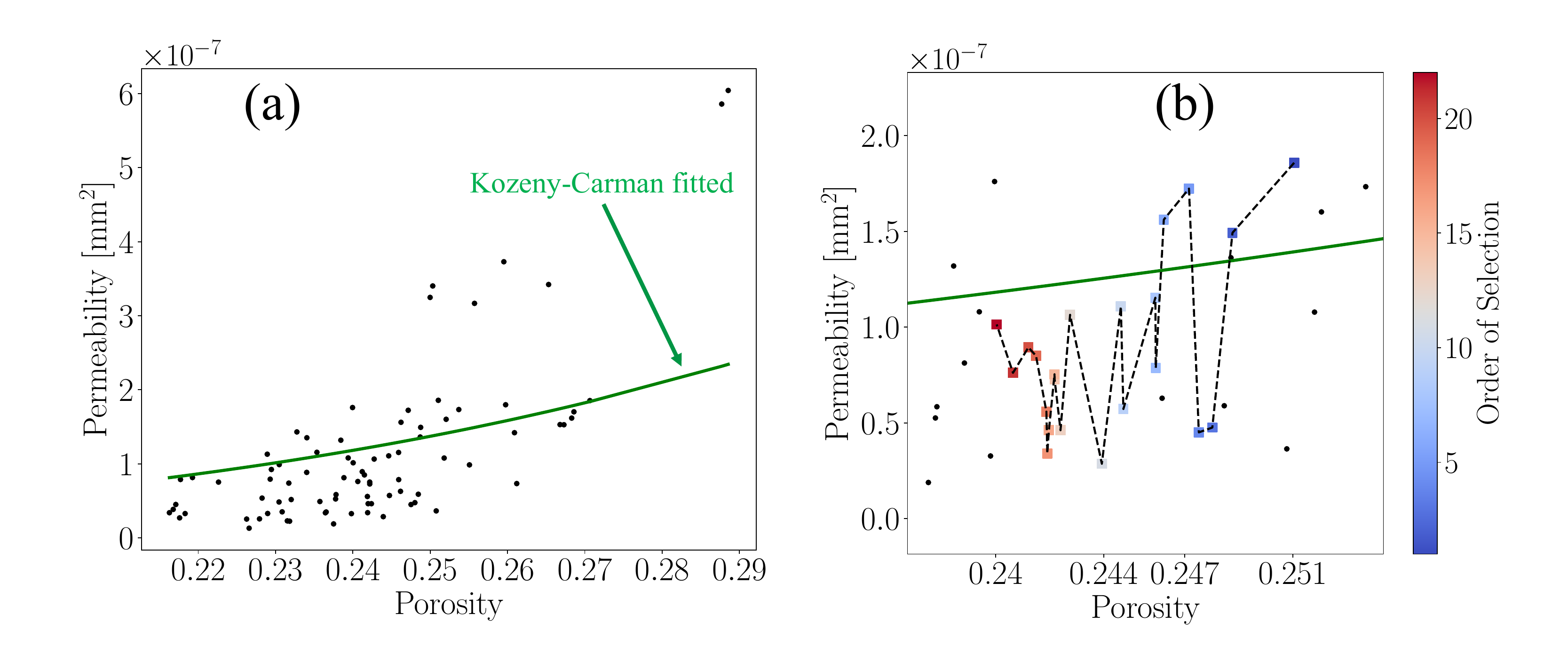}}
  \caption{(a) Fine-scale homogenized porosity-permeability data; (b) trajectory of data assignments for a quadrature point during the macro-scale simulation. The solid green line shows the fitted Kozeny-Carman curve to the points in figure (a). We observe a considerable discrepancy between the Kozeny-Carman model and ground truth points.
   \label{fig::fft_data}}
\end{figure}

Figure \ref{fig::bulk_shear_base} plots bulk and shear moduli as exponential functions fitted to curves provided in \citep{andra2013digital}, see \fig 2(a) and \fig 4(a) in the reference paper. These porosity dependent modului can be related to volumetric strain by $\phi = (1+\epsilon_v) \phi_0$ for the small deformation analysis \citep{wang2017unified}. Therefore, for each computational element with a distinct initial porosity $\phi_0$ we are able to calibrate model parameters  $p_0$, $c_r$, $c_{mu}$, and $\epsilon_{v0}$ to match bulk and shear moduli Eqs. \eqref{eq::bulk_eq} and \eqref{eq::shear_eq} with those so-called experimental curves shown in \fig \ref{fig::bulk_shear_base}. Since we do not have information about shear loading conditions in \citep{andra2013digital} we set $\epsilon_s=0$ for just parameter calibration process.

A collection of upscaled effective permeability for 84 Representative Volume Elements (RVEs) corresponding to 3D images of Berea sandstone is obtained by solving Stokes equations at microscale using an in-house developed Fast Fourier Transform (FFT) based solver \citep{ma2020fft, ma2020computational, maphase}. Figure \ref{fig::fft_data}(a) shows these 84 porosity-permeability pairs. The highly non-smooth behavior in \ref{fig::fft_data}(a) might be non-physical, but we assume there is such a case, and so we intend to challenge the data-driven solver. Isotropic Darcian flow is assumed to generate semi-synthetic data sets based on these fine-scale simulations, similar to previous problems. Since our goal in this paper is limited to show the robustness of formulation rather than application we stick to such a semi-synthetic data generation. Otherwise, someone needs to simulate numerous micro-scale samples to gather sufficient data. For each point in \fig \ref{fig::fft_data}(a), a data set of $801^2$ pressure gradient and fluid velocity pairs is created and labeled for a specific porosity. All generated data sets cover same range of pressure gradients as $-902 \mathrm{[MPa/m]} \le  \partial p/ \partial x  \le 1540\mathrm{[MPa/m]}$ and $-3190 \mathrm{[MPa/m]} \le  \partial p/ \partial x  \le 1650\mathrm{[MPa/m]}$. After many trials and errors, these ranges are found by changing the bounds to make sure final numerical pressure gradients belong to them. Numerical parameter for fluid metric function is set to $10^{-11}\tensor{I}$.

Since data sets have a dependence on porosity we propose a modification to the algorithm for material projection which consists of two steps. First, based on the current estimation of porosity at a quadrature point, the closest data set among 84 options will be selected. Then, inside the founded data set, the closest pressure gradient and fluid velocity pair will be reported similar to earlier problems.

\Fig \ref{fig::fix-pnt-itr-berea} depicts the performance of solver at each fixed-point iteration. Each of those jumps (peaks) corresponds to an initial iteration at a new time step. The initial guess at a new time step is expected to be off due to the new loading condition, and so we should expect those jumps. We expect to converge to a constant metric value within each time step along with fixed-point iterations without any further material projection. Generally, these figures confirm the same behavior, but some time steps do not converge exactly to a constant metric value. Metric values at those time steps slightly oscillate, see behavior around iteration 300 in the zoomed snapshot of \fig \ref{fig::fix-pnt-itr-berea}(a). However, the total numbers of material projections are close to zero, see corresponding iterations in the zoomed snapshot of \fig \ref{fig::fix-pnt-itr-berea}(b), and not more than 5 in the worst case happened in the simulation. Therefore, we can say the solver's overall performance is satisfactory because the exact minima for just a few material points (lest than $3\%$) are lost. This oscillatory behavior is not something unexpected since at some regions in \fig \ref{fig::fft_data}(a) porosity values are very close to each other while permeability values differ considerably and so, after some fixed-point iterations, there is a possibility for the solver to be trapped between several choices each selected from a different data set. It is well-known that heuristic approaches, e.g., fixed-point iteration used in our research, may not find the global optima for combinatorial optimization problems \citep{kanno2019mixed, guillermo2020framework}, i.e., trapped between some local minima. As suggested in \citep{kanno2019mixed} an alternative optimization problem can be formulated by a relaxation method which makes the optimization problem convex, and consequently, a global optimum is guaranteed. We have not studied such a formulation in the current research.
{
\begin{figure}[h]
 \centering
{\includegraphics[width=0.99\textwidth]{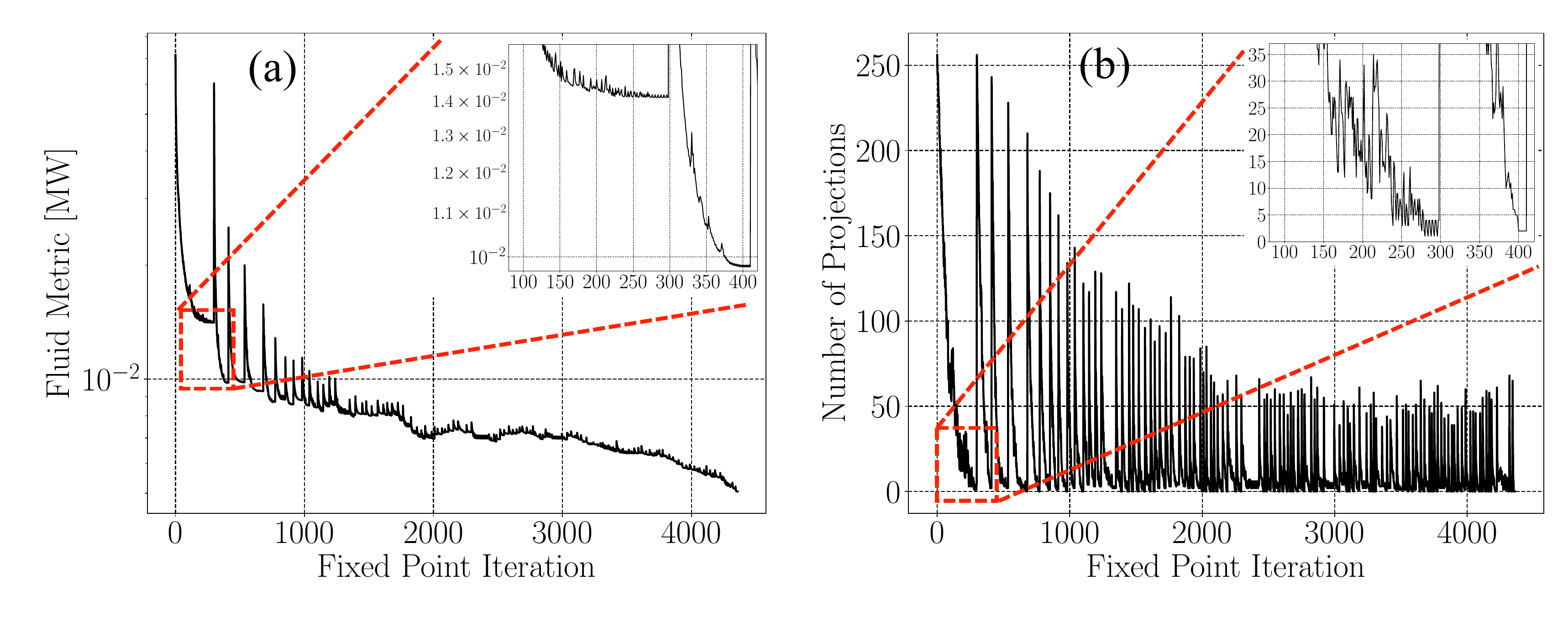}}
  \caption{Total fluid metric value (a) and the number of projected material points (b) at the end of each fixed-point iteration for 100 time steps.
   \label{fig::fix-pnt-itr-berea}}
\end{figure}

Porosity and pressure contours of selected data sets at the end of the material projection process are shown at different time stages in \fig \ref{fig::poro-contour-berea} and \fig \ref{fig::press-contour-berea}, respectively. As expected, for most of the quadrature points, the data points at lower porosity value are selected as time elapses. \Fig \ref{fig::fft_data}(b) showcases history of selected data sets in 100 time steps of the simulation for one of the quadrature points located at position $(0.3, 3.7)$.  Initially, a data set with porosity 0.251 is assigned to this point, and a data set with porosity 0.24 is selected at the end of simulation. This material point experiences almost 20 different permeability values with a considerable variance and no consensus trend, even for such a small variation of porosity values. If, instead, we used Kozeny–Carman relation to model porosity-dependent permeability we observed a smooth decay in permeability value by decreasing porosity.

Due to the high variance of the permeability dataset, selecting and calibrating a porosity-permeability model is by no mean a trivial ask. While a complicated model fluctuated with porosity is likely to overfit the data, a simple model such as the Kozeny-Carmen model may underfit the data and hence lead to significant errors (see solid green line in \fig \ref{fig::fft_data}(a)). While extending the dimensions of the parametric space (e.g., adding features \citep{sudakov2019driving} or geometric measures such as fabric tensors  \citep{sun2013multiscale} of the hydraulic model may improve the bias-variance trade-off, the increased dimensionality may also increase the difficulty of deriving calibrating the constitutive laws or the training of neural network models. 

Note that the data-driven approach does not impose or require any smoothness assumption among the data points in the phase space. 
While the assumptions of smoothness, continuity or convexity are not 
always physically justified, the lack thereof of the data-driven approach does 
impose a high demand on the data. To make a prediction, there must be a sufficient population or density of data locally distributed 
in the parametric space where the data point becomes admissible to the physics constraints. Recent work such as \citet{eggersmann2020model} introduces an approach based on tensor voting to make data-driven methods working with less intensive data, but the demand for data is still noticeably higher than the conventional approach. A model-free method that can operate under less data by incorporating physics constraints and statistics will be considered in the future but is out of the scope of this study.

\begin{figure}[!h]
\newcommand\imsz{6cm}
\newcommand\labeloffset{2cm}
\centering
\begin{tabular}{ m{\imsz} m{\imsz} }
 {\includegraphics[width=\imsz]{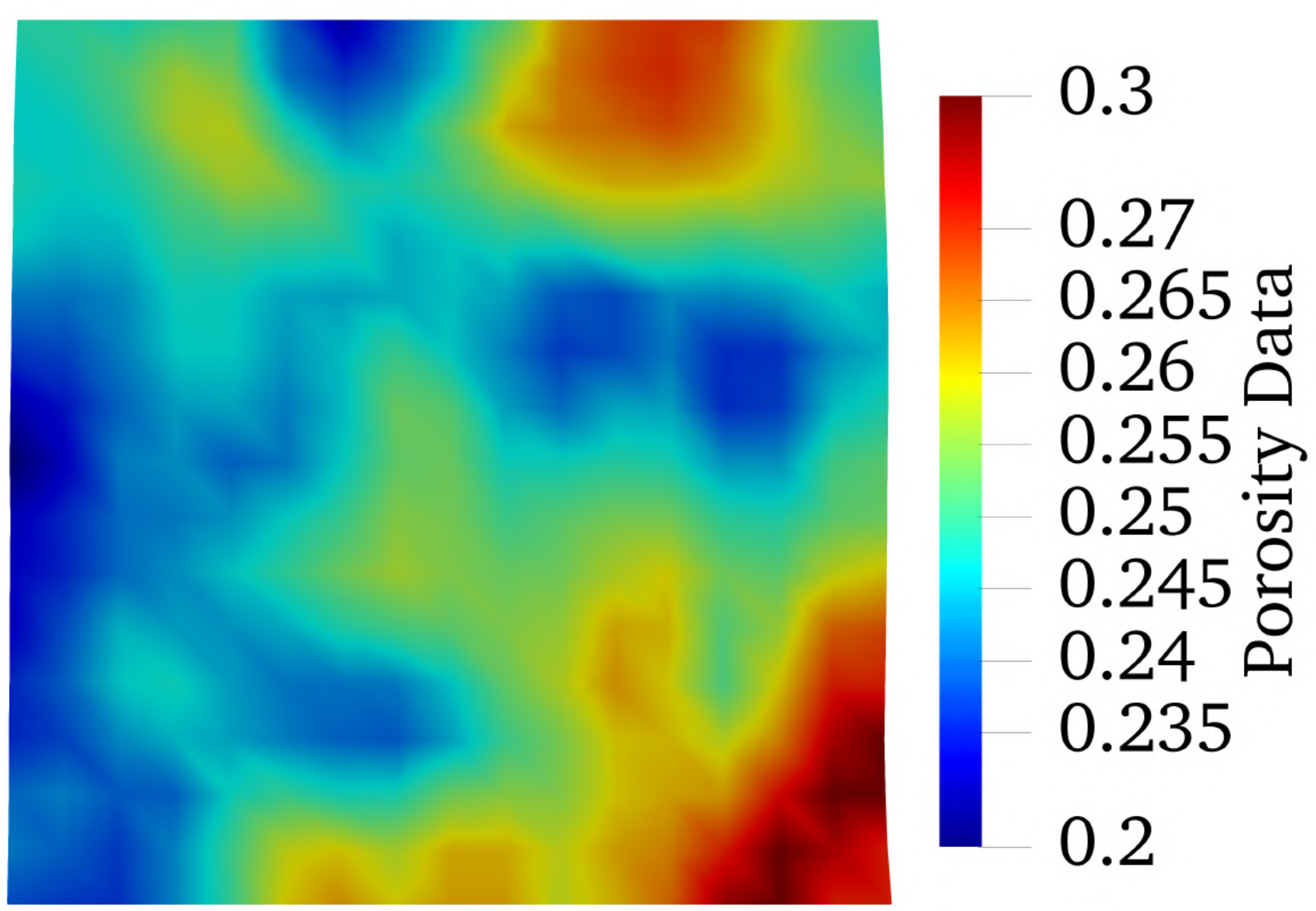}} & 
{\includegraphics[width=\imsz]{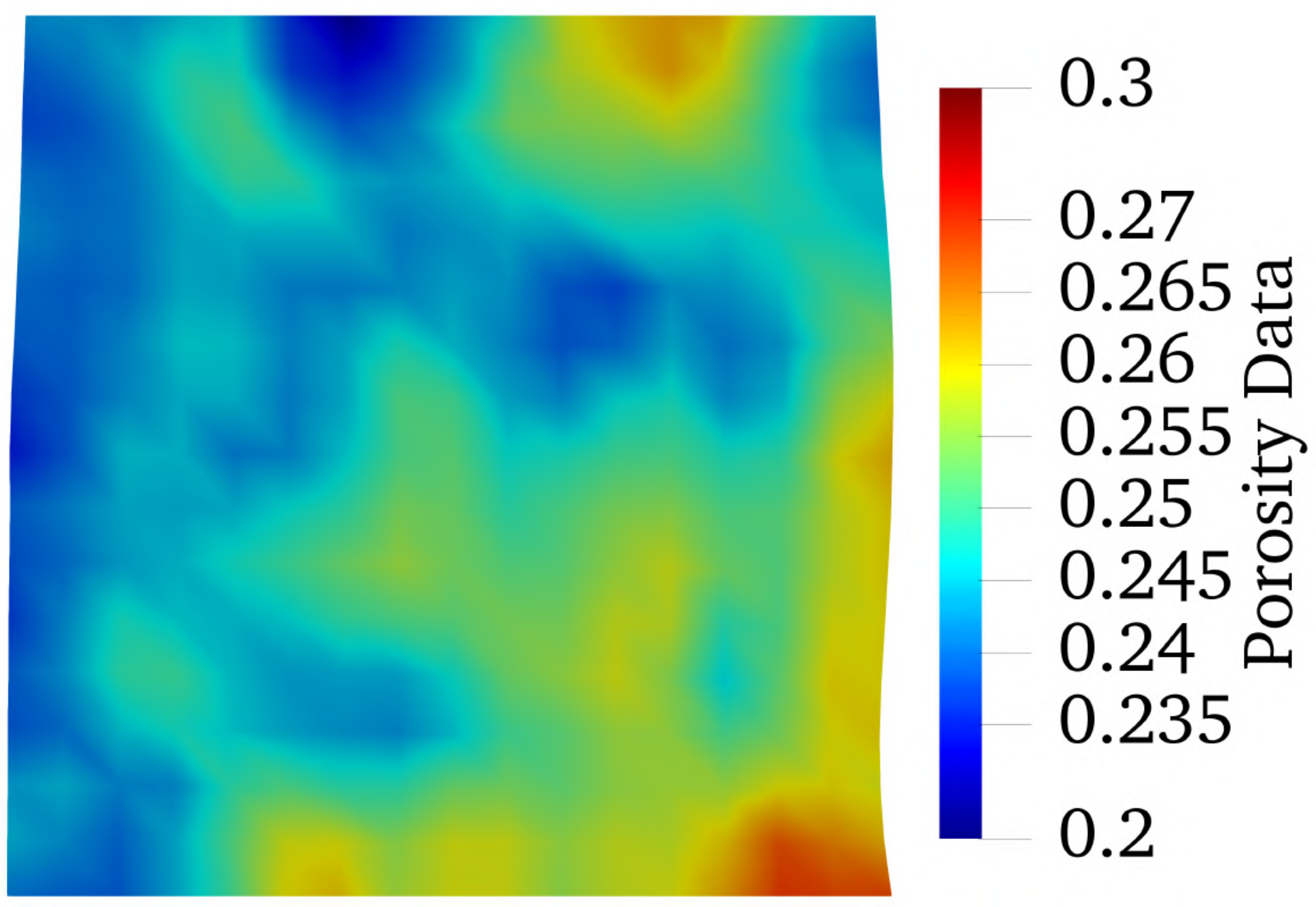}} \\ 
\hspace{1.1cm}$t=12 \ [\mathrm{hours}]$ & 
\hspace{0.9cm}$t=15.5 \ [\mathrm{days}]$\\
{\includegraphics[width=\imsz]{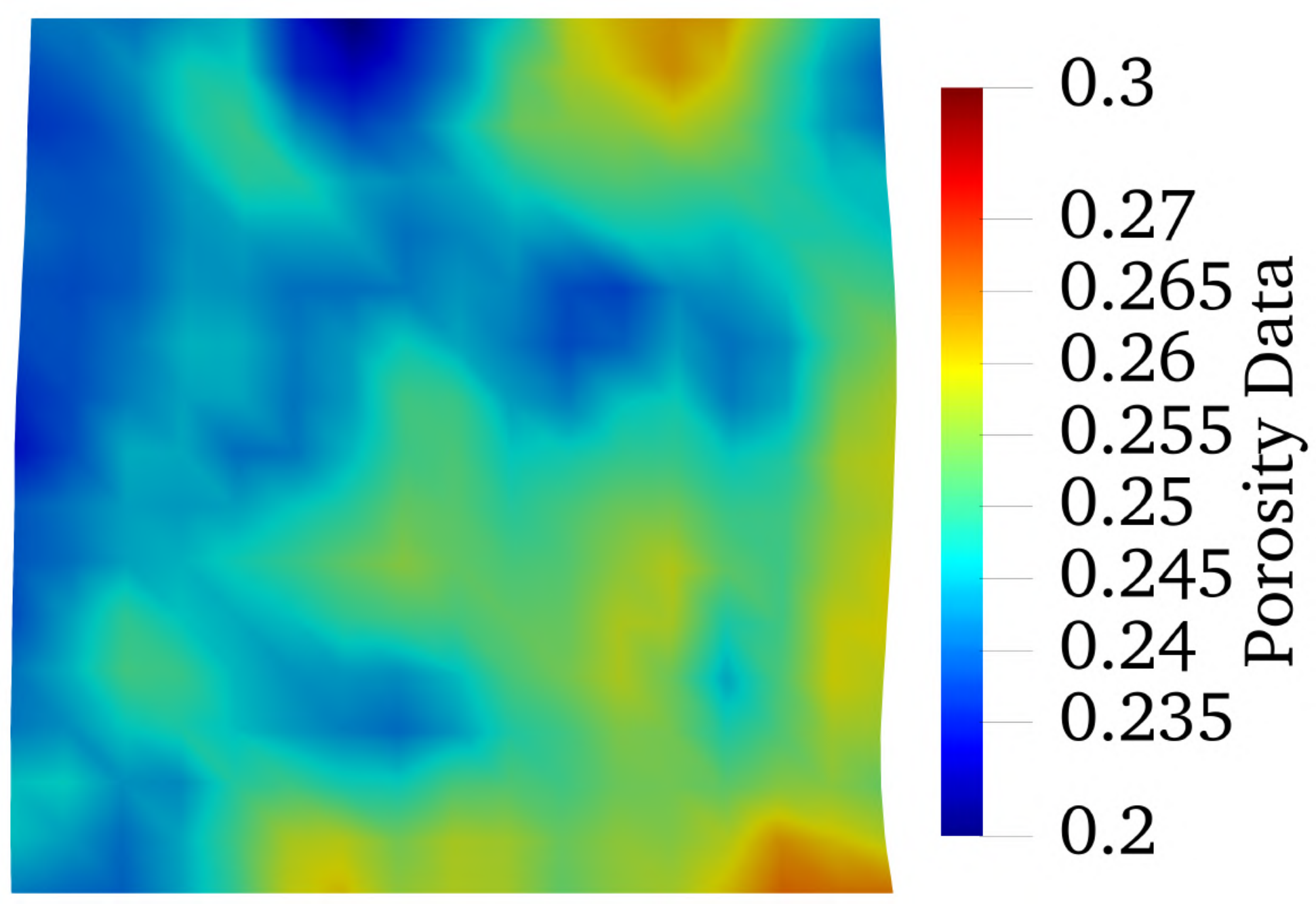}} & 
{\includegraphics[width=\imsz]{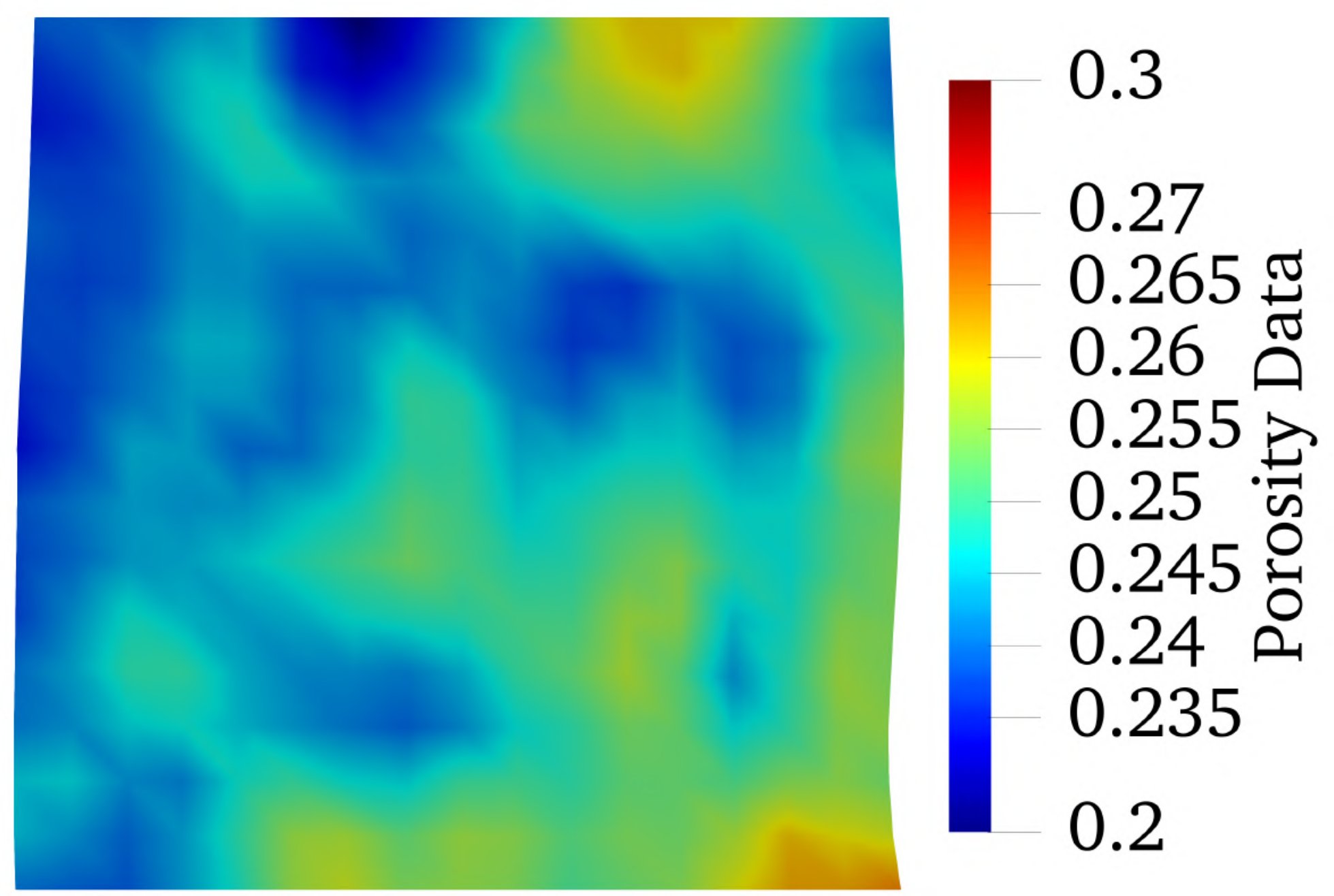}} \\
\hspace{1.cm}$t=25.5 \ [\mathrm{days}]$ &
\hspace{1.cm}$t=50.5 \ [\mathrm{days}]$\\
\end{tabular}
  \caption{Contours of assigned porosity on the deformed domain.
   \label{fig::poro-contour-berea}}
\end{figure}

\begin{figure}[!h]
\newcommand\imsz{6cm}
\newcommand\labeloffset{2cm}
\centering
\begin{tabular}{ m{\imsz} m{\imsz} }
 {\includegraphics[width=\imsz]{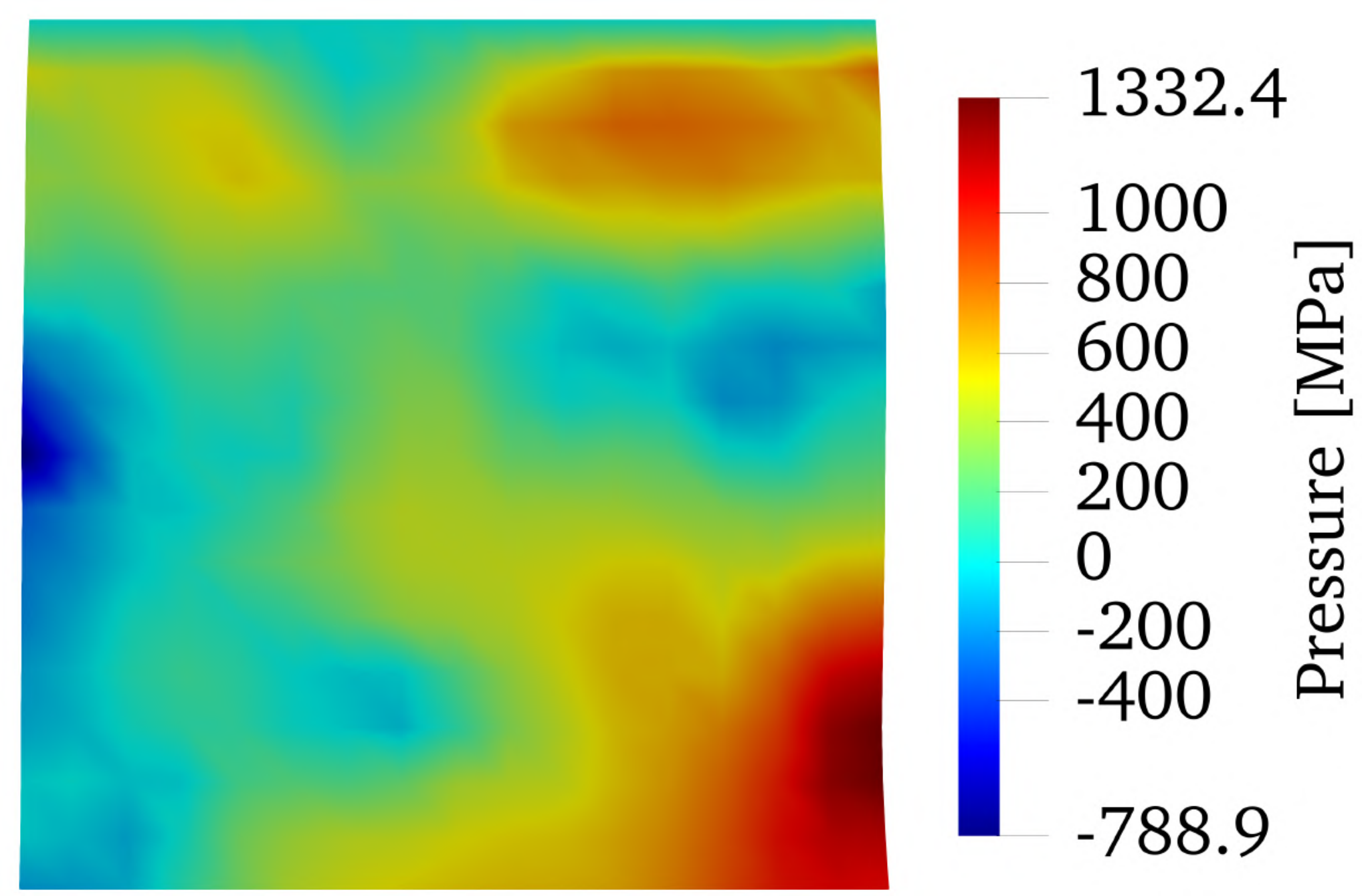}} & 
{\includegraphics[width=\imsz]{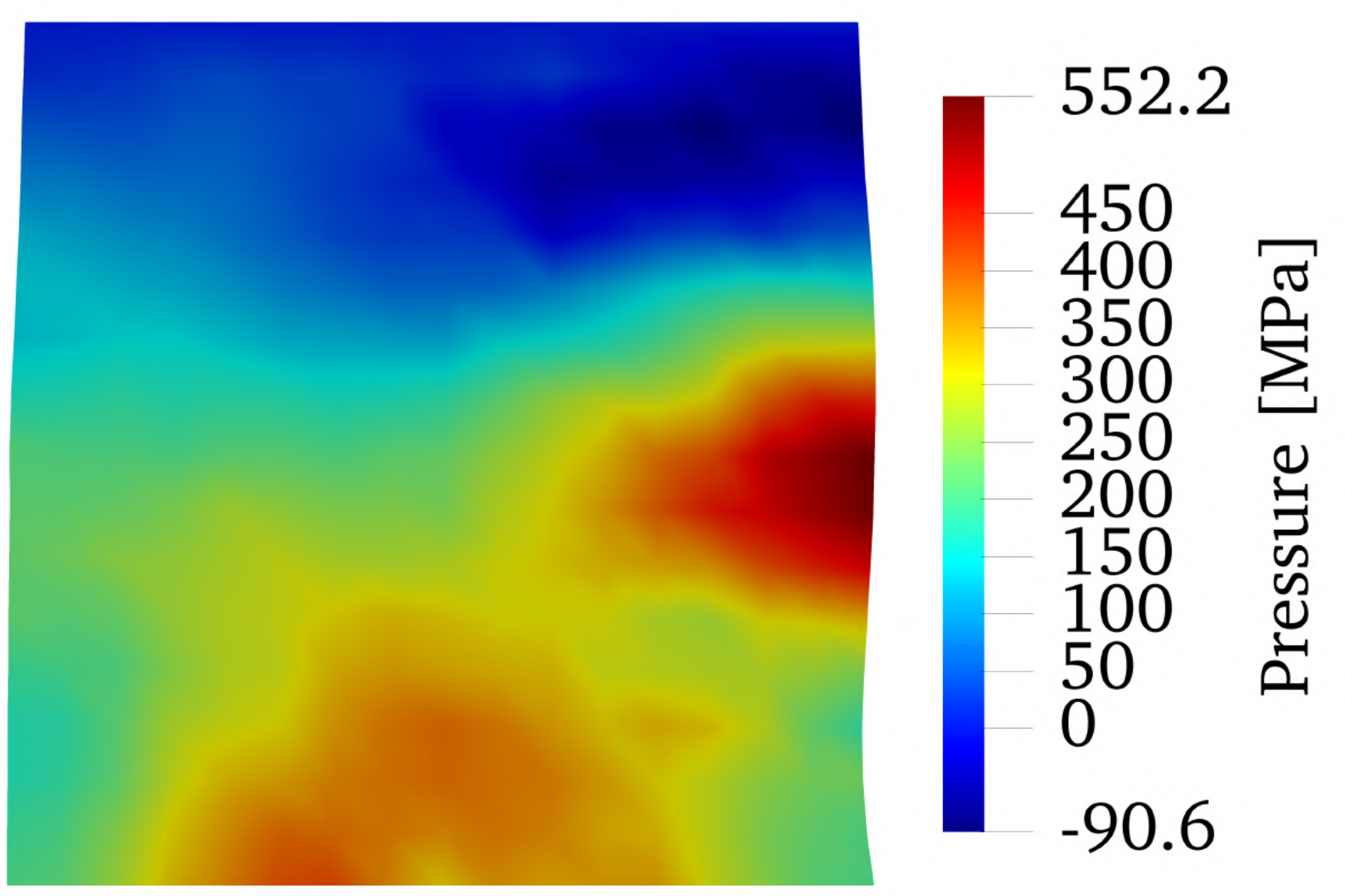}} \\ 
\hspace{1.1cm}$t=12 \ [\mathrm{hours}]$ & 
\hspace{0.9cm}$t=15.5 \ [\mathrm{days}]$\\
{\includegraphics[width=\imsz]{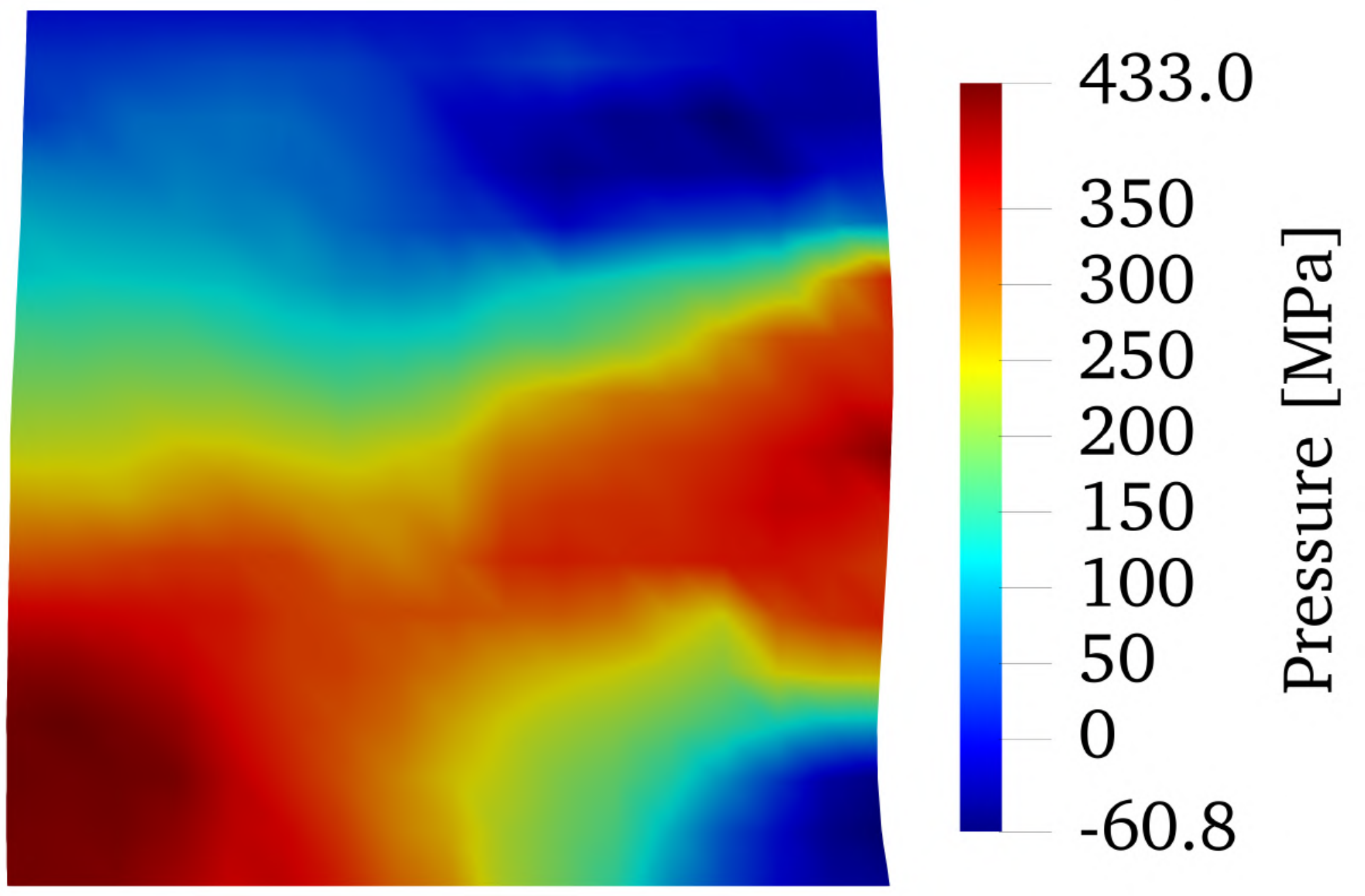}} & 
{\includegraphics[width=\imsz]{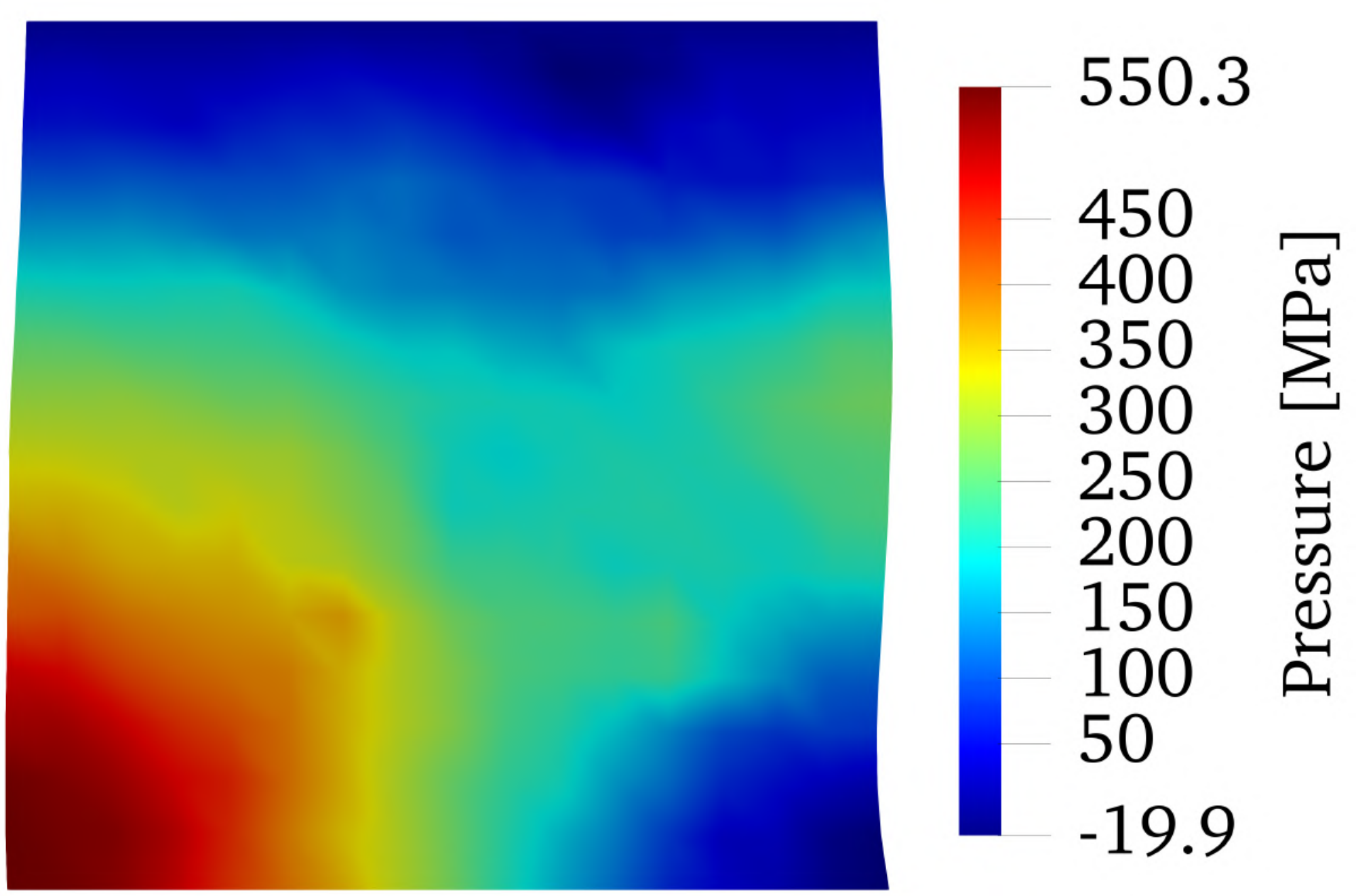}} \\
\hspace{1.cm}$t=25.5 \ [\mathrm{days}]$ &
\hspace{1.cm}$t=50.5 \ [\mathrm{days}]$\\
\end{tabular}
  \caption{Pressure contours on deformed body at different stages of loading.
   \label{fig::press-contour-berea}}
\end{figure}

\section{Conclusion}
We establish both a fully data-driven and hybridized data-driven/model-base framework 
that solve the poroelasticity problems either completely free of hand-crafted constitutive laws
or having them partially replaced by a data-driven scheme. 
By extending the metric-minimization algorithm 
originally proposed for single-physics elasticity problems,
we bypass the usages of material laws when the higher fluctuation hydraulic data may unavoidably lead to under-fitting. To overcome the significant computational time required to search the optimal data point from the discrete dataset, we introduce a k-d tree based nearest neighbor search for an energy metric for poromechanics problems. Comparisons between the k-d tree search and the brute-force search revealed that the k-d tree search may shorten the search time, especially when the dataset is large.

\section{Acknowledgments}
The authors would like to thank Professor Teng-fong Wong for providing the MicroCT images 
of the Berea sandstone and Dr. Ran Ma for programming the FFT solver that calculates the effective permeability of the Berea Sandstone and Dr. Guodong Zhang for providing the permeability database in \fig \ref{fig::fft_data}.
The authors are supported by the  NSF CAREER grant from Mechanics of Materials and Structures program at National Science Foundation under grant contract CMMI-1846875, and the Dynamic Materials and Interactions Program from the Air Force Office of Scientific 
Research under grant contracts FA9550-17-1-0169, and the Earth Materials and Processes
program from the US Army Research Office under grant contract 
W911NF-18-2-0306, as well as the
These supports are gratefully acknowledged. 

The views and conclusions contained in this document are those of the authors, 
and should not be interpreted as representing the official policies, either expressed or implied, 
of the sponsors, including the Army Research Laboratory or the U.S. Government. 
The U.S. Government is authorized to reproduce and distribute reprints for 
Government purposes notwithstanding any copyright notation herein.
\begin{appendices}
\section{K-D Tree Performance}\label{kd-tree-study}
Here we aim to clarify, in detail, why k-d tree search is a major step for the model-free formulation to reduce the computational cost and resource. To focus on just the k-d tree performance, we consider a static, single physics problem with a manufactured polynomial type solution which is fairly a trivial problem. 

Consider a linear Poisson equation governed steady-state response of fluid flow through a cube of unity length (depicted in \fig \ref{fig:cube-prob-def}) as follows:

\begin{align}
	&
	\diver{\vec{q}} + s = 0,
	\\&
	\vec{q} = -k \grad{p},
\end{align}

\begin{figure}[h]
 \centering
\includegraphics[width=0.6\textwidth]{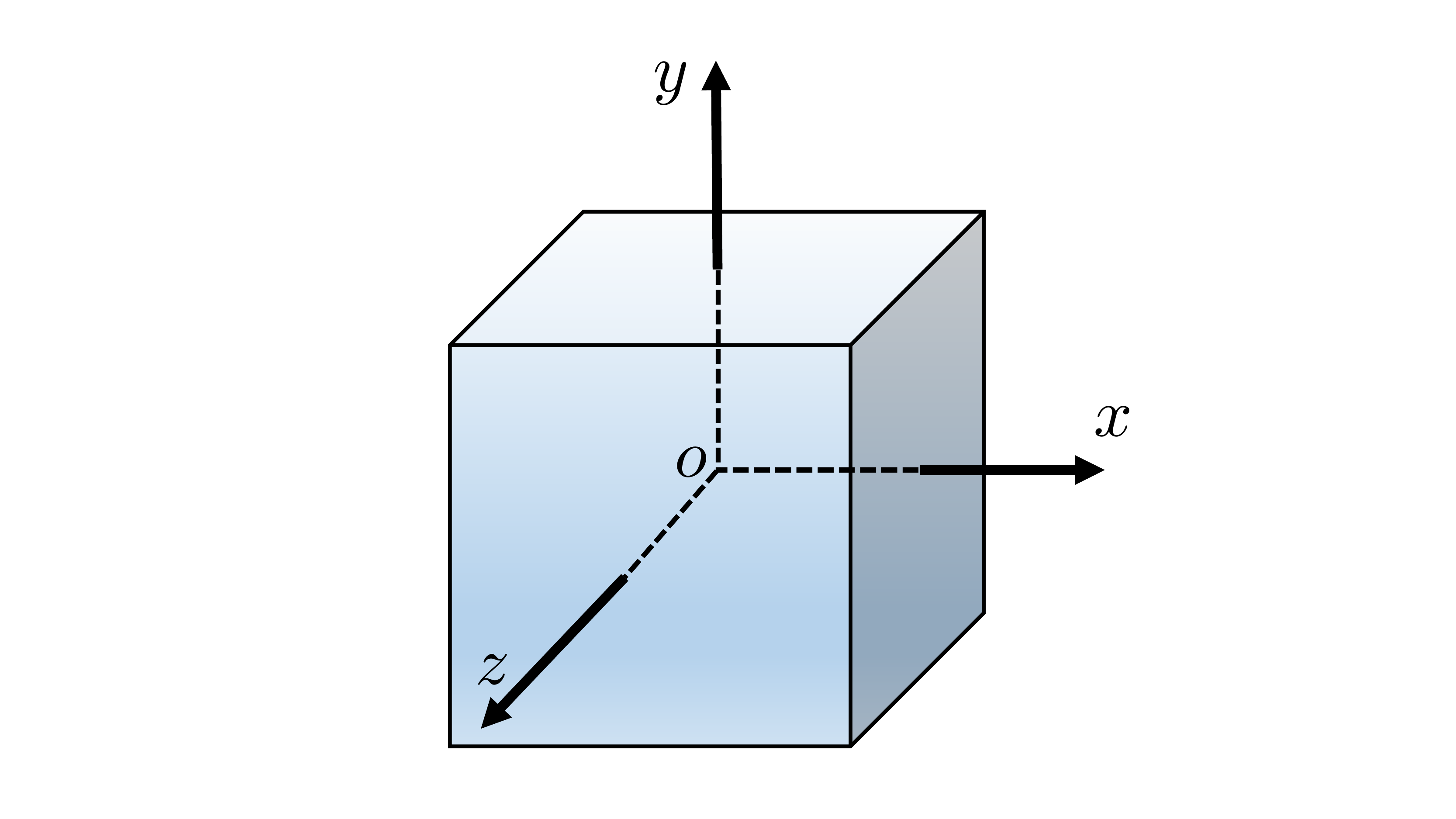}
  \caption{Cubic domain where the origin is placed at its centroid.\label{fig:cube-prob-def}}
\end{figure}

where $\vec{q}$, $s$, $k$, and $p$ are Darcy's velocity, source term, isotropic hydraulic conductivity, and pressures, respectively. The source term is designed to satisfy the solution $p = x^2 + y^2 + z^2$ where $(x,y,z)$ is the spatial coordinate in the cube domain. The data-driven formulation for the Poisson equation is a simplified version of the poromechanics formulation when the solid phase is ignored, and transient effects are not involved. Therefore, we have omitted the extra derivation for the sake of brevity, and interested readers can refer to \citep{nguyen2020variational}.

According to the exact solution for $p$, Dirichlet boundary conditions are applied over all external faces of the cube. Database consists of $\grad{p}$ and $\vec{q}$ pairs are sampled regularly in intervals $-1.1 \le \frac{\partial p }{\partial x}\le 1.1$, $-1.1 \le \frac{\partial p }{\partial y}\le 1.1$, and $-1.1 \le \frac{\partial p }{\partial z}\le 1.1$. For example, if each direction of the pressure gradient is sampled by $64$ equally distance points in the interval $[-1.1, 1.1]$, then a database with $64^3$ rows and $6$ columns has been artificially generated. In this problem, seven data sets ranging from $2^3$ to $128^3$ points are created. Hydraulic conductivity $k$ is set to unity. We use $4096$ structured hexahedron elements to discretize the cube. Trilinear Lagrangian basis functions are utilized, and numerical integration is calculated by one quadrature point.

Since the data-driven formulation might be sensitive to the initial random data assignment to each quadrature point at the first fixed-point iteration,  we have run the same BVP for each algorithm three times. Therefore, in total, $42$ time records are included in \fig \ref{fig:time-kdVSbf-3D} and \fig \ref{fig:time-kd-3D}(b). These figures suggest that the simulation time in this particular problem is relatively insensitive to the random initialization.

As shown in \fig \ref{fig:time-kdVSbf-3D}, the brute-force approach is more efficient than the k-d tree for small data sets ($N \le 16^3$), but after a threshold k-d tree exponentially outperforms the brute-force algorithm. Note that for $N=128^3$ the k-d tree search is almost two orders of magnitude more efficient. \Fig \ref{fig:time-kd-3D}(a) depicts the tree construction time which is an offline process. Time expressed in \fig \ref{fig:time-kd-3D}(b) is the simulation time from beginning to end, including post-processing, subtracted by the tree construction time.

\begin{figure}[t]
  \centering
  \subfigure[]{\includegraphics[width=0.5\textwidth]{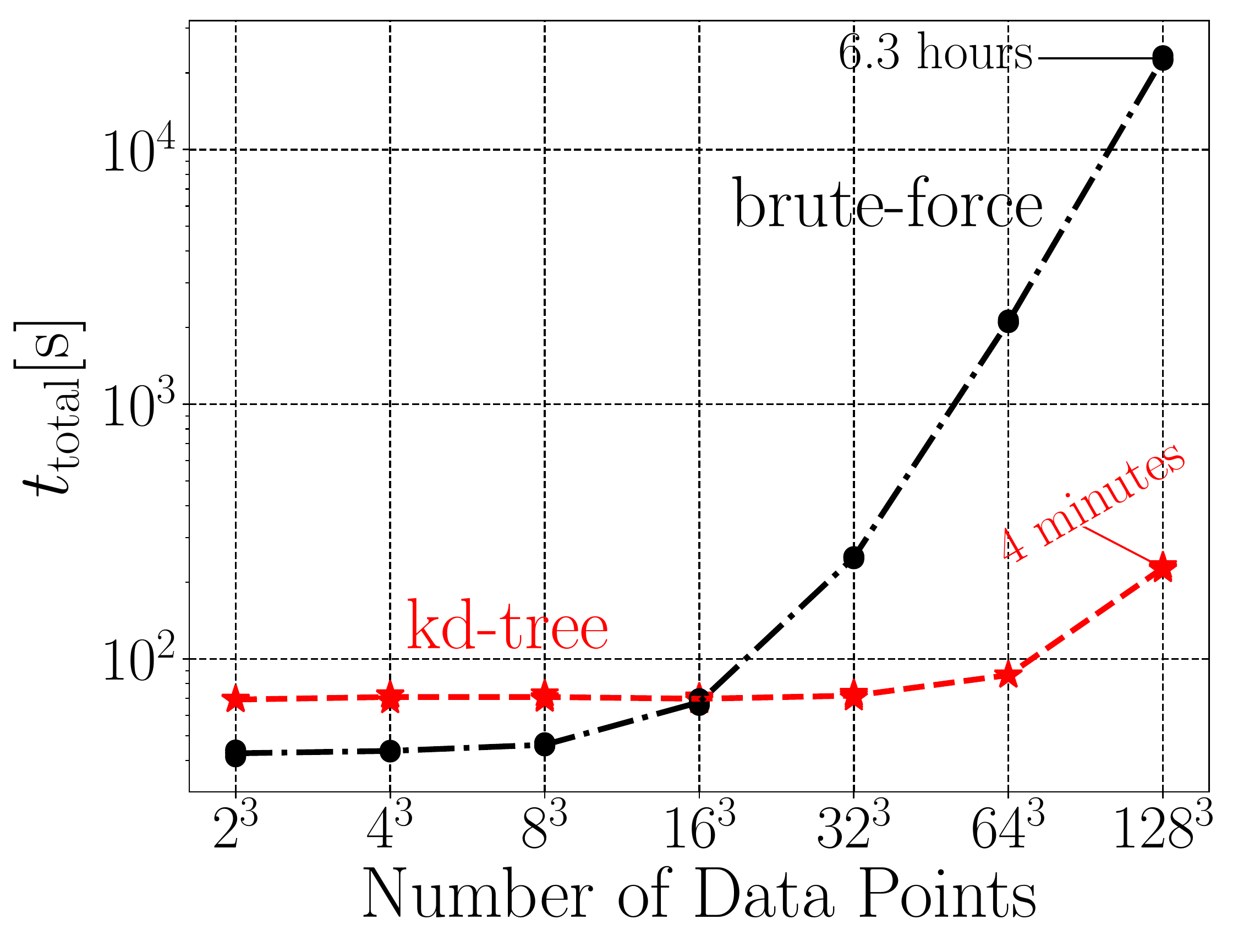}}
  \caption{ A comparison of total simulation time (including post-processing) from start to end for seven data sets with different sizes. Each data set, e.g. data set corresponding to $N=32^3$, is used for three simulations based on the k-d tree search and three simulations based on the brute-force search algorithms. All the $42$ time records are included in this figure. Dash and dash-dot lines pass through the mean value of points corresponding to the k-d tree and brute-force categories, respectively. Note that each simulation starts with a random assignment of data at the first fixed-point iteration.\label{fig:time-kdVSbf-3D}}
\end{figure}

\begin{figure}[t]
  \centering
  \subfigure[]{\includegraphics[width=0.45\textwidth]{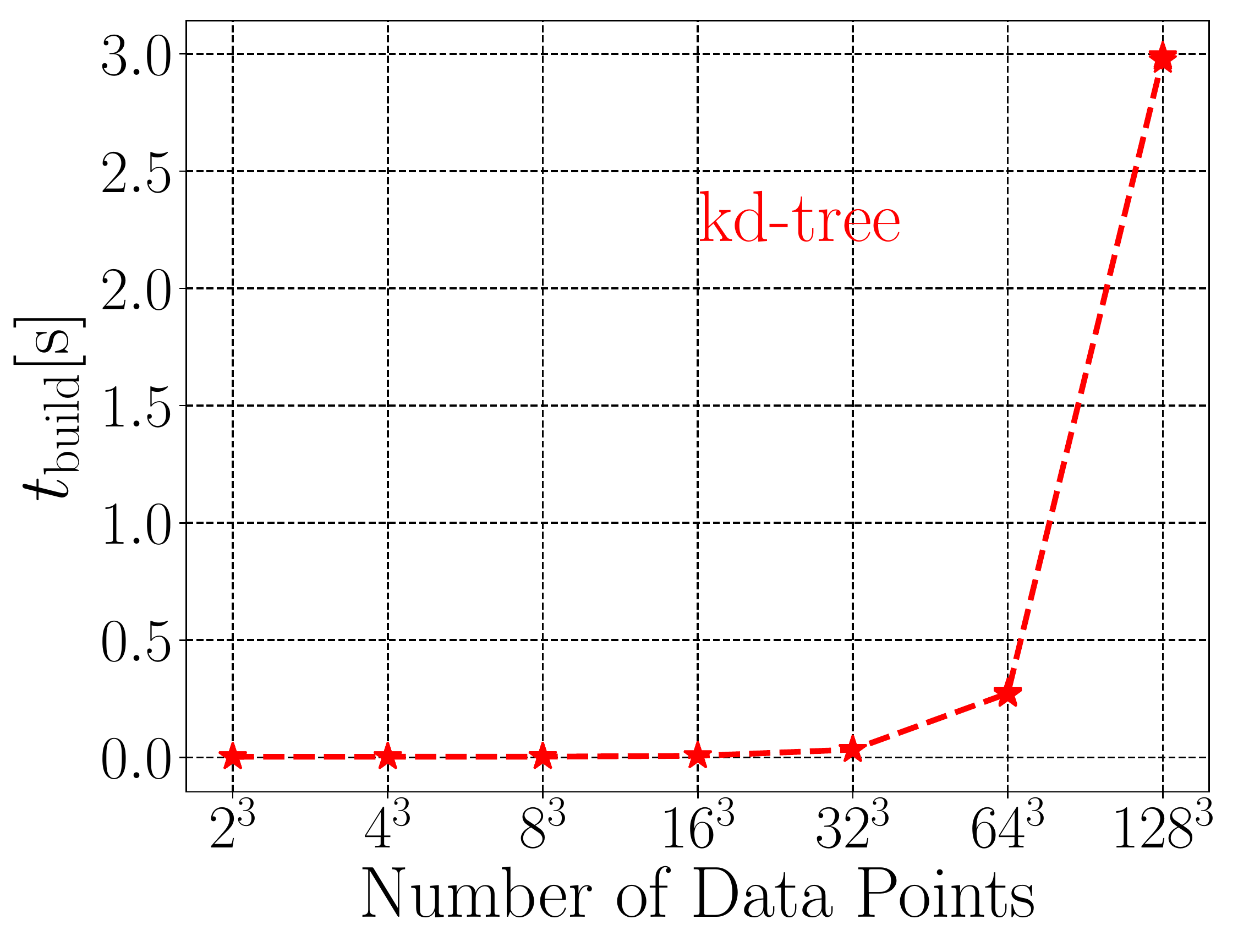}}
  \hspace{0.01\textwidth}
  \subfigure[]{\includegraphics[width=0.45\textwidth]{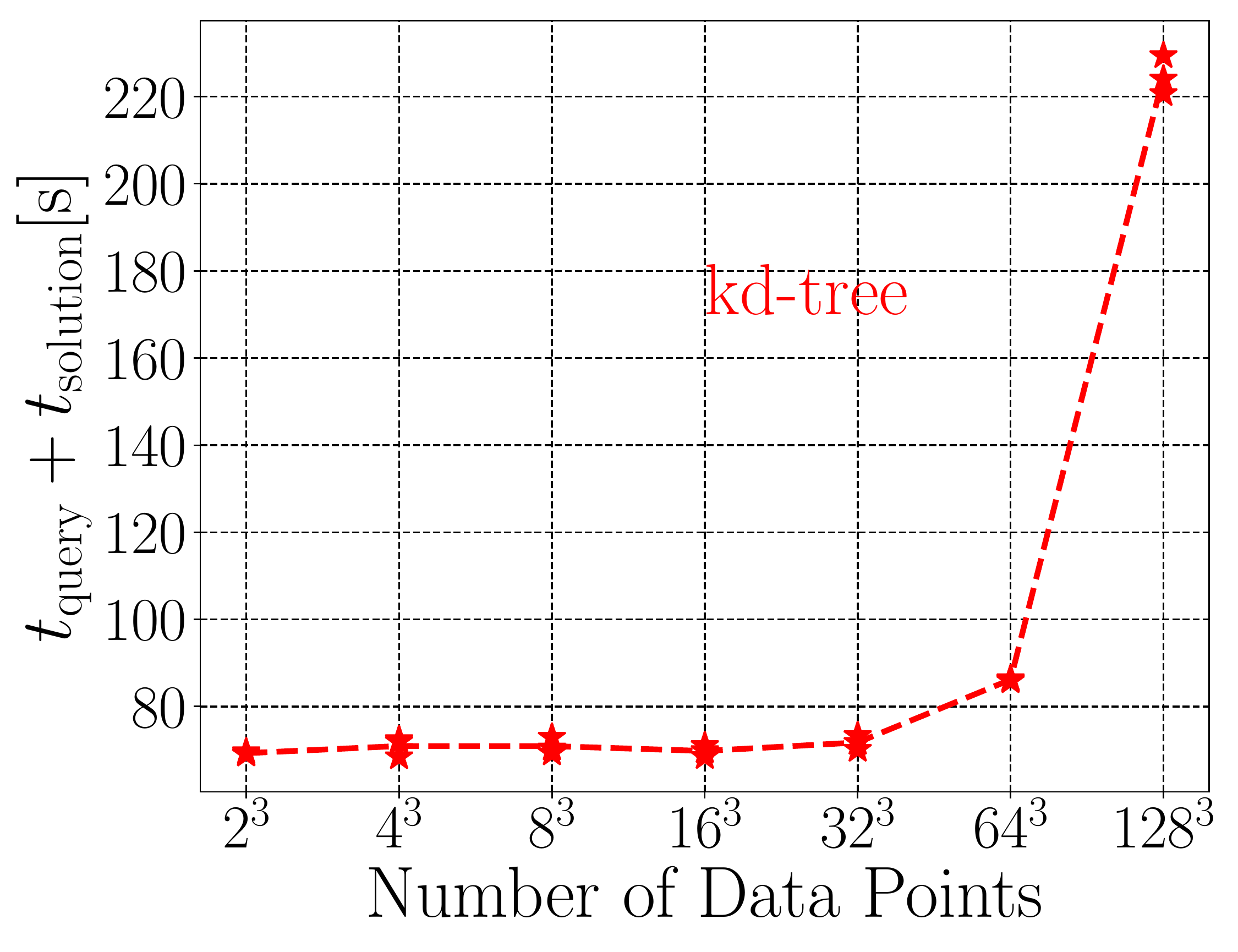}}
  \caption{(a) shows the tree construction time before the simulation for each data set. (b) shows the total time of simulation (including post-processing) subtracted by the tree construction time reported in (a). In (b), dash line passes through the mean value of time records for three trials corresponding to each data set (see caption \fig \ref{fig:time-kdVSbf-3D} for more information).\label{fig:time-kd-3D}}
\end{figure}

\section{Terzaghi's Analytical Solution}
\label{appx::terzaghi}
The exact solution for pressure and displacement profiles used in \ref{sec::problem-terzaghi} are obtained from the following equations (see \citep{wang2000theory} and \citep{Castelletto2015}):
\begin{equation}
 p(y,t) = \frac{4bM\bar{t}_y}{\pi (E+b^2M)} \sum_{i=0}^{\infty} \frac{(-1)^{i+1}}{2i+1} \exp\left(-\frac{(2i+1)^2 \pi^2 c_f}{4H^2}t\right) \cos\left( \frac{(2i+1)\pi}{2H} y \right),
\end{equation}

\begin{equation}
 u_y(y,t) = \frac{y \bar{t}_y}{E+b^2M} + \frac{b^2M\bar{t}_y}{E(E+b^2M)} \left[ y -  \frac{8H}{\pi^2} \sum_{i=0}^{\infty} \frac{(-1)^{i+1}}{(2i+1)^2} \exp\left(-\frac{(2i+1)^2 \pi^2 c_f}{4H^2}t\right) \sin\left( \frac{(2i+1)\pi}{2H} y \right) \right],
\end{equation}
where height $H=1\mathrm{m}$ and $0 \le y \le 1$ for the problem in \fig \ref{fig::comp-pess-disp-fullyDD-FEM-exact}. For the material properties in \ref{tab:terzaghi-params}, fluid diffusivity coefficient is $c_f = \frac{2kME}{b^2 M+E}$. Note that all the above relations are simplified for zero Poisson's ratio.

\section{Stress Relaxation Analytical Solution}\label{appx:strs-relax}
The following exact equation for traction history at the top boundary is used in \fig \ref{fig::strs-relax-top-strs-height-allDD} (see \citep{haider2007application}):

\begin{multline}
	t_z(z=H, t) = \frac{ \dot{\bar{u}}_z (2G+\lambda)}{H} \left(
	t \mathrm{Hev}(t_{\mathrm{ramp}} - t )  +  t_{\mathrm{ramp}} \mathrm{Hev}( t - t_{\mathrm{ramp}} ) 
	\right) 
	\\+
	\frac{2\dot{\bar{u}}_z \mu_{\mathrm{f}} H}{k} \sum_{i=0}^{\infty} \frac{1}{i^2\pi^2} 
	\left(
	\mathrm{Hev}(  t_{\mathrm{ramp}}  - t)
	( 1- e^{-a_i t} )
	+
	\mathrm{Hev}(  t  - t_{\mathrm{ramp}}  )
	( e^{-a_i ( t- t_{\mathrm{ramp}} ) }  - e^{-a_i t} )
\right),
\end{multline}
where $a_i = i^2\pi^2k(2G+\lambda)/H^2$ and $\mathrm{Hev}(\cdot)$ is the Heaviside step function, $\mathrm{Hev}(x < 0) = 0$ and $\mathrm{Hev}(x \ge 0) = 1$. $G$ and $\lambda$ are shear modulus and Lamé's parameter, respectively. $ \dot{\bar{u}}_z$ is the rate of applied displacement.

\end{appendices}
\bibliographystyle{plainnat}
\bibliography{main}

\end{document}